\documentclass[journal]{IEEEtran}
\usepackage{amsmath}
\usepackage{amsfonts}       % blackboard math symbols
\usepackage{multirow}

\usepackage{ifpdf}
\usepackage{cite}
\ifCLASSINFOpdf
   \usepackage[pdftex]{graphicx}
\else
   \usepackage[dvips]{graphicx}
\fi
\usepackage{amsmath}
\usepackage{algorithmic,algorithm}
\usepackage{array}
\ifCLASSOPTIONcompsoc
  \usepackage[caption=false,font=normalsize,labelfont=sf,textfont=sf]{subfig}
\else
  \usepackage[caption=false,font=footnotesize]{subfig}
\fi

\usepackage{xspace}
\makeatletter
\newcommand\onedot{\futurelet\@let@token\@onedot}
\def\@onedot{\ifx\@let@token.\else.\null\fi\xspace}
\def\eg{\emph{e.g}\onedot} 
\def\ie{\emph{i.e}\onedot} 
 
\def\etal{\emph{et al}\onedot}
\makeatother
\usepackage{color}
\usepackage[utf8x]{inputenc}
\begin{document}
\title{Unsupervised and Unregistered Hyperspectral Image Super-Resolution with Mutual Dirichlet-Net}


%\author{Michael~Shell,~\IEEEmembership{Member,~IEEE,}
%        John~Doe,~\IEEEmembership{Fellow,~OSA,}
%        and~Jane~Doe,~\IEEEmembership{Life~Fellow,~IEEE}% <-this % stops a space
%\thanks{M. Shell was with the Department
%of Electrical and Computer Engineering, Georgia Institute of Technology, Atlanta,

\author{Ying~Qu, \IEEEmembership{Member,~IEEE,}
Hairong~Qi, \IEEEmembership{Fellow,~IEEE,}
Chiman~Kwan \IEEEmembership{Senior Member,~IEEE,}
Naoto Yokoya, \IEEEmembership{Member,~IEEE,}
and Jocelyn Chanussot, \IEEEmembership{Fellow,~IEEE}
\thanks{Ying~Qu, and Hairong~Qi are with the Advanced Imaging and Collaborative Information Processing Group, Department of Electrical Engineering and Computer Science, University of Tennessee, Knoxville, TN 37996 USA (e-mail: yqu3@vols.utk.edu;  hqi@utk.edu).}
\thanks{Chiman~Kwan is with Applied Research LLC, Rockville, MD, 20850 USA (e-mail: chiman.kwan@arllc.net)}
\thanks{Naoto Yokoya is with RIKEN Center for
Advanced Intelligence Project (AIP) Tokyo, 103-0027, Japan. (e-mail:
naoto.yokoya@riken.jp)}
\thanks{Jocelyn Chanussot is with the Univ. Grenoble Alpes, Inria, CNRS, Grenoble INP, LJK, Grenoble, 38000, France.(e-mail: jocelyn@hi.is).}}

% and Aerospace Information Research Institute, Chinese Academy of Sciences, Beijing, PR China


\markboth{IEEE TRANSACTIONS ON GEOSCIENCE AND REMOTE SENSING}%
{Shell \MakeLowercase{\textit{et al.}}: Bare Demo of IEEEtran.cls for IEEE Journals}



\maketitle

\begin{abstract}
\textcolor{blue}{(Please find the final version from IEEE Transactions on Geoscience and Remote Sensing on IEEE Xplore. The code has been released on GitHub at https://github.com/yingutk/u2MDN.)}
Hyperspectral images (HSI) provide rich spectral information that has contributed to the successful performance improvement of numerous computer vision and remote sensing tasks. However, it can only be achieved at the expense of images' spatial resolution. Hyperspectral image super-resolution (HSI-SR) thus addresses this problem by fusing low resolution (LR) HSI with multispectral image (MSI) carrying much higher spatial resolution (HR). Existing HSI-SR approaches require the LR HSI and HR MSI to be well registered and the reconstruction accuracy of the HR HSI relies heavily on the registration accuracy of different modalities. In this paper, we propose an unregistered and unsupervised mutual Dirichlet-Net ($u^2$-MDN) to exploit the uncharted problem domain of HSI-SR \textit{without the requirement of multi-modality registration}. The success of this endeavor would largely facilitate the deployment of HSI-SR since registration requirement is difficult to satisfy in real-world sensing devices. The novelty of this work is three-fold. First, to stabilize the fusion procedure of two unregistered modalities, the network is designed to extract spatial and spectral information of two modalities with different dimensions through a shared encoder-decoder structure. Second, the mutual information (MI) is further adopted to capture the non-linear statistical dependencies between the representations from two modalities (carrying spatial information) and their raw inputs. By maximizing the MI, spatial correlations between different modalities can be well characterized to further reduce the spectral distortion. We assume the representations follow a similar Dirichlet distribution for its inherent sum-to-one and non-negative properties. Third, a collaborative $l_{2,1}$ norm is employed as the reconstruction error instead of the more common $l_2$ norm to better preserve the spectral information. Extensive experimental results demonstrate the superior performance of $u^2$-MDN as compared to the state-of-the-art.

\end{abstract}

\begin{IEEEkeywords}
Hyperspectral image, unregistered, super-resolution, mutual information, unsupervised deep learning
\end{IEEEkeywords}

% \fi
\IEEEpeerreviewmaketitle

\section{Introduction}
\label{sec:intro}
Hyperspectral image (HSI) collects hundreds of contiguous spectral representations of objects, which demonstrates advantages over the conventional multispectral image (MSI) or RGB image with much less spectral information ~\cite{chakrabarti2011statistics,bioucas2012hyperspectral}. Compared to conventional images, the rich spectral information of HSI can effectively distinguish visually similar objects that actually consist of different materials. Thus, HSI has been shown to enhance the performance of a wide range of computer vision and remote sensing tasks, such as, object recognition and classification~\cite{kwan2006novel,yokoya2017hyperspectral, haut2018active}, segmentation~\cite{aydav2018classification}, tracking~\cite{Uzkent_2017_CVPR_Workshops}, environmental monitoring~\cite{plaza2011foreword}, and change detection~\cite{borengasser2007hyperspectral}. 

\begin{figure}[t]
\setlength{\abovecaptionskip}{0.cm}
\setlength{\belowcaptionskip}{0.cm}
	\begin{center}
		\subfloat[]{\includegraphics[width=0.245\linewidth]{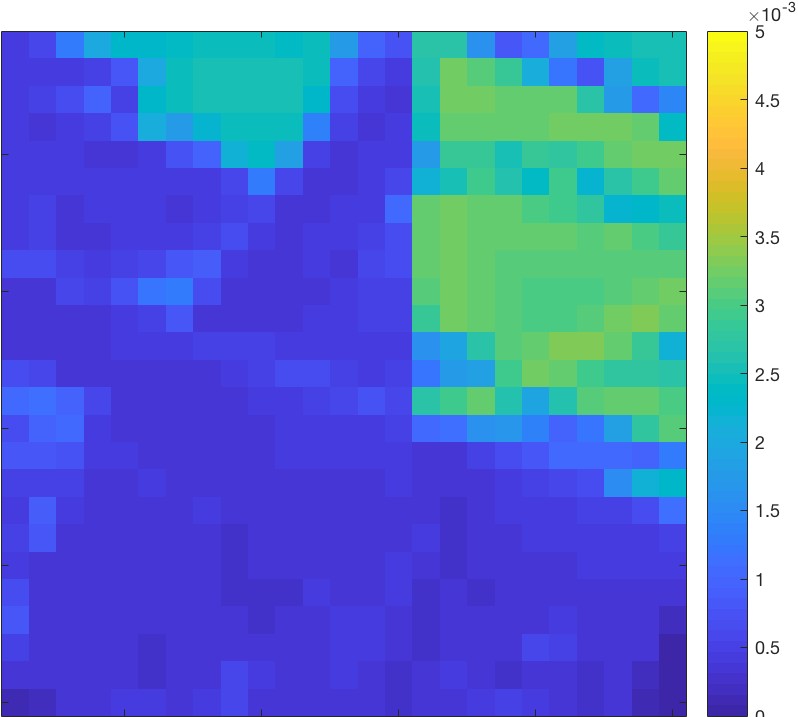}}
		\subfloat[]{\includegraphics[width=0.245\linewidth]{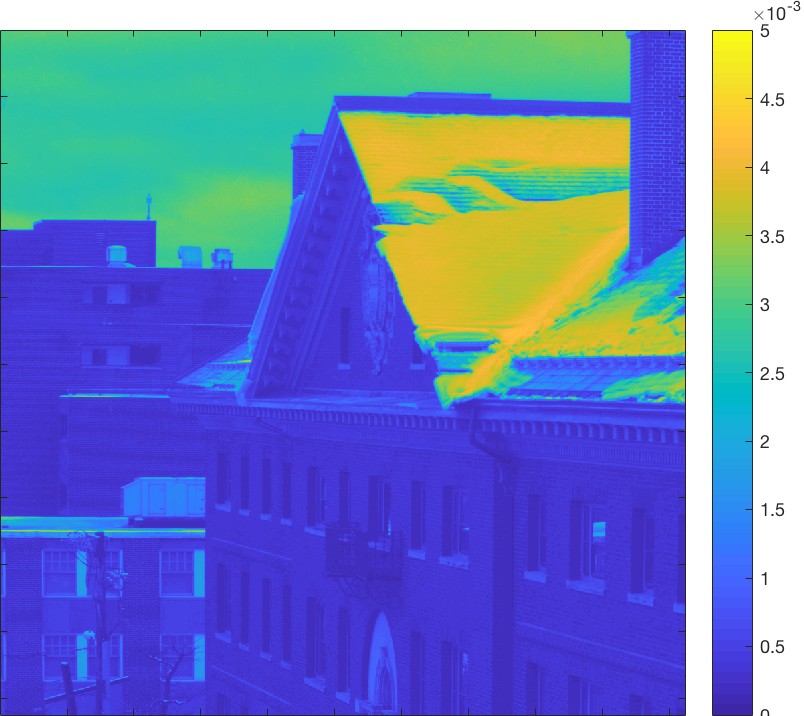}}
		\subfloat[]{\includegraphics[width=0.245\linewidth]{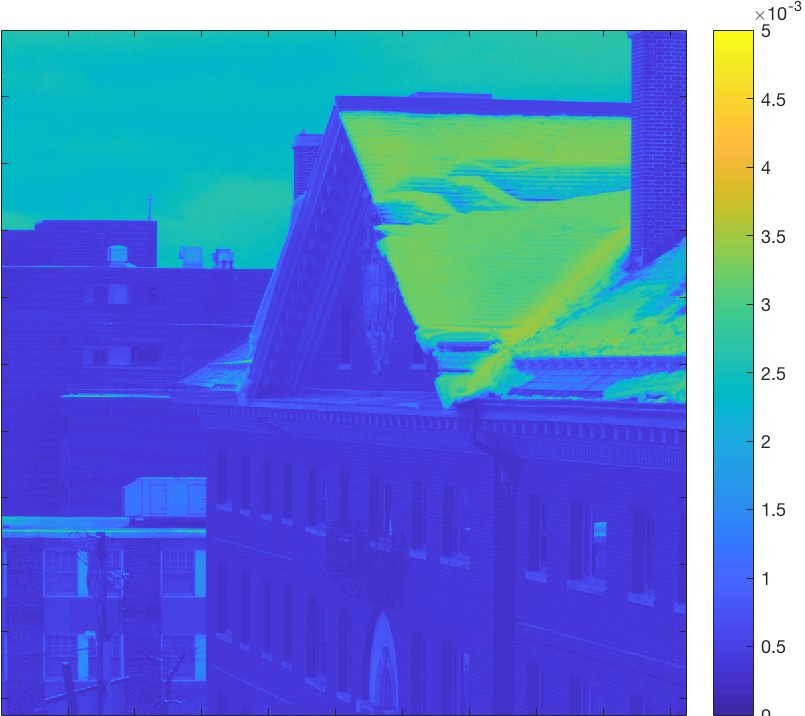}}
		\subfloat[]{\includegraphics[width=0.245\linewidth]{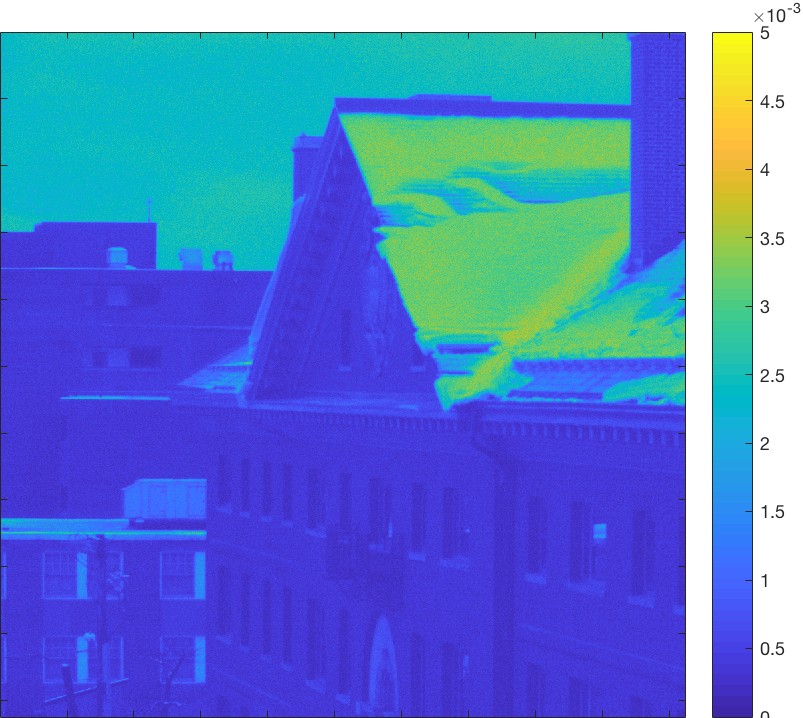}}
	\end{center}
	\caption{Unregistered hyperspectral image super-resolution.(a) First band of the 20 degree rotated and cropped LR HSI with 38\% information missing. (b) First band of the HR MSI. (c) First band of the reconstructed HR HSI  by the proposed methods. (d) First band of the reference HR HSI.}
	\label{fig:sample}
\end{figure}

During the HSI acquisition process, the finer the spectral resolution, the smaller the radiation energy that can reach the sensor for a particular spectral band within a narrow wavelength range. Thus, the high spectral resolution of HSI can only be achieved at the cost of its spatial resolution due to the hardware limitations ~\cite{kawakami2011high,akhtar2015bayesian}. On the contrary, we can obtain conventional MSI or RGB with a much higher spatial resolution by integrating the radiation energy over broad spectral bands which inevitably reduces their spectral resolution significantly~\cite{lanaras2015hyperspectral}. To improve the spatial resolution of HSI for better application performance, a natural way is to fuse the high spectral information extracted from HSI with the high-resolution spatial information extracted from conventional images to yield high resolution images in both spatial and spectral domains~\cite{vivone2015critical,yokoya2017hyperspectral}. This procedure is referred to as \textit{hyperspectral image super-resolution (HSI-SR)}~\cite{akhtar2015bayesian,lanaras2015hyperspectral}. %HSI-SR has been intensively studied and can be broadly divided into three categories 

HSI-SR can be broadly divided into three categories, traditional component substitution (CS) ~\cite{thomas2008synthesis,aiazzi2007improving} and multi-resolution analysis (MRA) based methods~\cite{aiazzi2006mtf}, matrix factorization based, and Bayesian-based approaches~\cite{loncan2015hyperspectral,yokoya2017hyperspectral}. Although HSI-SR has been intensively studied, spectral distortion can be easily introduced during the optimization procedure of methods from these categories. Recently, there have been several attempts to address the HSI-SR problem with deep learning where the mapping function between the LR HSI and HR HSI is learned using different frameworks~\cite{chang2018hsi,arun2020cnn}. However, the deep learning-based approaches are generally limited to handle image pairs with large spatial-scale differences and the learned mapping function may not be readily adapted to reconstruct HR HSI possessing different spectral characteristics or acquired from different sensors.
 
Despite a plethora of works on HSI-SR, all current approaches have at least one pre-requisite to solving the problem of HSI-SR, i.e., the two input modalities (HSI and MSI) must be well registered, and the quality of the reconstructed HR HSI relies heavily on the registration accuracy~\cite{yokoya2017hyperspectral,bioucas2012hyperspectral,zhou2017nonrigid,qu2018unsupervised,baronti2011theoretical}.  According to previous works, there are a few methods that introduce registration as a pre-step before data fusion~\cite{van2012multi,loncan2015hyperspectral,wei2015fast}. However, these pre-steps can only handle small-scale differences, \eg, two pixels/eight pixels offset in LR HSI/HR MSI \cite{zhou2017nonrigid}. Moreover, even in the registration community, HSI and MSI registration is a challenging problem itself as one pixel in LR HSI may cover hundreds of pixels in the corresponding HR MSI. The spectral difference is also large that both the spectral response function (SRF) and multi-band images have to be taken into consideration during registration ~\cite{chui2003new,fan2010spatial,myronenko2010point,ma2015robust,zhou2017nonrigid}. %Thus, most registration approaches can only handle small scale differences.

\begin{figure}[htb]
\setlength{\abovecaptionskip}{0.cm}
\setlength{\belowcaptionskip}{0.cm}
	\begin{center}
		\subfloat[]{\includegraphics[width=0.3\linewidth]{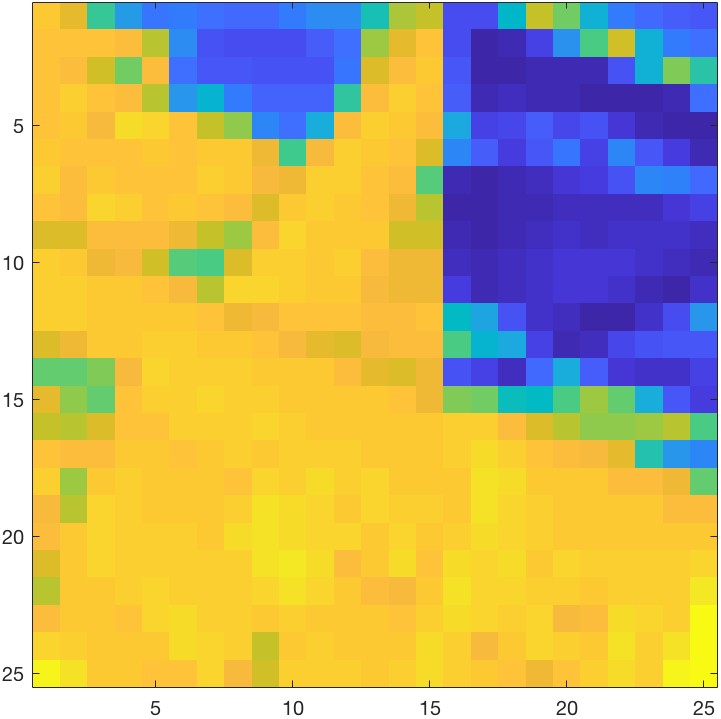}\label{fig:rgbresult:a}}\hspace{10mm}
		\subfloat[]{\includegraphics[width=0.3\linewidth]{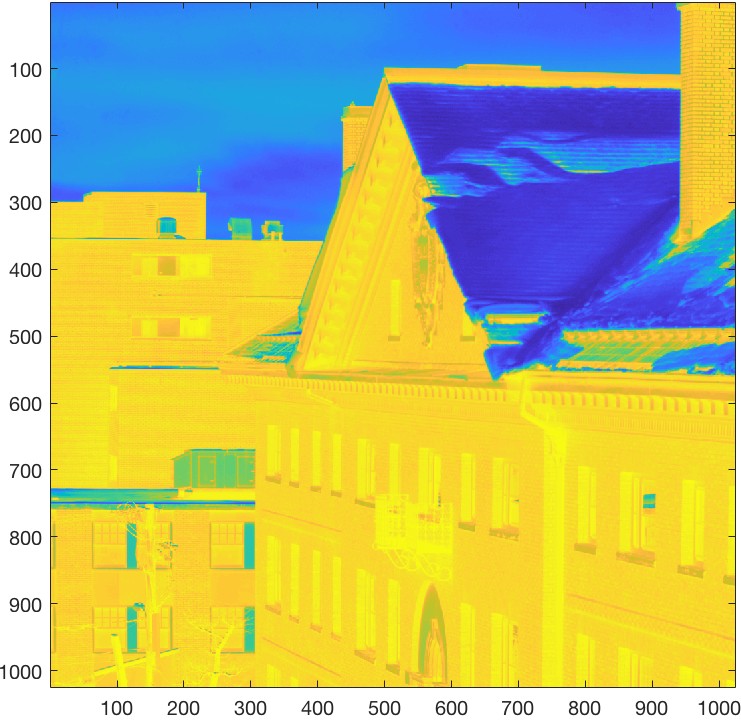}}
	\end{center}
	\caption{Learned hidden representations from unregistered (a) low resolution HSI and (b) high resolution MSI, respectively, as shown in Fig.~\ref{fig:sample}.}
	\label{fig:repres}
\end{figure}

% Thus substantial extensions are made to our previous work, to address the above problems. 
In this paper, an unsupervised network structure is proposed, aiming to solve the HSI-SR problem directly without multi-modality registration. An example is shown in Fig.~\ref{fig:sample}. We address the problem based on the assumption that, the pixels in the overlapped region of HR HSI and HR MSI can be approximated by a linear combination of the same spectral information (spectral bases) with the same corresponding spatial information (representations), which indicates how the spectral basis is constructed for each pixel. Since LR HSI is the down-sampled version of the HR HSI, ideally, its representations should be correlated with that of the HR MSI and HR HSI, \ie, they should follow similar patterns and distributions although possessing different resolutions, as shown in Fig.~\ref{fig:repres}. Therefore, to reconstruct HR HSI with minimum spectral distortion, the network is designed to decouple both the LR HSI and HR MSI into spectral bases and representations, such that their spectral bases are shared and their representations are correlated with each other. 

%to extract most correlated spectral bases and representations from two unregistered modalities, the network projects the LR HSI onto the same statistical space as HR MSI, such that the bases and representations could be extracted through the same encoder-decoder structure. 

The novelty of this work is three-fold. First, to stabilize the fusion procedure for two unregistered modalities, the network extracts both the spectral and spatial information of the multi-modalities through the same encoder-decoder structure, by projecting the LR HSI onto the same statistical space as HR MSI, as illustrated in Fig.~\ref{fig:flow}. The representations of the network are encouraged to follow a Dirichlet distribution to naturally meet the non-negative and sum-to-one physical constraints. Second, to prevent spectral distortion, we further adopt mutual information (MI) to extract optimal and correlated representations from multi-modalities. Since the two-modalities are unregistered, the correlated representations are learned by maximizing the MI between the representations and their own inputs during the network optimization. Third, a collaborative $l_{2,1}$ norm is employed as the reconstruction error instead of commonly used $l_2$ loss, so that the network is able to reconstruct individual pixels as accurately as possible. In this way, the network preserves the spectral information better. With the above design, the proposed network is able to work directly on unregistered images and the spectral distortion of the reconstructed HR HSI can be largely reduced. The proposed method is referred to as unregistered and unsupervised mutual Dirichlet Net, or $u^2$-MDN for short. 

$u^2$-MDN is an extension of our previous work uSDN~\cite{qu2018unsupervised}. However, uSDN is only effective on general HSI-SR problem with well-registered LR HSI and HR MSI. Here, we have made substantial extensions to address the challenges of HSI-SR with unregistered multi-modalities. To the best of our knowledge, this is the first effort to solving the HSI-SR problem directly on unregistered image pairs with unsupervised deep learning. The major improvements can be summarized from three perspectives. First, the network structure is different from that of the uSDN. Instead of adopting two deep learning networks as in uSDN, the proposed $u^2$-MDN is specifically designed to extract the representations of multi-modalities with only one encoder-decoder structure, which largely stabilizes the information extraction and fusion procedure given the unregistered multi-modalities. Second, uSDN minimizes spectral distortion of the reconstructed HR HSI by reducing the angular difference of the representations from multiple modalities, which fails to deal with unregistered cases, while the proposed $u^2$-MDN is able to handle both well-registered and unregistered cases by extracting correlated representations with mutual information through the mutual discriminative network. Third, instead of commonly used $l_2$ loss adopted by uSDN, the collaborative $l_{2,1}$ norm is introduced in the proposed $u^2$-MDN to better preserve the spectral information.

\section{Related Work}
\subsection{Hyperspectral Image Super-Resolution}
The problem of HSI-SR originates from multispectral image super-resolution (MSI-SR) in the remote sensing field, where the spatial resolution of MSI is further improved by a high-resolution panchromatic image (PAN). Traditional widely utilized MSI-SR methods can be roughly categorized into two groups: the component substitution (CS) and the multi-resolution analysis (MRA) based approaches. Generally, CS--based approaches~\cite{thomas2008synthesis} project the given data onto a predefined space where the spectral information and spatial information are separated. Subsequently, the spatial component is substituted with the one extracted from PAN~\cite{aiazzi2007improving}. Several methods based on CS have been proposed to address the problem of hyper-sharpening and achieved promising results with different criteria~\cite{schaepman2015advanced,selva2015hyper,selva2018improving}. MRA-based approaches achieve the spatial details by first applying a spatial filter to the HR images. Then the spatial details are injected into the LR HSI~\cite{mallat1989theory, liu2000smoothing, burt1983laplacian,aiazzi2006mtf,loncan2015hyperspectral}. Although these traditional pan-sharpening approaches can be extended to solve the HSI-SR problem, they usually suffer from severe spectral distortions ~\cite{loncan2015hyperspectral,akhtar2015bayesian,dian2017hyperspectral}. %as discussed in Sec.~\ref{sec:intro}.

Recent approaches consist of Bayesian-based and matrix factorization-based methods~\cite{loncan2015hyperspectral,yokoya2017hyperspectral}. Bayesian approaches estimate the posterior distribution of the HR HSI given LR HSI and HR MSI. The unique framework of Bayesian offers a convenient way to regularize the solution space of HR HSI by employing a proper prior distribution such as Gaussian. Different methods vary according to the different prior distributions adopted. Wei \etal proposed a Bayesian Naive method~\cite{wei2014bayesian} based on the assumption that the representation coefficients of HR HSI follow a Gaussian distribution. However, this assumption does not always hold especially when the ground truth HR HSI contains complex textures. Instead of using Gaussian prior, dictionary-based approaches solve the problem under the assumption that HR HSI is a linear combination of properly chosen over-complete dictionary and sparse coefficients~\cite{wei2015hyperspectral}. Simoes \etal proposed HySure~\cite{simoes2015convex}, which takes into account both the spatial and spectral characteristics of the given data. This approach solves the problem through vector-based total variation regularization. Akhtar  \etal~\cite{akhtar2015bayesian} introduced a non-parametric Bayesian strategy to solve the HSI-SR problem. The method first learns a spectral dictionary from LR HSI under the Bayesian framework. Then it estimates the spatial coefficients of the HR MSI by Bayesian sparse coding. Eventually, the HR HSI is generated by combining the spatial dictionary with the spatial coefficients. However, the spectral information extracted from LR HSI may not be the optimal spectral bases for MSI, since MSI is not utilized during the optimization procedure.

Matrix factorization-based approaches have been actively studied recently~\cite{kawakami2011high,yokoya2012coupled,lanaras2015hyperspectral,veganzones2016hyperspectral}, with Kawakami \etal~\cite{kawakami2011high} being the first that introduced matrix factorization to solve the HSI-SR problem. The method learns a spectral basis from LR HSI and then uses this basis to extract sparse coefficients from HR MSI with non-negative constraints. Similar to Bayesian-based approaches, the HR HSI is generated by linearly combining the estimated bases with the coefficients. Yokoya \etal~\cite{yokoya2012coupled} decomposed both the LR HSI and HR MSI alternatively to achieve the optimal non-negative bases and coefficients that are used to generate HR HSI. Wycoff \etal~\cite{wycoff2013non} solved the problem with alternating direction method of multipliers (ADMM). Lanaras \etal~\cite{lanaras2015hyperspectral} further improved the fusion results by introducing a sparse constraint. However, most methods~\cite{yokoya2012coupled,wycoff2013non,lanaras2015hyperspectral} are based on the same assumption that the down-sampling function between the spatial coefficients of HR HSI and LR HSI is known beforehand. In practice, this assumption is not always true due to the complex environmental conditions. 

Most of these approaches focus on the spectral characteristics of the HSI, where the spectral information of the HSI is extracted while the spatial relationship between pixels is untouched. Recently, there have been a few approaches proposed to address the HSI-SR problem based on tensor decomposition~\cite{cstf2018,chang2020weighted, xu2019nonlocal,xu2020hyperspectral}, which explored both the spectral and spatial correlations of the HSI by learning a core tensor and the dictionaries along three dimensions,~\ie, the spectral dimension, and two spatial dimensions. In this way, the information of each dimension can be represented with its own dictionary, while the core-tensor is shared among multi-modalities. Although this formulation works well on well-registered images, it is problematic on unregistered image pairs, since the core-tensor cannot be shared between HSI and MSI with large displacements. In addition, it might limit the reconstruction ability of the method on remote sensing images which have only a few redundant structures on the spatial domain of HSI.\\

Chen~\etal~\cite{Chen2014SIRF} proposed to simultaneously register images during the fusing process. However, it only works on panchromatic and MSI. Zhou~\etal~\cite{zhou2019} proposed an integrated approach for registration and fusion, which addressed the problem of HSI-SR on unregistered image pairs. However, the registration is still a required step, and the fusion and registration are performed independently, which would introduce additional errors during optimization.

\subsection{Deep learning based Super-Resolution}
Deep learning attracts increasing attention for natural image super-resolution since 2014 when Dong \etal first introduced convolution neural network (CNN) to solve the problem of natural image super-resolution and demonstrated state-of-the-art restoration quality \cite{dong2016image}. Ledig \etal proposed a method based on generative adversarial network and skipped residual network. The method employed perceptual loss through the VGG network which can recover photo-realistic textures from heavily down-sampled images~\cite{ledig2016photo}. Usually, natural image SR methods only work up to 8 times upscaling. There have been several attempts to address the MSI-SR or HSI-SR with deep learning in a supervised fashion. In 2015, a modified sparse tied-weights denoising autoencoder was proposed by Huang \textit{et al.}~\cite{huang2015new} to enhance the resolution of MSI. The method assumes that the mapping function between LR and HR PAN is the same as the one between LR and HR MSI. Masi \etal proposed a supervised three-layer SRCNN~\cite{masi2016pansharpening} to learn the mapping function between LR MSI and HR MSI. Similar to~\cite{masi2016pansharpening}, Wei \etal~\cite{wei2017boosting} learned the mapping function with deep residual network. Li \etal~\cite{li2017hyperspectral} solved the HSI-SR problem by learning a mapping function with spatial constraint strategy and convolutional neural network (CNN). Dian \etal~\cite{dian2018deep} initialized the HR HSI from the fusion framework via the Sylvester equation. Then, the mapping function is trained between the initialized HR-HSI and the reference HR HSI through deep residual learning. Xie~\etal~\cite{2019Multispectral} reduced spectral distortions of the reconstructed HR HSI by exploiting the approximate low-rankness prior along the spectral domain of the HSI. However, these supervised deep learning-based methods can not be readily adopted on HSI-SR for real applications %due to the reasons elaborated in Sec.~\ref{sec:intro}.  
due to three reasons. First, the scale differences between LR HSI and HR MSI can reach as large as 10, \ie, one pixel in HSI covers 100 pixels in MSI. In some applications, the scale difference can even be 25~\cite{kwan2017blind} and 30~\cite{kwan2018mars}. But most existing super-resolution methods only work on up to $8$ times upscaling. Second, they are designed to find an end-to-end mapping function between the LR images and HR images under the assumption that the mapping function is the same for different images. However, the mapping function may not remain the same for images acquired with different sensors. Even for the data collected from the same sensor, the mapping function for different spectral bands may not be the same. Thus the assumption may cause severe spectral distortion. Third, training a mapping function is a supervised problem that requires a large dataset, the down-sampling function, and the availability of the HR HSI, making supervised learning unrealistic for HSI.

Recently, we proposed an unsupervised uSDN~\cite{qu2018unsupervised}, which addressed the problem of HSI-SR with deep network models. Specifically, it extracts the spectral and spatial information through two encoder-decoder networks from the two modalities.  The angular difference between the LR HSI and HR MSI representations is minimized to reduce the spectral distortion for every ten iterations. Fu~\etal~\cite{fu2019hyperspectral} proposed an unsupervised CNN-based method for HSI super-resolution, which learns a mapping function between the RGB space and the spectral space with spatial constraint for the HR HSI. Zheng~\etal~\cite{zheng2020coupled} proposed an unsupervised method with learnable downsampling function based on the theory of linear unmixing. These methods can achieve promising results for different HSI datasets. However, they are specifically designed for well-registered image pairs. 

\section{Problem Formulation}
\label{sec:formulate}
Given the LR HSI, $\bar{\mathbf{Y}}_h \in \mathbb{R}^{m\times n\times L}$, where $m$, $n$ and $L$ denote its width, height and number of spectral bands, respectively, and the unregistered HR MSI with overlapped region, $\bar{\mathbf{Y}}_m \in \mathbb{R}^{M \times N \times l}$, where $M$, $N$ and $l$ denote its width, height and number of spectral bands, respectively, the goal is to reconstruct the HR HSI $\bar{\mathbf{X}} \in \mathbb{R}^{M\times N \times L}$ based on the content of HR MSI. In general, MSI has much higher spatial resolution than HSI, and HSI has much higher spectral resolution than MSI, \ie., $M\gg m$, $N \gg n$ and $L \gg l$.

To facilitate the subsequent processing, we unfold the 3D images into 2D matrices, $\mathbf{Y}_h \in \mathbb{R}^{mn\times L}$, $\mathbf{Y}_m \in \mathbb{R}^{MN \times l}$ and $\mathbf{X}\in \mathbb{R}^{MN \times L}$, such that each row represents the spectral reflectance of a single pixel.  Since each pixel in both LR HSI and HR MSI can be approximated by a linear combination of $c$ spectral bases $\mathbf{D}$~\cite{lanaras2015hyperspectral,akhtar2015bayesian,qu2018unsupervised}, the matrices can be further decomposed as
\begin{align}
\begin{split}\label{equ:blrhsi}
&\mathbf{Y}_h = \mathbf{S}_h\mathbf{D}_h
\end{split}\\
\begin{split}\label{equ:bmsi}
&\mathbf{Y}_m = \mathbf{S}_m\mathbf{D}_m
\end{split}\\
\begin{split}\label{equ:bhrhsi}
&\mathbf{X} = \mathbf{S}_m\mathbf{D}_h
\end{split}\end{align}
where $\mathbf{D}_h\in\mathbb{R}^{c \times L}$, $\mathbf{D}_m\in\mathbb{R}^{c \times l}$ denote the spectral bases of LR HSI and HR MSI, respectively. $\mathbf{S}_h\in\mathbb{R}^{mn\times c}$, $\mathbf{S}_m\in\mathbb{R}^{MN\times c}$ denote the coefficients of LR HSI and HR MSI, respectively, Since $\mathbf{S}_h$ or $\mathbf{S}_m$ indicate how the spectral bases are combined for individual pixels at specific locations, they preserve the spatial structure of HSI. Note that the benefit of unfolding the data into 2D matrices is that, the extraction procedure  can decouple each pixel without changing the relationship of the pixel and its neighborhood pixels, thus the reconstructed image has less artifacts~\cite{lanaras2015hyperspectral,akhtar2015bayesian,qu2018unsupervised}.

In real applications, although the areas captured by LR HSI and HR MSI might not be registered well, they always have overlapping regions, and the LR HSI includes all the spectral basis of HR MSI  \ie., they share the same type of materials carrying specific spectral signatures. The relationship between LR HSI and HR MSI can be expressed as 
\begin{align}
\begin{split}\label{equ:chm}
&\mathcal{C}_h \neq \mathcal{C}_m, \quad \mathcal{C}_h \cap \mathcal{C}_m \neq \emptyset,\quad \mathbf{D}_m =\mathbf{D}_h\mathcal{R},
\end{split}
%\begin{split}\label{equ:shm}
%&\mathbf{D}_m =\mathbf{D}_h\mathcal{R},
%\end{split}
\end{align}
where $\mathcal{C}_h$ and  $\mathcal{C}_m$ denote the contents of LR HSI  and HR MSI, respectively.  $\mathcal{R}\in\mathbb{R}^{L \times l}$ is the prior transformation matrix of sensor~\cite{kawakami2011high, yokoya2012coupled, wei2015hyperspectral,loncan2015hyperspectral,simoes2015convex,vivone2015critical,lanaras2015hyperspectral,dian2017hyperspectral,qu2018unsupervised}, which describes the relationship between HSI and MSI bases. 

With $\mathbf{D}_h\in\mathbb{R}^{c \times L}$ carrying the high-resolution spectral information and $\mathbf{S}_m \in\mathbb{R}^{MN\times c}$ carrying the high-resolution spatial information, the desired HR HSI, $\mathbf{X}$, is generated by Eq. \eqref{equ:bhrhsi}. 

The challenges to solve this problem are that 1) the ground truth $\mathbf{X}$ is not available, and 2) the LR HSI and HR MSI do not cover the same region. To solve this unsupervised and unregistered HR-HSI problem, the key is to take advantage of the shared spectral information $\mathbf{D}_h$ among different modalities. In addition, the representations of both modalities specifying the spatial information of scene should meet the non-negative and sum-to-one physical constraints. Moreover, in the ideal case, for the pixels in the overlapped region between LR HSI and HR MSI, their spatial information should follow similar patterns, because they carry the information of how the reflectance of shared materials (spectral basis) are mixed in each location. Therefore, the network should have the ability to learn correlated spatial and spectral information from unregistered multi-modality images to maximize its ability to prevent spectral distortion.

\section{Proposed Approach}
\begin{figure}
\setlength{\abovecaptionskip}{0.cm}
\setlength{\belowcaptionskip}{0.cm}
	{\includegraphics[width=1\linewidth]{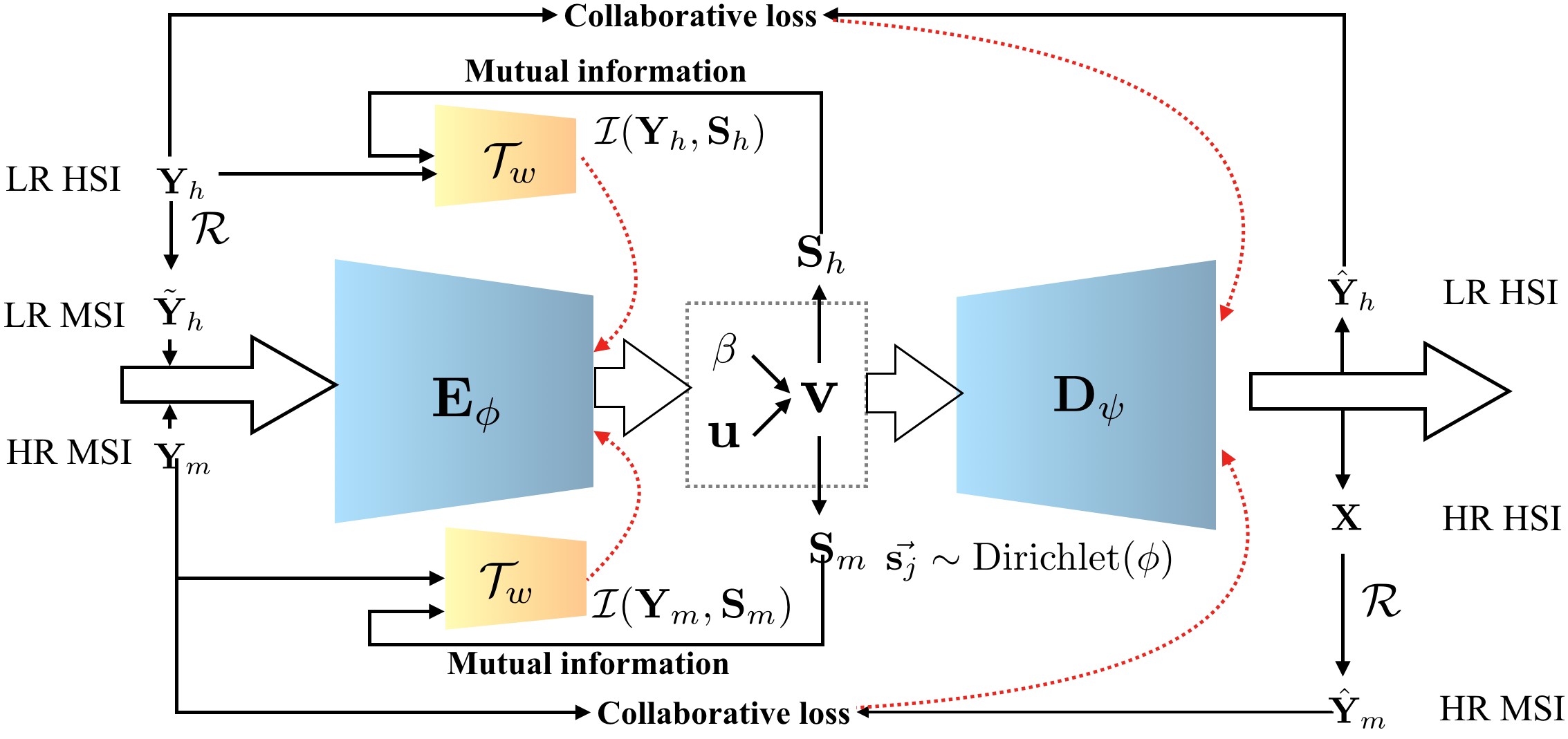}}
	\caption{Simplified architecture of $u^2$-MDN.}
	\label{fig:flow}
\end{figure}

We propose an unsupervised architecture for unregistered LR HSI and HR MSI as shown in Fig.~\ref{fig:flow}. Here, we highlight the structural uniqueness of the network. To extract correlated spectral and spatial information of unregistered multi-modalities, the network projects the LR HSI into the same statistical space as HR MSI, so that the two modalities can share the same encoder and decoder. The encoder enforces the representations (carrying spatial information) of both modalities to follow a Dirichlet distribution, to naturally meet the non-negative and sum-to-one physical properties.  In order to prevent spectral distortion, mutual information is introduced during optimization to maximize the correlation between the representations of LR HSI and HR MSI. And the collaborative $l_{2,1}$ loss is adopted to encourage the network to extract accurate spectral and spatial information from both modalities. 

\subsection{Network Architecture}
\label{sec:arch}
As shown in Fig.~\ref{fig:flow}, the network reconstructs both the LR HSI $\mathbf{Y}_h$ and HR MSI $\mathbf{Y}_m$ by sharing the same encoder and decoder network structure. Since the number of the spectral band $L$ of the HSI  $\mathbf{Y}_h$ is much larger than that of the spectral band $l$ of MSI $\mathbf{Y}_m$, we project $\mathbf{Y}_h$ into an $l$ dimensional space by $\tilde{\mathbf{Y}}_h = \mathbf{Y}_h\mathcal{R}$, such that $\tilde{\mathbf{Y}}_h$ represents the LR MSI lying in the same space as HR MSI. In this way, both modalities are linked to share the same encoder structure without additional parameters. 

On the other hand, the spectral information $\mathbf{D}_m$ of MSI is highly compressed from that of HSI, \ie, $\mathbf{D}_m =\mathbf{D}_h\mathcal{R}$. Thus, it is very unstable and difficult to directly extract $\mathbf{D}_h$, carrying high spectral resolution from, MSI with low-spectral resolution. But the spectral basis of HR MSI can be transformed from those of LR HSI which possesses more spectral information, \ie, $\hat{\mathbf{Y}}_m = \mathbf{S}_m \mathbf{D}_m = \mathbf{S}_m \mathbf{D}_h \mathcal{R} = \mathbf{X} \mathcal{R}$. Therefore, in the network design, both modalities share the same decoder structure $\mathbf{D}_h$. The transformation matrix $\mathcal{R}$ is added as fixed weights to reconstruct the HR MSI $\hat{\mathbf{Y}}_m$. Then the output of the layer before the fixed weights is actually $\mathbf{X}$, according to Eq.~\eqref{equ:bhrhsi}. 

Let us define the input domain as $\mathcal{Y} = \{\tilde{\mathbf{Y}}_h, \mathbf{Y}_m\}$, output domain as $\hat{\mathcal{Y}} = \{\hat{\mathbf{Y}}_h, \mathbf{X}\}$, and the representation domain as $\mathcal{S} = \{\mathbf{S}_h, \mathbf{S}_m\}$, the encoder of the network $\text{E}_{\phi}:\mathcal{Y}\rightarrow\mathcal{S}$, maps the input data to low-dimensional representations (latent variables on the Bottleneck hidden layer), \ie, $p_{\phi}(\mathcal{S}\vert {\mathcal{Y}})$ and the decoder  $\text{D}_{\psi}:\mathcal{S}\rightarrow\hat{\mathcal{Y}}$ reconstructs the data from the representations, \ie, $p_{\psi}(\hat{\mathcal{Y}} \vert \mathcal{S})$. Note that the bottleneck hidden layer $\mathcal{S}$ behaves as the representation layer that reflects the spatial information, and the weights $\psi$ of the decoder $\text{D}_{\psi}$ serve as $\mathbf{D}_h$ in Eq.~\eqref{equ:blrhsi}, respectively. This correspondence is further elaborated below. 

Taking the procedure of training LR HSI as an example. The LR HSI is reconstructed by $\hat{\mathbf{Y}}_h = \mathbf{D}_{\psi}(\mathbf{S}_h)$, where $\mathbf{S}_h = \mathbf{E}_{\phi}(\mathbf{Y}_h)$. Since $\mathbf{Y}_h$ carries the high-resolution spectral information, to better extract the spectral basis, part of the network should simulate the prior relationship described in Eq.~\eqref{equ:blrhsi}. That is, the representation layer $\mathbf{S}_h$ acts as the proportional coefficients and the weights $\psi$ of the decoder correspond to the spectral basis $\mathbf{D}_h$ in Eq.~\eqref{equ:blrhsi}. Therefore, in the network structure, we define $\psi = \mathbf{W}_1\mathbf{W}_2...\mathbf{W}_k = \mathbf{D}_h$ with identity activation function without bias, where $\mathbf{W}_k$ denotes the weights in the $k$th layer. In this way,  $\mathbf{D}_h$  preserves the spectral information of LR HSI, and the latent variables $\mathbf{S}_h$ preserves the spatial information effectively. More implementation details will be described in Sec.~\ref{sec:diri}. 

Eventually, the desired HR HSI is generated directly by $\mathbf{X} = \mathbf{S}_m\mathbf{D}_h$.  Note that the dashed lines in Fig.~\ref{fig:flow} show the path of back-propagation which will be elaborated in Sec.~\ref{sec:opt}. 

\subsection{Mutual Dirichlet Network with Collaborative Constraint}
\label{sec:diri}
\begin{figure}
\setlength{\abovecaptionskip}{0.cm}
\setlength{\belowcaptionskip}{0.cm}
	\begin{center}
	{\includegraphics[width=0.8\linewidth]{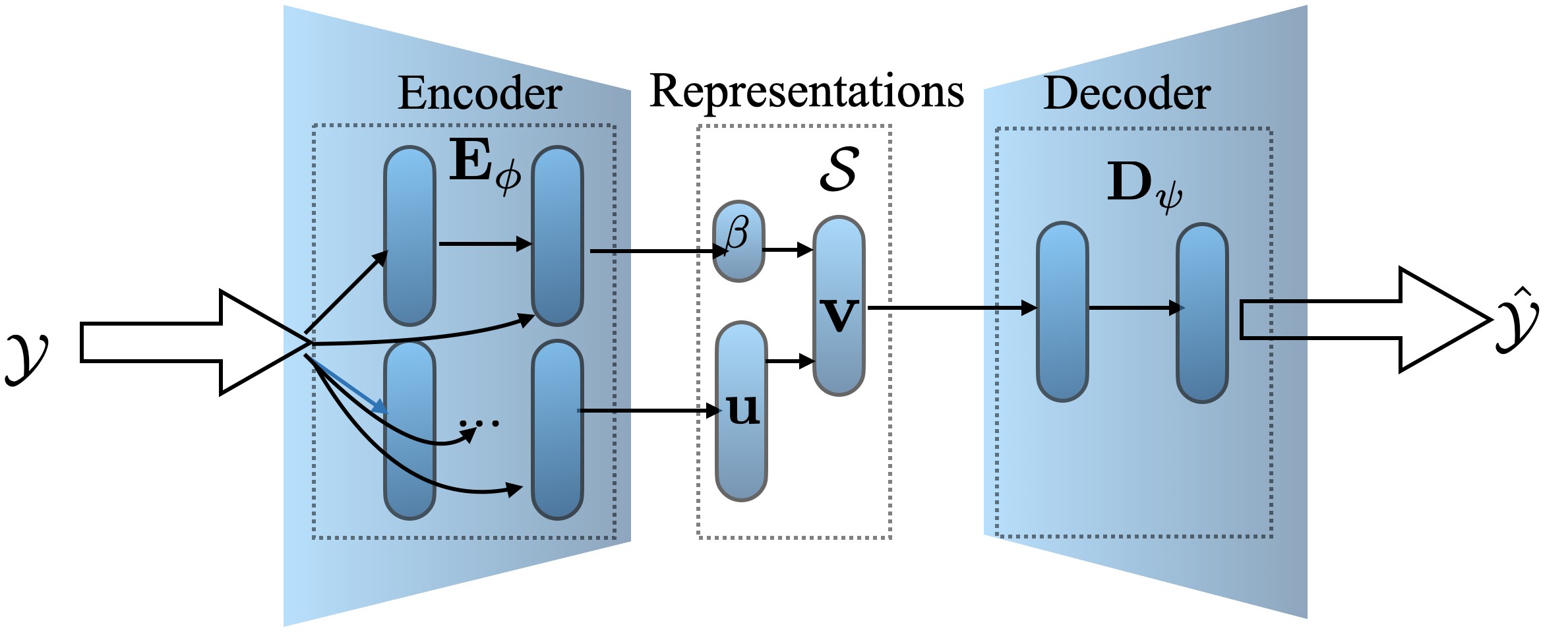}}
	\caption{Details of the encoder-decoder structure.}
	\label{fig:flow1}
	\end{center}
\end{figure}

To  extract better spectral information and naturally incorporate the physical requirements of spatial information, \ie, non-negative and sum-to-one, the representations $\mathcal{S}$ are encouraged to follow a Dirichlet distribution. In addition, the network should have the ability to learn the correlated and optimized representations generated from the encoder $\mathbf{E}_{\phi}$ for both modalities. Thus, in the network design, we maximize the mutual information (MI) between the representations of LR HSI, $\mathbf{S}_h$, and HR MSI ,$\mathbf{S}_m$, by maximizing the MI between the input images and their own representations. To further reduce the spectral distortion, the collaborative $l_{2,1}$ loss is incorporated into the network instead of the traditional $l_2$ reconstruction loss. The detailed encoder-decoder structure and the MI structure are shown in Fig.~\ref{fig:flow1} and Fig.~\ref{fig:flow2}, respectively. 

\subsubsection{Dirichlet Structure}
To generate representations with Dirichlet distribution, we incorporate the stick-breaking structure between the encoder and representation layers. The stick-breaking process was first proposed by Sethuranman~\cite{sethuraman1994constructive} back in 1994. It can be illustrated as breaking a unit-length stick into $c$ pieces, the length of which follows a Dirichlet distribution. Nalisnick and Smyth, and Qu \etal successfully coupled the expressiveness of networks with the Bayesian nonparametric model through a stick-breaking process~\cite{nalisnick2016deep,qu2018unsupervised}. Here, we follow the work of \cite{nalisnick2016deep,qu2018unsupervised}, which draw the samples of $\mathcal{S}$ from Kumaraswamy distribution \cite{kumaraswamy1980generalized}. 

The stick-breaking process is integrated into the network between the encoder $\mathbf{E}_{\phi}$ and the decoder $\mathbf{D}_{\psi}$, as shown in Fig.~\ref{fig:flow}. Assuming that the generated representation row vector is denoted as $\mathbf{s}_i = \{s_{ij}\}_{1 \leq j \leq c}$, we have $0\leq s_{ij}\leq1$, and $\sum_{j=1}^{c}{s_{ij}}=1$.  Each variable $s_{ij}$ can be defined as

\begin{equation}
s_{ij} =\left\{
\begin{array}{ll}
v_{i1} \quad & \text{for} \quad j = 1\\
v_{ij}\prod_{k<j}(1-v_{ik}) \quad &\text{for} \quad j>1 , 
\end{array}\right.
\label{equ:stick}
\end{equation}
where  $v_{ik}\sim \text{Beta}(u, \alpha,\beta)$.  Since it is difficult to draw samples directly from the Beta distribution, we draw samples from the inverse transform of Kumaraswamy distribution. The benefit of the Kumaraswamy distribution is that it has a closed-form CDF, and it is equivalent to the Beta distribution when $\alpha = 1$ or $\beta = 1$. Let $\alpha=1$, we have
\begin{equation}
v_{ik}\sim 1-(1-u_{ik}^\frac{1}{\beta_{i}}).
\label{equ:draw}
\end{equation}
Both parameters $u_{ik}$ and $\beta_{i}$ are learned through the network for each row vector as illustrated in Fig.~\ref{fig:flow}.  Because $\beta>0$, a softplus is adopted as the activation function \cite{dugas2001incorporating} at the ${\beta}$ layer. Similarly, a sigmoid \cite{han1995influence} is used to map ${u}$ into $(0,1)$ range at the $\mathbf{u}$ layer. Due to the fact that the spectral signatures of data are different for each image pair, the network only trains one group of data, \ie, LR HSI $\mathbf{Y}_h$ and HR MSI $\mathbf{Y}_m$,  to reconstruct its own HR HSI $\mathbf{X}$. Therefore, to increase the representation power of the network, the encoder of the network is densely connected, \ie, each layer is fully connected with all its subsequent layers \cite{huang2016densely}.

\subsubsection{Mutual Dirichlet Network}
Before further describing the details of the network, we first explain the reason that motivates this design. Given unregistered multi-modalities LR HSI, $\mathbf{Y}_h$ and HR MSI, $\mathbf{Y}_m$, and the desired HR HSI, $\mathbf{X}$, each pixel of which indicates the mixed spectral reflection of the captured area. The overlapped region of the three modalities is defined by $\mathcal{C}$. Ideally, each pixel in the overlapped region of these three modalities should possess the same spectral signatures. In addition, the corresponding proportional coefficients of $\mathbf{X}$ and $\mathbf{Y}_m$ should be the same for a given pixel within $\mathcal{C}$. Since $\mathbf{Y}_h$ is a down-sampling and transformed version of $\mathbf{X}$, its proportional coefficients (representations) should follow the same pattern as that of $\mathbf{X}$ and $\mathbf{Y}_m$, \ie, $\mathbf{S}_h$ and $\mathbf{S}_m$ should be highly correlated although with different resolution. One example is shown in Fig.~\ref{fig:sample}. Therefore, to generate HR HSI with low spectral distortion, it is necessary to encourage the representations $\mathbf{S}_h$ and $\mathbf{S}_m$ to follow similar patterns. However, traditional constraints like correlation may not work properly, because the input LR HSI and HR MSI are not registered with each other and the mapping function $\mathbf{E}_{\phi}$, between the input $\mathcal{Y}$ and the representations $\mathcal{S}$, holds the non-linear property. Therefore, we introduce MI, which captures the non-linear statistical dependencies between variables~\cite{kinney2014equitability}, to reinforce the representations of LR HSI and HR MSI to follow similar patterns with statistics. 

Mutual information has been widely used for multi-modality registrations~\cite{zitova2003image,woo2015multimodal}. It is a Shannon-entropy-based measurement of mutual independence between two random variables, \eg, $\mathbf{S}_h$ and $\mathbf{S}_m$. The mutual information $\mathcal{I}( \mathbf{S}_h; \mathbf{S}_m)$ measures how much uncertainty of one variable  ($\mathbf{S}_h$ or  $\mathbf{S}_m$) is reduced given the other variable ($\mathbf{S}_m$ or $\mathbf{S}_h$). Mathematically, it is defined as 

\begin{equation}
\begin{array}{ll}
\mathcal{I}(\mathbf{S}_h; \mathbf{S}_m) &= H(\mathbf{S}_h) - H(\mathbf{S}_h \vert \mathbf{S}_m)\\
&= \int_{\mathcal{S}_h\times\mathcal{S}_m}\log\frac{d\mathbb{P}_{\mathbf{S}_h\mathbf{S}_m}}
{d\mathbb{P}_{\mathbf{S}_h}\otimes d\mathbb{P}_{\mathbf{S}_m}}d\mathbb{P}_{\mathbf{S}_h\mathbf{S}_m}
\end{array}
\end{equation}
where $H$ indicates the Shannon entropy, $H(\mathbf{S}_h\vert \mathbf{S}_m)$ is the conditional entropy of  $\mathbf{S}_h$ given $\mathbf{S}_m $. $d\mathbb{P}_{\mathbf{S}_h\mathbf{S}_m}$ is the joint probability distribution, and $\mathbb{P}_{\mathbf{S}_h}$, $\mathbb{P}_{\mathbf{S}_m}$ denote the marginals. Belghazi \etal ~\cite{belghazi2018mine} introduced an MI estimator, which allows neural network to estimate MI through back-propagation, by adopting the concept of Donsker-Varadhan representation~\cite{donsker1983asymptotic}.

In order to maximally preserve the spectral information of the reconstructed HR HSI, our goal is to encourage the two representations $\mathbf{S}_h$ and $\mathbf{S}_m$ to follow similar patterns by maximizing their MI, $\mathcal{I}(\mathbf{S}_h; \mathbf{S}_m)$, during the optimization procedure. Since $\mathbf{S}_h = \mathbf{E}_{\phi}(\mathbf{Y}_h)$ and $\mathbf{S}_m = \mathbf{E}_{\phi}(\mathbf{Y}_m)$, the MI can also be expressed as $\mathcal{I}(\mathbf{E}_{\phi}(\mathbf{Y}_h); \mathbf{E}_{\phi}(\mathbf{Y}_m))$. However, it is difficult to maximize such MI directly with neural networks, because the two modalities do not match with each other in our scenario. Therefore, we maximize the average MI between the representations and their own inputs, \ie, $\mathcal{I}(\mathbf{Y}_h, \mathbf{E}_{\phi}(\mathbf{Y}_h))$ and $\mathcal{I}(\mathbf{Y}_m, \mathbf{E}_{\phi}(\mathbf{Y}_m))$. The benefit of doing this is two-fold. First, by optimizing the encoder weights $\mathbf{E}_{\phi}$, it is able to greatly improve the quality of individual representations~\cite{hjelm2018learning}. Thus it helps the network to preserve the spectral and spatial information better. Second, since the multi-modalities, \ie,$\mathbf{Y}_h$ and $\mathbf{Y}_m$, are correlated, and the dependencies (MI) between the representations and multi-modalities are maximized, it also maximizes the MI, $\mathcal{I}(\mathbf{S}_h; \mathbf{S}_m)$, between different modalities, such that $\mathbf{S}_h$ and $\mathbf{S}_m$ are encouraged to follow similar patterns. Let's explain it with a toy example. We assume that both $\mathbf{Y}_h$ and $\mathbf{Y}_m$ cover the same material `brick', the spectral pixel of which in the image pairs are denoted by $\mathbf{y}_h$ and $\mathbf{y}_m$, respectively, and $\tilde{\mathbf{y}}_h=\mathbf{y}_h\mathcal{R}$. $\tilde{\mathbf{y}}_h$, and $\mathbf{y}_m$ may not be identical to each other in real applications, but they are correlated and should possess similar spectral information. By maximizing the MI between the image and their representations, we are able to find a better representation $\mathbf{s}_h$ which reduces the uncertainty of $\tilde{\mathbf{y}}_h$ to a large extent, and also a better representation $\mathbf{s}_m$, which reduces the uncertainty of $\tilde{\mathbf{y}}_m$ to a large extent. Since $\tilde{\mathbf{y}}_h$ and $\mathbf{y}_m$ are similar, $\mathbf{s}_m$ and $\mathbf{s}_h$ should also be similar. In this way, the MI can regularize the solution space, such that $\mathbf{S}_h$ and $\mathbf{S}_m$ have similar patterns.

Taking  $\mathcal{I}(\mathbf{Y}_h, \mathbf{E}_{\phi}(\mathbf{Y}_h))$ as an example. It is equivalent to Kullback-Leibler (KL) divergence~\cite{belghazi2018mine} between the joint distribution $\mathbb{P}_{\mathbf{Y}_h\mathbf{E}_{\phi}(\mathbf{Y}_h)}$ and the product of the marginals $\mathbb{P}_{\mathbf{Y}_h}\otimes \mathbb{P}_{\mathbf{E}_{\phi}(\mathbf{Y}_h)}$. Let $\mathbb{P} = \mathbb{P}_{\mathbf{Y}_h\mathbf{E}_{\phi}(\mathbf{Y}_h)}$ 
and $\mathbb{Q} = \mathbb{P}_{\mathbf{Y}_h}\otimes \mathbb{P}_{\mathbf{E}_{\phi}(\mathbf{Y}_h)}$, we can further express MI  as 
\begin{equation}
\mathcal{I}(\mathbf{Y}_h, \mathbf{E}_{\phi}(\mathbf{Y}_h)) = \mathbb{E}_{\mathbb{P}}
[\log\frac{d\mathbb{P}}{d\mathbb{Q}}] = D_{KL}(\mathbb{P}\Vert\mathbb{Q})
%D_{KL}(\mathbb{P}_{\mathbf{Y}_h\mathbf{E}_{\phi}(\mathbf{Y}_h)} \Vert \mathbb{P}_{\mathbf{Y}_h}\otimes \mathbb{P}_{\mathbf{E}_{\phi}(\mathbf{Y}_h)}),
\label{equ:kl}
\end{equation}
Such MI can be maximized by maximizing the KL-divergence's lower bound based on Donsker-Varadhan (DV) representation~\cite{donsker1983asymptotic}. Since we do not need to calculate the exact MI, we introduce an alternative lower bound based on Jensen-Shannon which works better than the DV-based objective function~\cite{hjelm2018learning}. 

\begin{figure}
\setlength{\abovecaptionskip}{0.cm}
\setlength{\belowcaptionskip}{0.cm}
	\begin{center}
	{\includegraphics[width=0.6\linewidth]{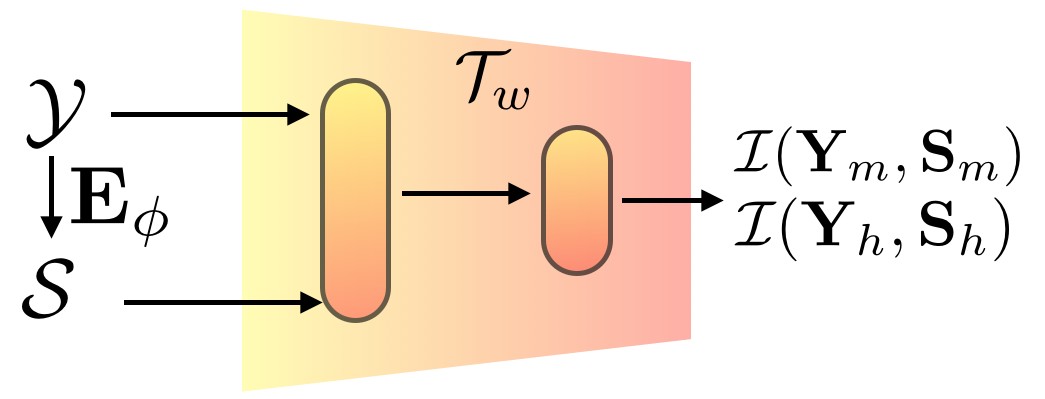}}
	\caption{Details of the MI structure.}
	\label{fig:flow2}
	\end{center}
\end{figure}

In the network design, an additional network $\mathcal{T}_w:\mathcal{Y}\times\mathcal{S}\rightarrow\mathbb{R}$ is built with two fully-connected layers, whose weights are denoted as $w$. During the training procedure, the raw image and the extracted representations are stacked and fed into the network as shown in Fig.~\ref{fig:flow2}. Then the estimator can be defined as
\begin{equation}
\begin{array}{ll}
\mathcal{I}_{\phi,w}(\mathbf{Y}_h, \mathbf{E}_{\phi}(\mathbf{Y}_h)) :&=
\mathbb{E}_{\mathbb{P}}[-sp(-\mathcal{T}_{w,\phi}(\mathbf{Y}_h,\mathbf{E}_{\phi}(\mathbf{Y}_h))]
\end{array}
\label{equ:js}
\end{equation}
%&- \mathbb{E}_{\mathbb{P}\times\tilde{\mathbb{P}}}[sp(\mathcal{T}_{w,\phi}(\mathbf{Y}_h',\mathbf{E}_{\phi}(\mathbf{Y}_h))]\\
%where $sp(x) =\log(1+e^x) $ and $\mathbf{Y}_h'$ is an input sampled from $\tilde{\mathbb{P}}=\mathbb{P}$ by shuffling the input data. Since both $\mathbf{E}_\phi$ and $\mathcal{T}_w$ are used to find the optimal representations, they are updated together. Combined with the MSI MI, the objective function is defined as
where $sp(x) =\log(1+e^x) $. Note that we ignore the negative samples in DV-based objective function~\cite{hjelm2018learning}, which are usually generated by shuffling the input data. Because it is unstable to train the network with random shifting input data given only two input data pairs. Since both $\mathbf{E}_\phi$ and $\mathcal{T}_w$ are used to find the optimal representations, they are updated together. Combined with the MSI MI, the objective function is defined as
\begin{equation}
\label{optmiall}
\begin{array}{ll}
\mathcal{L}_{\mathcal{I}}(\phi,w) &= \mathcal{I}_{\phi,w}(\mathbf{Y}_h, \mathbf{E}_{\phi}(\mathbf{Y}_h))\\
&+ \mathcal{I}_{\phi,w}(\mathbf{Y}_m, \mathbf{E}_{\phi}(\mathbf{Y}_m))
\end{array}
\end{equation} 
Since the encoder $\mathbf{E}_{\phi}$ and the estimation network of MI $\mathcal{T}_w$ for both LR HSI  and HR MSI  share the same weights $\phi$ and $w$, their optimized representations follow similar patterns. More optimization details are described in Sec.~\ref{sec:opt}.

In order to extract better spectral information, we adopt the collaborative reconstruction loss with $l_{2,1}$ norm \cite{nie2010efficient} instead of traditional $l_2$ norm for both LR HSI and HR MSI. The objective function for $l_{2,1}$ loss is defined as 
\begin{equation}
\begin{array}{ll}
\mathcal{L}_{2,1}(\phi,\psi)  &=  \Vert D_\psi(E_\phi(\mathbf{Y}_h))- \mathbf{Y}_h\Vert_{2,1}\\
&+\Vert D_\psi(E_\phi(\mathbf{Y}_m))- \mathbf{Y}_m\Vert_{2,1}
\end{array}
\end{equation}
where $\Vert X \Vert_{2,1} = \sum_{i=1}^{m}\sqrt{\sum_{j=1}^{n}X_{i,j}^2}$. $l_{2,1}$ norm can be treated as the sequential application of the $l_2$ norm on each pixel vector, followed by the $l_1$ norm on the image to enforce the reconstruction errors of the entire image to be sparse, that is, most of the reconstruction errors of individual pixels to be zero, such that the individual pixels would be reconstructed as accurately as possible.
% $l_{2,1}$ norm will encourage the rows of the reconstruction error to be sparse. That is, the network is designed to learn individual pixels as accurate as possible. 
In this way, it extracts better spectral information and further reduces the spectral distortion.

\subsection{Optimization and Implementation Details}
\label{sec:opt}
The objective functions of the proposed network architecture can then be expressed as:
\begin{equation}
\label{equ:optall}
\mathcal{L}(\phi,\psi,w) = \mathcal{L}_{2,1}(\phi,\psi) - \lambda\mathcal{L}_{\mathcal{I}}(\phi,w) + \mu\Vert\psi\Vert_F^2
\end{equation}
where $l_2$ norm is applied on the decoder weights $\psi$ to prevent over-fitting. $\lambda$ and $\mu$ are the parameters that balance the trade-off between reconstruction error, negative of mutual information and weight loss, respectively. 

Before feeding into the network, the spectral vectors in LR HSI and HR MSI are transformed to zero-mean vectors by reducing the vector mean of their own image. Since the spectral information of MSI has been compressed too much (\eg, HSI has 31 bands, but MSI has 3 bands), the decoder of the network is only updated by LR HSI data to stabilize the network. The number of the input nodes is equal to the band number of HR MSI $l$. LR HSI $\mathbf{Y}_h$ is projected into a $l$ dimensional space by $\tilde{\mathbf{Y}}_h = \mathbf{Y}_h\mathcal{R}$ before feeding into the network, while HR MSI is directly fed into the network. The number of the output nodes is chosen based on the band number of LR HSI $L$. When the input of the network is ${\mathbf{Y}}_h$, the output of the decoder is $\hat{\mathbf{Y}}_h$. When the input of the network is ${\mathbf{Y}}_m$, the reconstructed $\hat{\mathbf{Y}}_m$ is generated by multiplying the output of the decoder with fixed weights $\mathcal{R}$.

The encoder-decoder is constructed with fully-connected layers and the detailed structure is shown in Fig.~\ref{fig:flow1}. The input of the encoder has $l$ neurons carrying each pixel of the image, which is densely connected by stacking with all its subsequent layers. Let's take $l=8$ as an example, the input layer has 8 neurons, and we assume that the second and the third layers have 3 neurons, respectively.  The input layer is passed to the second layer by stacking the first layer on top of the second layer. Then the stacked layer is passed to the third layer by stacking 11 neurons on top of the third layer. In this way, the encoder is densely connected. The layer $\mathbf{v}$ is drawn with Eq.~\eqref{equ:draw} given layer $\mathbf{u}$ and layer $\beta$, which are learned by back-propagation. $\beta$ has only one node, which is learned by a two-layer densely-connected fully-connected neural network. It denotes the distribution parameter of each pixel. $u$ has 15 nodes, which are learned by a four-layer densely-connected neural-network. The representation layer $\mathcal{S}$ with 15 nodes is constructed with $\mathbf{v}$ and $\mathbf{\beta}$, according to Eq.~\eqref{equ:stick}. The decoder has two fully-connected layers. The number of nodes and the activation functions for different layers are shown in Table~\ref{tab:layers}. 

\begin{table}[htb]
\setlength{\abovecaptionskip}{0.cm}
\setlength{\belowcaptionskip}{0.cm}
	\caption{The number of layers and nodes in the proposed network.}
	\label{tab:layers}
	\begin{center}
%		\begin{tabular}{c|cccc}
		  \begin{tabular}{p{0.9cm}|p{1.5cm} p{2.5cm} p{0.6cm} p{0.65cm}}
			\hline
			& $\mathbf{u}$/$\mathbf{\beta}$ encoder&$\mathbf{u}$/${\beta}$/$\mathbf{v}$&$\mathcal{T}_w$&decoder\\
			\hline
			$\#$layers &4/2&1/1/1&2&2\\
			$\#$nodes &[3,3,3,3]/[3,3]& 15/1/15&[18,1]&[15,15]\\
			activation&linear&sigmoid/softplus/linear&sigmoid&linear\\	
			\hline
		\end{tabular}
	\end{center}
\end{table}

The training is done in an unsupervised fashion without ground truth HR HSI. Given multi-modalities LR HSI and HR MSI, the network is optimized with back-propagation to extract their correlated spectral bases and representations, as illustrated in Fig.~\ref{fig:flow} with red-dashed lines. The training process stops when the reconstruction error of the network does not decrease anymore. Then we can feed the HR MSI into the trained network, and obtain the reconstructed HR HSI, $\mathbf{X}$, from the output of the decoder.

\section{Experiments and Results}
\label{sec:exp}
\subsection{Datesets}
\label{sec:exp:data}
The proposed $u^2$-MDN has been extensively evaluated with two widely used benchmark datasets, CAVE \cite{yasuma2010generalized} and Harvard \cite{chakrabarti2011statistics}, and five remote sensing datasets, Hyperspec Chikusei,  CASI University of Houston, ROSIS-3 University of Pavia, HYDICE Washington DC Mall~\cite{yokoya2017hyperspectral} and real data without simulation, as summarized in Table~\ref{tab:data}. 

\subsubsection{Cave dataset}
The CAVE dataset consists of 32 HR HSI images and each of which has a dimension of $512\times 512$ with 31 spectral bands taken within the wavelength range 400--700 nm at an interval of 10 nm. 
\subsubsection{Harvard dataset}
The Harvard dataset includes 50 HR HSI images with both indoor and outdoor scenes. The images are cropped to $1024\times 1024$, with 31 bands taken at an interval of 10 nm within the wavelength range of 420--720 nm.
\subsubsection{Hyperspec Chikusei dataset}
The dataset was taken by Headwall’s Hyperspec-VNIR-C  sensor over Chikusei, Ibaraki, Japan. The image has a ground sampling distance (GSD) of 2.5 m and was cropped to $540\times 420$ with 128 bands, covering the wavelength range from 363 to 1018 nm. Please refer to~\cite{yokoya2016airborne} for more details. 
\subsubsection{University of Houston dataset}
This dataset was acquired by ITRES CASI-1500 sensor over the University of Houston campus with a GSD of 2.5 m~\cite{debes2014hyperspectral}. It was cropped to $320\times 540$ with 144 bands taken within the wavelength range 364--1046 nm.
\subsubsection{University of Pavia dataset}
The dataset was taken by the reflective optics spectrographic imaging system (ROSIS-3) sensor over the University of Pavia, Italy, with a GSD of 1.3 m. It was cropped to $560\times 320$ with 103 spectral bands taken within the wavelength range 430--830 nm.  
\subsubsection{Washington DC Mall dataset}
The dataset was acquired by the hyperspectral digital imagery collection experiment (HYDICE) sensor over the Mall in Washington DC, USA at a GSD of 2.5 m. The image was cropped to $420\times 300$ with 191 bands covering the wavelength range from 400 to 2500 nm.
\subsubsection{Real dataset without simulation}
The LR HSI over the Cuprite mining district, Nevada, US, was acquired by Hyperion with a GSD of 30 m, the image size of which is 100$\times$153 with 167 bands taken within the wavelength range from 426 to 2355 nm. The HR MSI is the SWIR data of WorldVeiw3 with a GSD of 7.5 m, the image size of which is 460$\times$670 with 8 bands covering the wavelength range from 1209 to 2329 nm. Both rigid and nonrigid deformation exist as shown in Figs.~\ref{fig:real:a} and~\ref{fig:real:b}.
%The LR HSI was acquired by Hyperion, the resolution of which is $100\times 153$ with 167 bands taken within the wavelength range 426--2355 nm. The HR MSI is the SWIR data of WorldVeiw3, the resolution of which is $460\times 670$ with 8 bands covering the wavelength range from 1209 to 2329 nm. 

\subsection{Experimental Setup}
\label{sec:exp:setup}
For real applications, the mis-registration of two modalities is crucial for HSI-SR~\cite{baronti2011theoretical,zhou2017nonrigid,zhou2019}. To demonstrate how misregistration would influence the performance of HSI-SR, we conduct two groups of experiments to evaluate the various approaches,~\ie, the experiments on well-registered image pairs, and on unregistered image pairs. By conducting experiments in these two scenarios, we intend to show that misregistration would influence the performance of HSI-SR significantly. Therefore, it is very important to develop algorithms that can directly work on unregistered image pairs.

% However, since not all the approaches can reconstruct HR HSI from unregistered LR HSI and HR MSI. Thus, for fair comparison, 

The well-registered image pairs are generated in two different ways following the widely-used protocols for benchmark datasets~\cite{dong2016hyperspectral, lanaras2015hyperspectral, cstf2018} and the Walds protocol~\cite{wald1997fusion,yokoya2017hyperspectral} for remote sensing datasets. 
\begin{itemize}
	\item For benchmark HSI datasets, CAVE~\cite{yasuma2010generalized} and Harvard~\cite{chakrabarti2011statistics}, the image pairs are generated with the extreme Super-Resolution (SR) ratio of 32, where the LR HSI $\mathbf{Y}_h$ is obtained by averaging the HR HSI over $32\times 32$ disjoint blocks. The HR MSI with 3 bands are generated by multiplying the HR HSI with the given spectral response matrix $\mathcal{R}$ of Nikon D700~\cite{lanaras2015hyperspectral,akhtar2015bayesian,qu2018unsupervised}. Note that we adopt this setting because it is the same protocol used by state-of-the-art methods~\cite{akhtar2015bayesian, dong2016hyperspectral, lanaras2015hyperspectral, cstf2018} on general hyperspectral images. In addition, for remote sensing applications, the scale difference can even be 25~\cite{kwan2017blind} and 30~\cite{kwan2018mars}. With such settings, we are able to evaluate the proposed method in extreme scenarios. 
	\item For remote sensing datasets, the image pairs are simulated with the Walds protocol~\cite{wald1997fusion}, where the LR HSI is generated by applying a Gaussian filter with its full width at half maximum (FWHM) equal to the SR ratio, to match a plausible system modulation transfer function (MTF)~\cite{loncan2015hyperspectral,yokoya2017hyperspectral,selva2018improving}. The MSI is generated by degrading the HR HSI in the spectral domain using MSI spectral reflection functions (SRFs) from different sensors as filters. The datasets are listed in Table~\ref{tab:data}. Please refer to~\cite{yokoya2017hyperspectral} for more details.  Note that since the scales are different between the real LR HSI and HR MSI for different sensors~\cite{kwan2017blind,kwan2018mars,yokoya2017hyperspectral}, the SR ratio is set to 4, 5, 6 and 8, to evaluate the robustness of the proposed method. The noise is added to the image with a signal-to-noise-ratio (SNR) of 30 dB in all bands. 
	\end{itemize}

The unregistered image pairs are generated in the same way as that of the well-registered image pairs, except that the LR HSI images are further distorted with rigid or nonrigid deformations. 
\begin{itemize}
	\item For benchmark HSI datasets, CAVE~\cite{yasuma2010generalized} and Harvard~\cite{chakrabarti2011statistics}, it is easier to introduce rigid deformation. Thus, the LR HSI is further rotated with $5^{\circ}$ and cropped by 15\% of its surrounding pixels, \eg, for images in the CAVE dataset, 39,322 pixels of the MSI are not covered in the LR HSI; and for images in the Harvard dataset, 157,290 pixels of the MSI are not covered in the LR HSI.
	\item For remote sensing datasets, it is usually unavoidable to introduce nonrigid deformation~\cite{baronti2011theoretical}. Thus, following the protocol in~\cite{2017Block,zhou2019}, the nonrigid distortion is emulated by introducing random shifts in pixels. 	
	\item For real data, the LR HSI is directly captured from Hyperion and the HR MSI is captured from WorldView3. Both rigid and nonrigid deformations exist as shown in Figs.~\ref{fig:real:a} and~\ref{fig:real:b}.
\end{itemize}

%	In the remote sensing ﬁeld, the registration process is never completely eﬀective so that some local misregistrations remains.

%However, not all the approaches can reconstruct HR HSI from unregistered LR HSI and HR MSI. Thus, for fair comparison, we first evaluate the performance on individual well-registered image pairs with extreme super-resolution (sr) factor 32, where the LR HSI $\mathbf{Y}_h$ is obtained by averaging the HR HSI over $32\times 32$ disjoint blocks. The HR MSI with 3 bands are generated by multiplying the HR HSI with the given spectral response matrix $\mathcal{R}$ of Nikon D700~\cite{lanaras2015hyperspectral,akhtar2015bayesian,qu2018unsupervised}.

\begin{table}[htbp]
\setlength{\abovecaptionskip}{0.cm}
\setlength{\belowcaptionskip}{0.cm}
	\caption{Dataset pairs from different sensors used in the experiments.}
	\label{tab:data}
	\centering
	\begin{tabular}{l | c c c c}
		\hline
		Dataset & HSI sensor& MSI sensor& SR Ratio \\
		\hline
		CAVE & Apogee Alta U260& Nikon & 32 \\
		Harvard & Nuance FX & Nikon  & 32 \\
		Chikusei & Hyperspec & WorldView2  & 6\\
		Houston & CASI & Sentinel-2 & 5\\
		Pavia & ROSIS-3 & QuickBird  & 8\\
		Washington & HYDICE & QuickBird  & 4\\
		Real data &Hyperion & WorldView3 & 4\\
		\hline
	\end{tabular}%
\end{table}

\begin{figure}[htbp]
\setlength{\abovecaptionskip}{0.cm}
\setlength{\belowcaptionskip}{0.cm}
	\begin{center}
		\subfloat[balloon]{\includegraphics[width=0.25\linewidth]{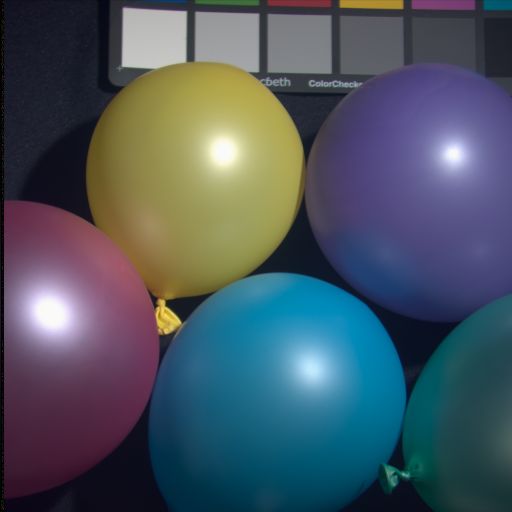}\label{fig:individual:a}}
		\subfloat[cloth]{\includegraphics[width=0.25\linewidth]{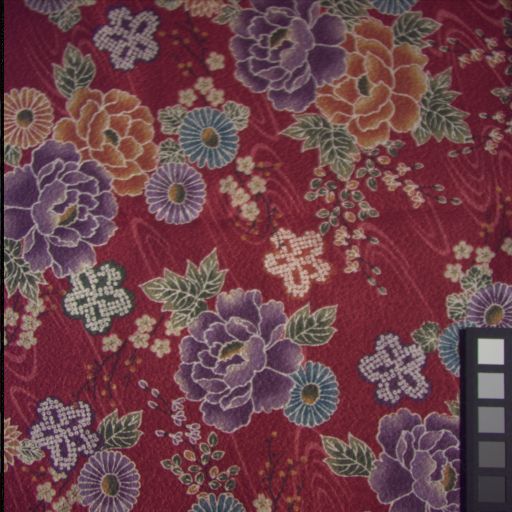}\label{fig:individual:b}}
		\subfloat[pompoms]{\includegraphics[width=0.25\linewidth]{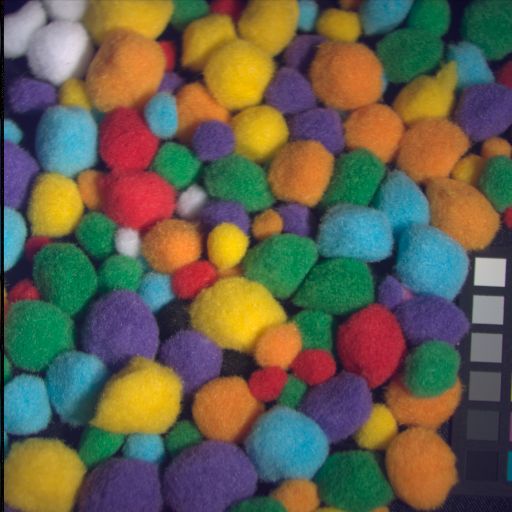}\label{fig:individual:c}}
		\subfloat[spool]{\includegraphics[width=0.25\linewidth]{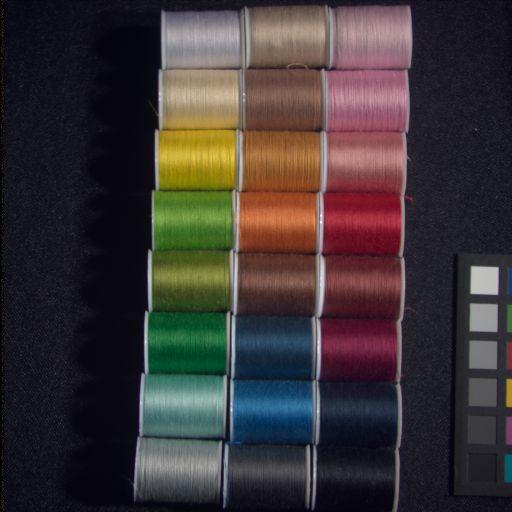}\label{fig:individual:d}}\hfill
		\subfloat[img1]{\includegraphics[width=0.3\linewidth]{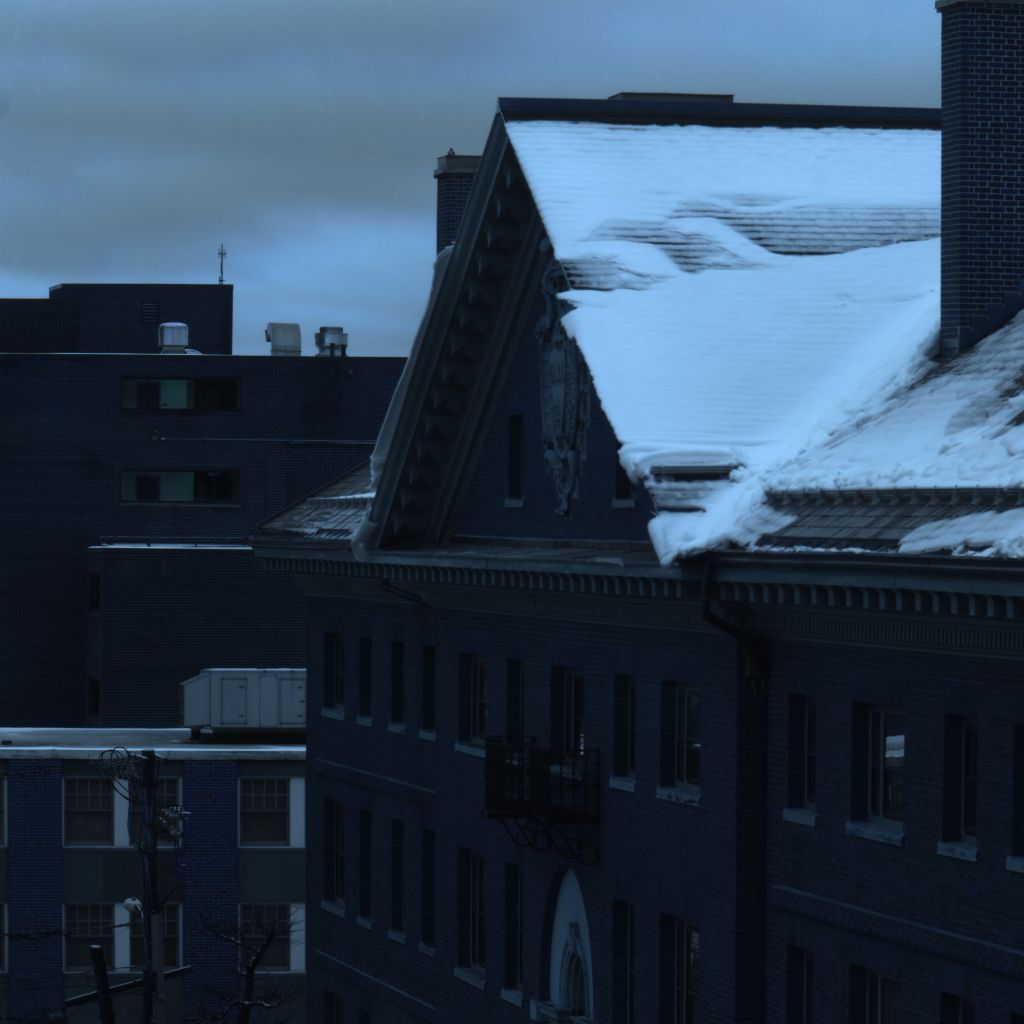}\label{fig:individual:e}}
		\subfloat[imgb5]{\includegraphics[width=0.3\linewidth]{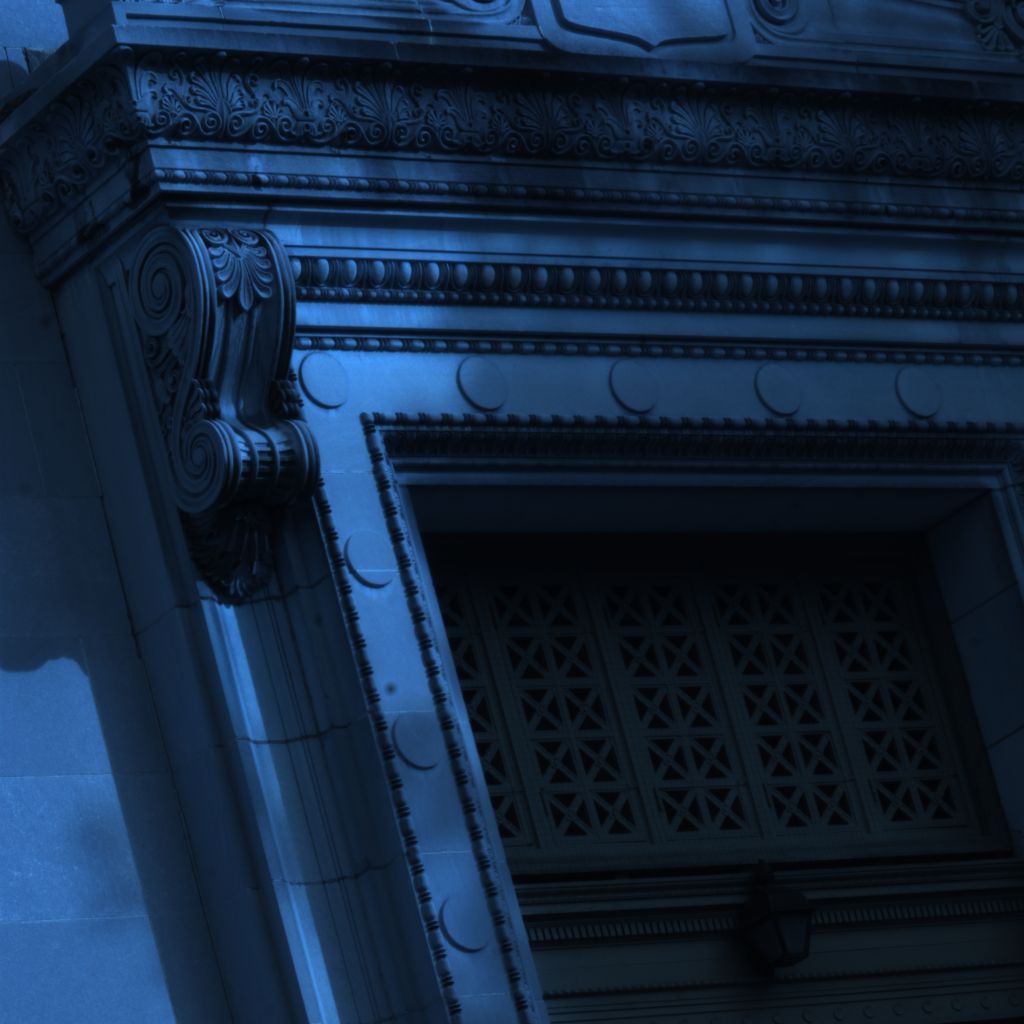}\label{fig:individual:f}}
		\subfloat[imgc5]{\includegraphics[width=0.3\linewidth]{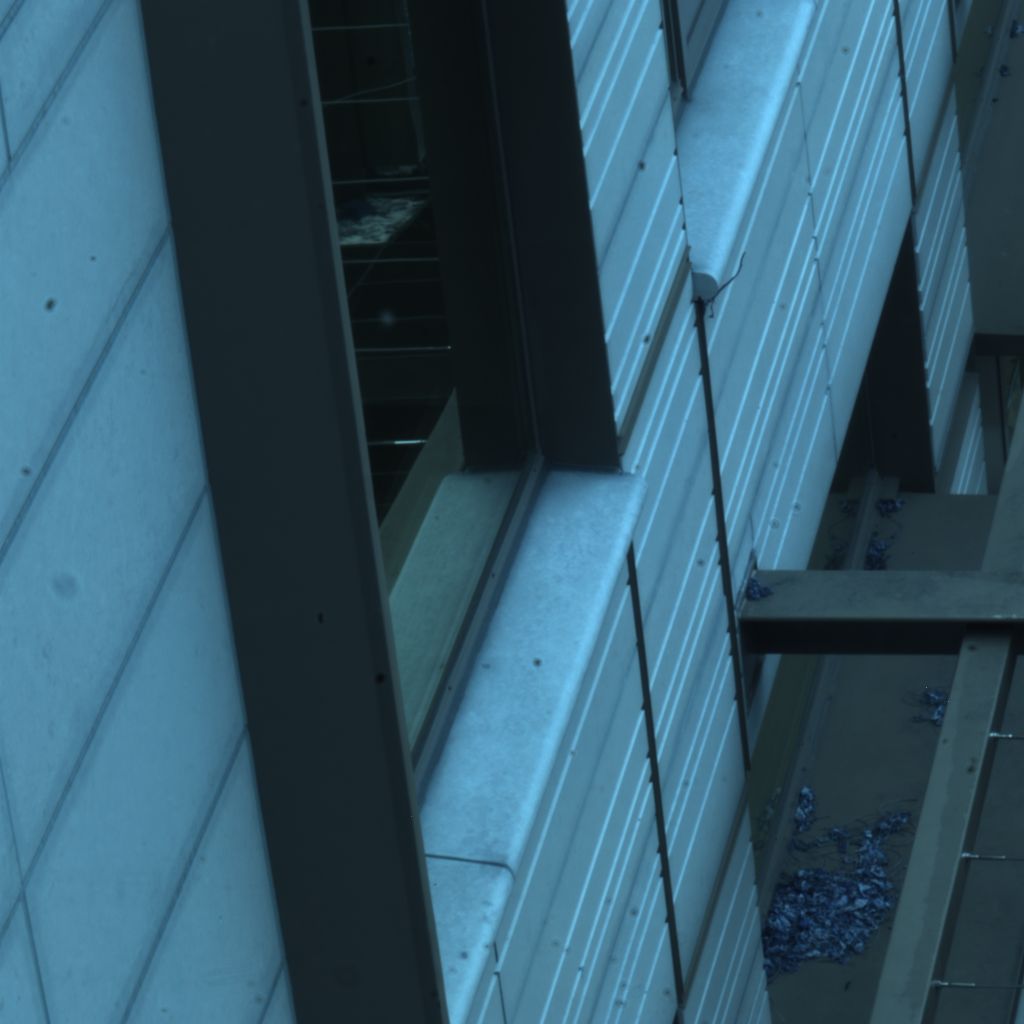}\label{fig:individual:g}}\hfill
	\end{center}
	\caption{The HR MSI of individual test images from the CAVE~\cite{yasuma2010generalized} (top row) and Harvard~\cite{chakrabarti2011statistics} (bottom row) datasets.}
	\label{fig:individual}
\end{figure}

The results of the proposed method on individual images in Fig.~\ref{fig:individual} are compared with nine state-of-the-art methods, including traditional methods such as CS-based GSA~\cite{aiazzi2007improving} and  MRA-based SFIM~\cite{liu2000smoothing}, matrix factorization based methods such as CNMF~\cite{yokoya2012coupled} and Lanaras' CSU ~\cite{lanaras2015hyperspectral}, Bayesian-based methods such as HySure~\cite{simoes2015convex}, sparse-coding based methods such as NSSR~\cite{dong2016hyperspectral}, tensor-based method~\cite{cstf2018}, the integrated registration and fusion method~\cite{zhou2019}, and the uSDN method~\cite{qu2018unsupervised} that belong to different categories of HSI-SR. These methods also reported the best performance \cite{loncan2015hyperspectral,lanaras2015hyperspectral,qu2018unsupervised}, with the original code made available by the authors. Note that the proposed $u^2$-MDN is unsupervised,~\ie, the HR HSI is not available during the training procedure. Thus, for a fair comparison, only unsupervised methods are included in the experiments. The average results on the datasets are also reported to evaluate the robustness of the proposed method.

\begin{figure}[htbp]
\setlength{\abovecaptionskip}{0.cm}
\setlength{\belowcaptionskip}{0.cm}
	\begin{center}
		\subfloat[Chikusei]{\includegraphics[width=0.35\linewidth]{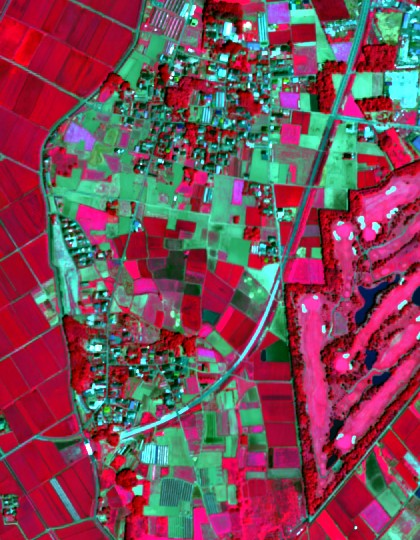}\label{fig:rs:a}}\hspace{0.3mm}
		\subfloat[Houston]{\includegraphics[width=0.45\linewidth]{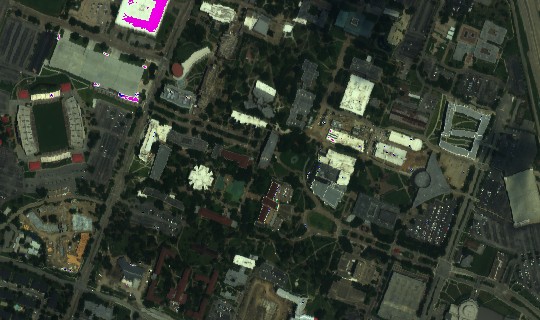}\label{fig:rs:b}}\\
		\subfloat[Pavia]{\includegraphics[width=0.35\linewidth]{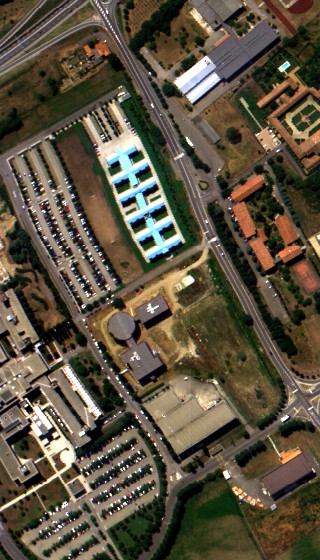}\label{fig:rs:c}}\hspace{0.3mm}
		\subfloat[Washington]{\includegraphics[width=0.35\linewidth]{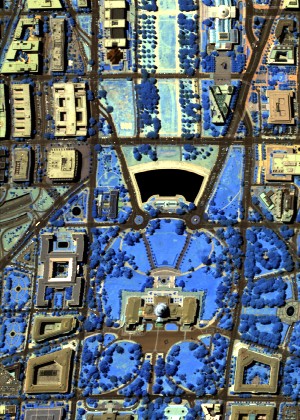}\label{fig:rs:d}}
	\end{center}
	\caption{Color composite of the remote sensing datasets from ~\cite{yokoya2017hyperspectral}. The reference HR HSI of the (a) Chikusei, (b) Houston, (c) Pavia and (d) Washington datasets.}
	\label{fig:rs}
\end{figure}

For rigid deformation, since the resolution of HSI does not match that of the degraded MSI,~\ie, there exists large displacement between two modalities, only five methods may reconstruct HR HSI from unregistered images without large errors. Thus, the proposed method is compared with these five state-of-the-art methods, \ie, GSA ~\cite{aiazzi2007improving}, SFIM~\cite{liu2000smoothing}, CNMF~\cite{yokoya2012coupled}, NSSR~\cite{dong2016hyperspectral}, and the integrated registration and fusion method~\cite{zhou2019} on unregistered image pairs. Note that, as discussed in Sec.~\ref{sec:formulate}, in order to work on unregistered image pairs, the LR HSI should include all the spectral bases of HR MSI. For the CAVE and Harvard datasets, not all the image pairs meet this requirement after rotation and cropping. Thus, we choose seven commonly used image pairs from the benchmark dataset, where the LR HSI includes all the spectral bases of HR MSI even after rotation and cropping. The chosen image pairs are shown in Fig.~\ref{fig:individual}. The remote sensing images are shown in Fig.~\ref{fig:rs}.

\subsection{Evaluation Metrics}
For quantitative comparison, the erreur relative globale adimensionnelle de synthèse (ERGAS), the peak signal-to-noise ratio (PSNR), and the spectral angle mapper (SAM) are applied to evaluate the quality of the reconstructed HSI. 

ERGAS provides a measurement of the band-wise normalized root of mean square error (RMSE) between the reference HSI, $\mathbf{X}$, and the reconstructed HSI, $\hat{\mathbf{X}}$, with the best value at 0~\cite{wald1997fusion}.  It is defined as

\begin{equation}
\text{ERGAS}(\mathbf{X},\hat{\mathbf{X}}) = \frac{100}{\text{sr}}\sqrt{{\frac{1}{L}}\sum_{i=1}^{L}\frac{\text{mean}\Vert\mathbf{X}_i-\hat{\mathbf{X}_i}\Vert_2^2}{(\text{mean}\mathbf{X}_i)^2}},
\end{equation}
where $\text{sr}$ denotes the sr factor between the HR MSI and LR HSI, $L$ denotes the number of spectral bands of the reconstructed $\hat{\mathbf{X}}$.

PSNR is the average ratio between the maximum power of the image and the power of the residual errors in all the spectral bands. A larger PSNR indicates a higher spatial quality of the reconstructed HSI. For each image band of HSI, the PSNR is defined as
\begin{equation}
\text{PSNR}(\mathbf{X}_i,\hat{\mathbf{X}}_i)=10\cdot\log_{10}\bigg(\frac{\max(\mathbf{X}_i)^2}{\text{mean}\Vert\mathbf{X}_i-\hat{\mathbf{X}}_i\Vert_2^2}\bigg)
\end{equation}

SAM~\cite{kruse1993spectral} is commonly used to quantify the spectral distortion of the reconstructed HSI. The larger the SAM, the worse the spectral distortion of the reconstructed HSI. For each HSI pixel $\hat{\mathbf{X}_j}$, the SAM is defined as
\begin{equation}
\text{SAM}(\mathbf{X}_j,\hat{\mathbf{X}_j}) = \arccos\left(\frac{\mathbf{X}_j^T\hat{\mathbf{X}_j}}{\Vert\mathbf{X}_j\Vert_2\Vert\hat{\mathbf{X}}_j\Vert_2}\right)
\end{equation}
The global SAM is estimated by averaging the SAM over all the pixels in the entire image. 

\begin{table*}[htbp]
\setlength{\abovecaptionskip}{0.cm}
\setlength{\belowcaptionskip}{0.cm}
	\centering
	\caption{Benchmarked results in terms of ERGAS (E), PSNR (P) and SAM (S) on well-registered image pairs.}
	\label{tab:bench_evaluate}
%	\begin{tabular}{c|ccc|ccc|ccc|ccc|ccc|ccc|ccc}
	\begin{tabular}{l|p{0.35cm}p{0.35cm}p{0.35cm}|p{0.35cm}p{0.35cm}p{0.35cm}|p{0.35cm}p{0.35cm}p{0.41cm}|p{0.35cm}p{0.35cm}p{0.41cm}|p{0.35cm}p{0.35cm}p{0.35cm}|p{0.35cm}p{0.35cm}p{0.35cm}|p{0.35cm}p{0.35cm}p{0.35cm}}
		\hline
		\multirow{3}{*}{Methods}&\multicolumn{12}{|c|}{CAVE}&\multicolumn{9}{|c}{Harvard} \\
		\cline{2-22}
		{}&\multicolumn{3}{|c|}{balloon}&\multicolumn{3}{|c|}{cloth}&\multicolumn{3}{|c|}{pompoms}&\multicolumn{3}{|c|}{spool}&\multicolumn{3}{|c|}{img1}&\multicolumn{3}{|c|}{imgb5}&\multicolumn{3}{|c}{imgc5}\\
		\cline{2-22}
	&E&P&S&E&P&S&E&P&S&E&P&S&E&P&S&E&P&S&E&P&S\\
		\hline
		GSA     &0.19&41.89&4.07&0.40&32.51&5.95&0.37&34.78&7.39&0.41&39.61&9.53&0.12&40.41&2.19&0.16&39.07&2.19&0.12&38.82&\textbf{1.67}\\
		\hline
		SFIM    &0.59&33.52&8.45&0.54&30.59&5.25&3.76&25.39&11.89&2.93&28.63&19.71&0.23&32.62&2.10&0.29&33.15&3.52&0.23&35.62&2.84\\
		\hline
		CNMF    &0.26&39.27&9.71&0.54&30.52&6.55&0.31&35.45&6.32&0.54&37.28&16.77&0.15&37.25&2.86&0.17&39.06&2.14&0.13&38.49&2.64\\
		CSU     &0.19&41.52&4.68&0.40&33.47&5.52&0.28&36.81&6.01&0.45&39.64&6.84&0.12&39.12&2.30&0.18&39.01&2.37&0.12&39.05&2.38\\
		\hline
		HySure  &0.34&37.08&9.92&0.53&30.22&7.13&0.52&31.68&10.97&0.55&37.47&15.54&0.18&35.82&4.27&0.34&35.52&3.45&0.19&36.75&2.34\\
		\hline
		NSSR		&0.16&43.2&3.35&0.31&33.3&4.58&0.26&37.71&5.31&0.45&39.41&6.91&0.14&39.91&2.24&0.17&39.12&2.17&0.12&38.87&1.87\\
		\hline
		CSTF		&\textbf{0.14}&\textbf{44.71}&3.97&0.39&32.51&5.25&0.27&36.72&6.09&{0.38}&\textbf{42.06}&8.61&0.21&33.73&2.77&0.25&34.98&2.46&0.22&32.48&1.96\\
		\hline
		Integrated&	0.28&37.75&2.64&1.47&21.55&8.73&0.52&30.29&5.99&1.03&30.94&6.77&0.32&29.81&2.68&0.63&26.29&2.31&0.27&30.47&1.79\\
		uSDN    &0.20&41.54&4.56&{0.35}&33.48&\textbf{4.16}&{0.25}&37.84&5.43&0.40&38.49&13.01&0.12&39.30&2.27&0.16&39.72&2.10&\textbf{0.11}&39.12&2.58\\
		$u^2$-MDN &{0.16}&43.59&\textbf{1.93}&\textbf{0.30}&\textbf{34.85}&{4.31}&\textbf{0.19}&\textbf{39.12}&\textbf{3.46}&\textbf{0.37}&{40.08}&\textbf{4.47}&\textbf{0.11}&\textbf{40.97}&\textbf{2.06}&\textbf{0.15}&\textbf{39.76}&\textbf{2.08}&\textbf{0.11}&\textbf{39.19}&{1.77}\\
		\hline
	\end{tabular}
\end{table*}

\subsection{Experimental Results on Registered Image Pairs} 
For a fair comparison, we first perform experiments on the general case when LR HSI and HR MSI are well registered. Table~\ref{tab:bench_evaluate} show the experimental results of 7 groups of commonly benchmarked images from the CAVE and Harvard datasets \cite{lanaras2015hyperspectral,akhtar2015bayesian,qu2018unsupervised, dong2016hyperspectral}. Table~\ref{tab:rs_evaluate} show the experimental results of the remote sensing images. The average results of the datasets are shown in Table~\ref{tab:average}. Note that, in order to show how the method works in different scenarios, the data are not normalized for evaluation. Since the intensities of the Harvard dataset are quite small, the ERGAS of the reconstructed images is generally smaller than those of the CAVE dataset and remote sensing dataset.

\begin{table*}[htbp]
\setlength{\abovecaptionskip}{0.cm}
\setlength{\belowcaptionskip}{0.cm}
  \centering
  \caption{Remote sensing results in terms of ERGAS, PSNR and SAM on well-registered image pairs.}
    \label{tab:rs_evaluate}%
    \begin{tabular}{l|ccc|ccc|ccc|ccc}
    \hline
     \multirow{2}{*}{Methods}&\multicolumn{3}{|c|}{Chikusei}&\multicolumn{3}{|c}{Houston}&\multicolumn{3}{|c}{Pavia}&\multicolumn{3}{|c}{Washington} \\
	\cline{2-13}
    &ERGAS&PSNR&SAM&ERGAS&PSNR&SAM&ERGAS&PSNR&SAM&ERGAS&PSNR&SAM\\
    \hline
    GSA   & 1.432 &42.1264&1.4478& \textbf{2.7859} &34.133&1.8443& 1.0661 &38.7949&3.5647& 3.518 &37.2308&2.187 \\
    \hline
    SFIM  & 1.2284 &47.4358&\textbf{0.9379}& 2.9415 &33.9958&0.9938& 0.7274 &42.9283&2.312& 3.0356 &39.2045&1.2382 \\
    \hline
    CNMF  & 1.479 &47.8427&1.1602& 2.9896 &33.1454&1.3882& 0.7712 &43.2417&2.3623& \textbf{3.0341} &39.1491&1.388 \\
    CSU   & 2.4705 &35.8506&1.9208& 3.2773 &32.3793&2.0193& 1.7283 &33.9385&3.5754& 4.2854 &34.1841&1.9706 \\
    \hline
    HySure & \textbf{1.2216} &48.7601&1.0934& 2.9619 &\textbf{34.5328}&1.7281& 0.7767 &43.2719&2.6094& 3.3232 &39.0&1.6808 \\
    \hline
    NSSR  & 2.6427 &33.5161&2.5263& 4.7663 &29.2931&5.3182& 3.7068 &28.8702&5.7786& 9.1737 &29.9297&4.0385 \\
    \hline
    CSTF  & 1.9024 &38.1548&1.7884& 4.0207 &29.5598&6.4& 1.1877 &37.3&4.0719& 22.4659 &20.4012&20.1433\\
    \hline
    Integrated & 1.3854	&43.4116&1.4104&4.0627 &28.9168&3.9936&1.1773&37.9724&3.5066&5.8183 &29.5106&4.3237\\
    uSDN & 1.7861 &42.8702&1.3035& 3.6198 &32.5059&5.698& 1.0221 &39.4535&3.1874& 6.7819 &30.1769&5.3259 \\
    Proposed & 1.4717 &\textbf{50.2839}&1.0578& {2.8659} &34.0584&\textbf{0.8865}& \textbf{0.715} &\textbf{43.8022}&\textbf{2.3053}& 3.8988 &\textbf{39.2144}&\textbf{1.2298} \\
    \hline
    \end{tabular}%
\end{table*}%

\begin{table*}[htbp]
\setlength{\abovecaptionskip}{0.cm}
\setlength{\belowcaptionskip}{0.cm}
	\centering
	\caption{The average ERGAS, PSNR and SAM scores over well-registered benchmarked and remote sensing datasets.}
	\label{tab:average}
	\begin{center}
		\begin{tabular}{l|ccc|ccc|ccc}
			\hline
			\multirow{2}{*}{Methods}&\multicolumn{3}{|c|}{CAVE}&\multicolumn{3}{|c}{Harvard}&\multicolumn{3}{|c}{Remote Sensing}\\
			\cline{2-4}\cline{5-7}\cline{8-10}
			&ERGAS&PSNR&SAM&ERGAS&PSNR&SAM&ERGAS&PSNR&SAM\\
			\hline
			GSA&0.34&37.2&6.74& 0.13&39.43&2.02&2.2005&38.0713& 2.261 \\
			\hline
			SFIM&1.96&29.53&11.33&0.25&33.8&2.82&\textbf{1.9832}& 40.8911& 1.3705\\
			\hline
			CNMF&0.41&35.63&9.84&0.15&38.27&2.55& 2.0685& 40.8447& 1.5747\\	
			CSU&0.33&37.86&5.76&0.14&39.06&2.35& 2.9404& 34.0881& 2.3715\\
			\hline
			HySure&0.49 &34.11&10.89&0.24&36.03&3.35&2.0709& 41.3912& 1.7779\\
			\hline
%			NSSR&0.66 &32.02&10& 0.41&30.55&3.85\\
			NSSR&0.30&38.41&5.04&0.14&39.3&2.09& 5.0724& 30.4023& 4.4154\\
			\hline
			CSTF&0.30&39.00&5.98&0.41&33.73&2.4& 7.3942& 31.3540& 8.1009 \\
			\hline
			Integrated&0.83 &30.13&6.03& 1.09&28.86&2.26& 2.8672& 33.3008& 3.9413\\	
			uSDN&0.30&37.84&6.79&0.13&39.38&2.32& 3.3025& 36.2516& 3.8787\\
			$u^2$-MDN &\textbf{0.26}&\textbf{39.41}&\textbf{3.54}&\textbf{0.12}&\textbf{39.97}&\textbf{1.97}& 2.2379&\textbf{41.8397}&\textbf{1.3699}\\		 
			\hline
		\end{tabular}
	\end{center}
\end{table*}

We observe that CS-based GSA~\cite{aiazzi2007improving} is stable on both the benchmarked and remote sensing datasets. However, it could not preserve the spectral information well especially on the benchmarked datasets. Matrix-factorization-based CSU \cite{lanaras2015hyperspectral} works better than CNMF \cite{yokoya2012coupled} on the benchmarked CAVE and Harvard datasets. However, its performance is worse than that of CNMF on the remote sensing dataset, whose number of spectral bands is higher than that of the benchmarked dataset. MRA-based SFIM~\cite{liu2000smoothing}, Bayesian-based HySure~\cite{simoes2015convex} and the integrated fusion approach~\cite{zhou2019} could achieve relatively good performance on the remote sensing datasets, but their performance drops significantly on the benchmarked CAVE and Harvard datasets. On the contrary, sparse-coding-based NSSR~\cite{dong2016hyperspectral} and tensor-based CSTF~\cite{cstf2018} could achieve much more competitive performance on the benchmarked datasets than on the remote sensing datasets. Note that for NSSR, the most effective step on the CAVE dataset is a post-processing step from~\cite{wycoff2013non}, which actually degrades the performance on remote sensing datasets with more numbers of spectral bands. Thus, the post-processing step is disabled on the remote sensing datasets to improve the reconstruction accuracy. The tensor-based CSTF could achieve competitive results on the CAVE dataset, which has a redundant spatial structure. However, its performance drops on the remote sensing datasets with less redundant spatial structure.

The deep-learning-based uSDN~\cite{qu2018unsupervised} preserves spectral information well on both the benchmarked and remote sensing datasets. However, it can only work on well-registered images due to its network design with angular difference regularization. Based on the average results shown in Table~\ref{tab:average}, the proposed $u^2$-MDN network powered by the mutual information and collaborative $l_{2,1}$ loss shows comparable, if not better, performance as compared to the state-of-the-art approaches in terms of ERGAS, PSNR, and SAM, and quite stable for different types of input images regardless of the number of spectral bands and SR ratios. In addition, it is very effective in preserving the spectral signature of the reconstructed HR HSI, showing much-improved performance, especially measured by SAM on the CAVE data. This further demonstrates the robustness of the proposed $u^2$-MDN.

\subsection{Experimental Results on Unregistered Image Pairs}

\begin{table*}[htbp]
\setlength{\abovecaptionskip}{0.cm}
\setlength{\belowcaptionskip}{0.cm}
	\centering
	\caption{Results on unregistered (rigid distorted) benchmarked images in terms of ERGAS, PSNR and SAM.}
	\label{tab:un_evaluate}
	\begin{tabular}{l|p{0.3cm}p{0.35cm}p{0.43cm}|p{0.3cm}p{0.35cm}p{0.43cm}|p{0.32cm}p{0.35cm}p{0.43cm}|p{0.32cm}p{0.35cm}p{0.43cm}|p{0.35cm}p{0.35cm}p{0.43cm}|p{0.35cm}p{0.35cm}p{0.35cm}|p{0.35cm}p{0.35cm}p{0.35cm}}
		\hline
		\multirow{3}{*}{Methods}&\multicolumn{12}{|c|}{CAVE}&\multicolumn{9}{|c}{Harvard} \\
		\hline
		{}&\multicolumn{3}{|c|}{balloon}&\multicolumn{3}{|c|}{cloth}&\multicolumn{3}{|c|}{pompoms}&\multicolumn{3}{|c|}{spool}&\multicolumn{3}{|c|}{img1}&\multicolumn{3}{|c|}{imgb5}&\multicolumn{3}{|c}{imgc5}\\
		\cline{2-22}
		&E&P&S&E&P&S&E&P&S&E&P&S&E&P&S&E&P&S&E&P&S\\
		\hline
		GSA&	0.82&27.71&14.35&0.76&27.59&9.75&1.18&23.54&22.13&1.07&29.92&17.16&1.65&23.60&10.66&0.40&20.07&4.90&0.69&22.18&5.32\\
		SFIM&1.51&22.47&12.69&1.01&24.74&9.65&1.82&19.50&14.89&1.88&25.30&21.02& 1.33&17.41&3.28&0.68&25.38&4.44&0.89&19.93&3.96\\
		CNMF&0.71&29.18&10.63&0.69&27.84&8.12&0.83&26.67&11.88&0.63&34.62&17.03&0.74&22.29&3.85&0.34&31.39&3.97&0.48&25.34&3.16\\
		NSSR&0.52&32.59&8.07&0.72&27.16&8.05&0.76&27.45&10.22&1.03&32.80&15.94&0.61&25.83&5.29&0.50&29.72&6.80&0.35&28.66&2.64\\
		Integrated&1.14&24.68&9.34&1.68&19.84&11.25&1.82&19.38&17.60&1.65&29.81&13.60&1.00&19.80&4.05&0.68&25.40&3.24&0.87&20.25&2.40 \\
		$u^2$-MDN&\textbf{0.30}&\textbf{38.61}&\textbf{3.48}&\textbf{0.40}&\textbf{32.89}&\textbf{6.08}&\textbf{0.37}&\textbf{33.64}&\textbf{4.87}&\textbf{0.56}&\textbf{36.25}&\textbf{6.78}&\textbf{0.13}&\textbf{39.42}&\textbf{2.32}&\textbf{0.25}&\textbf{36.90}&\textbf{2.73}&\textbf{0.14}&\textbf{36.29}&\textbf{2.26}\\	
		\hline
	\end{tabular}
\end{table*}

\begin{table*}[htbp]
  \centering
	\caption{Results on unregistered (nonrigid distorted) remote sensing images in terms of ERGAS, PSNR and SAM.}
	\label{tab:rs_unevaluate}%
    \begin{tabular}{l|ccc|ccc|ccc|ccc}
    \hline
     \multirow{2}{*}{Methods}&\multicolumn{3}{|c|}{Chikusei}&\multicolumn{3}{|c}{Houston}&\multicolumn{3}{|c}{Pavia}&\multicolumn{3}{|c}{Washington} \\
	\cline{2-13}
    &ERGAS&PSNR&SAM&ERGAS&PSNR&SAM&ERGAS&PSNR&SAM&ERGAS&PSNR&SAM\\
    \hline
    GSA   & 4.4617 &27.3028&7.6554& 5.2167 &27.6816&6.7581& 3.5149 &27.161&12.6803& 7.1542 &28.0469&7.8453 \\
    SFIM  & 6.8057 &23.8192&5.8405& 11.1495&22.5494&8.3767& 5.2757 &23.8248&9.2951& 10.7509 &23.9122&8.7243 \\
    CNMF  & 6.5318 &25.5878&4.7582& 5.6255 &28.0016&5.5382& 4.4232 &25.2803&7.9551& 25.8519 &29.4765&5.9344 \\
    NSSR  & 5.8158 &25.8475&5.4245& 7.9435 &23.7692&8.4696& 5.0728 &25.5778&8.2277& 18.4162 &22.3786&9.6114 \\
    Integrated & 8.9107 &21.4644&7.5126& 14.4247 &18.7631&10.2734& 7.4938 &20.7214&11.9717& 15.0896 &20.9486&11.0361 \\
    $u^2$-MDN & \textbf{1.5843}&\textbf{45.4919}&\textbf{1.0899}& \textbf{2.9458} &\textbf{33.6519}&\textbf{1.0497}& \textbf{0.8898} &\textbf{40.4446}&\textbf{2.4371}& \textbf{4.0777} &\textbf{38.6361}&\textbf{1.2757}\\
    \hline
    \end{tabular}%
\end{table*}%

In this section, two unregistered scenarios are studied,~\ie, rigid distorted benchmarked datasets, and nonrigid distorted remote sensing datasets, as described in Sec.~\ref{sec:exp:setup}. Note that, since the pixels in the HSI and MSI do not match with each other, the reconstruction errors are expected to be increased.

\begin{table*}[htbp]
\setlength{\abovecaptionskip}{0.cm}
\setlength{\belowcaptionskip}{0.cm}
	\centering
	\caption{The average ERGAS, PSNR and SAM scores over unregistered benchmarked and remote sensing datasets.}
	\label{tab:rs_average}
	\begin{center}
		\begin{tabular}{l|ccc|ccc|ccc}
			\hline
			\multirow{2}{*}{Methods}&\multicolumn{3}{|c|}{CAVE}&\multicolumn{3}{|c}{Harvard}&\multicolumn{3}{|c}{Remote Sensing}\\
			\cline{2-4}\cline{5-7}\cline{8-10}
			&ERGAS&PSNR&SAM &ERGAS&PSNR&SAM &ERGAS&PSNR&SAM\\
			\hline
			GSA&0.96&27.19&15.85& 0.91&21.95&6.96  &5.09 &27.5481&8.7348 \\
			SFIM&1.56&23.00&14.56 &0.97&20.91&3.89 &8.50 &23.5264&8.0592\\
			CNMF&0.72&29.58&11.92 &0.52&26.34&3.66 &10.61&27.0866&6.0465\\	
			NSSR&0.76&30&10.57 &0.49&28.07&4.91    &9.31 &24.3933&7.9333\\
			Integrated&1.57&23.43&12.95 &0.85&21.82&3.23 &11.48&20.4744& 10.1985\\	
			$u^2$-MDN &\textbf{0.41}&\textbf{35.35}&\textbf{5.30}&\textbf{0.17}&\textbf{37.54}&\textbf{2.44} &\textbf{2.3744}&\textbf{39.5561}&\textbf{1.4631}\\
			\hline
		\end{tabular}
	\end{center}
\end{table*}
%To solve the problem of unregistered hyperspectral image super-resolution, the LR HSI should contain all the spectral bases in the HR MSI. Since one pixel in the LR HSI covers 1024 pixels in the HR MSI, when the LR HSI is rotated and cropped, certain spectral information contained in HR MSI is corrupted or missing. Thus, the reconstruction error is expected to be increased. 

\begin{figure}[htbp]
\setlength{\abovecaptionskip}{0.cm}
\setlength{\belowcaptionskip}{0.cm}
	\begin{center}
		\subfloat[]{\includegraphics[width=0.45\linewidth]{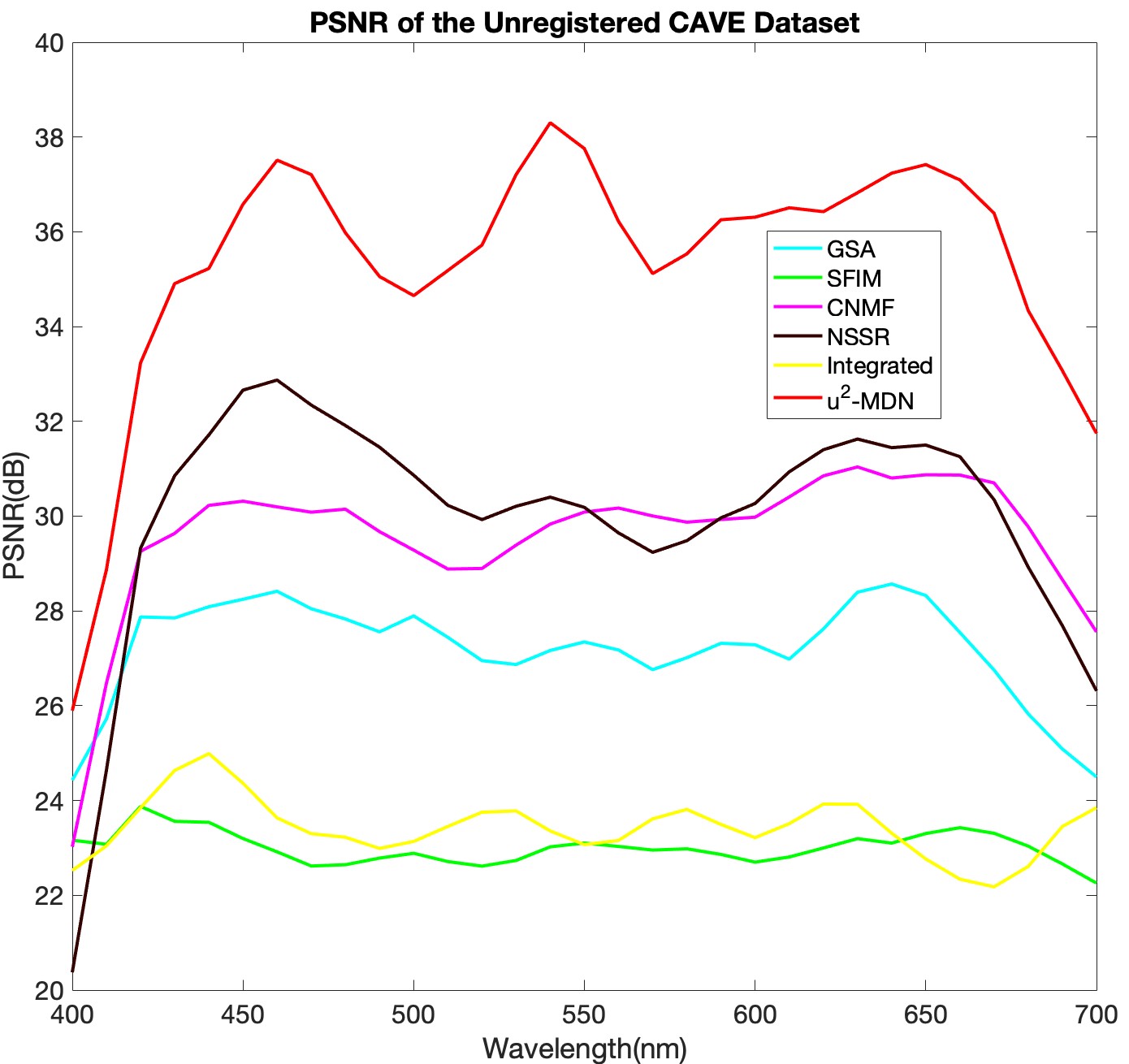}\label{fig:avg_psnr:a}}
		\subfloat[]{\includegraphics[width=0.48\linewidth]{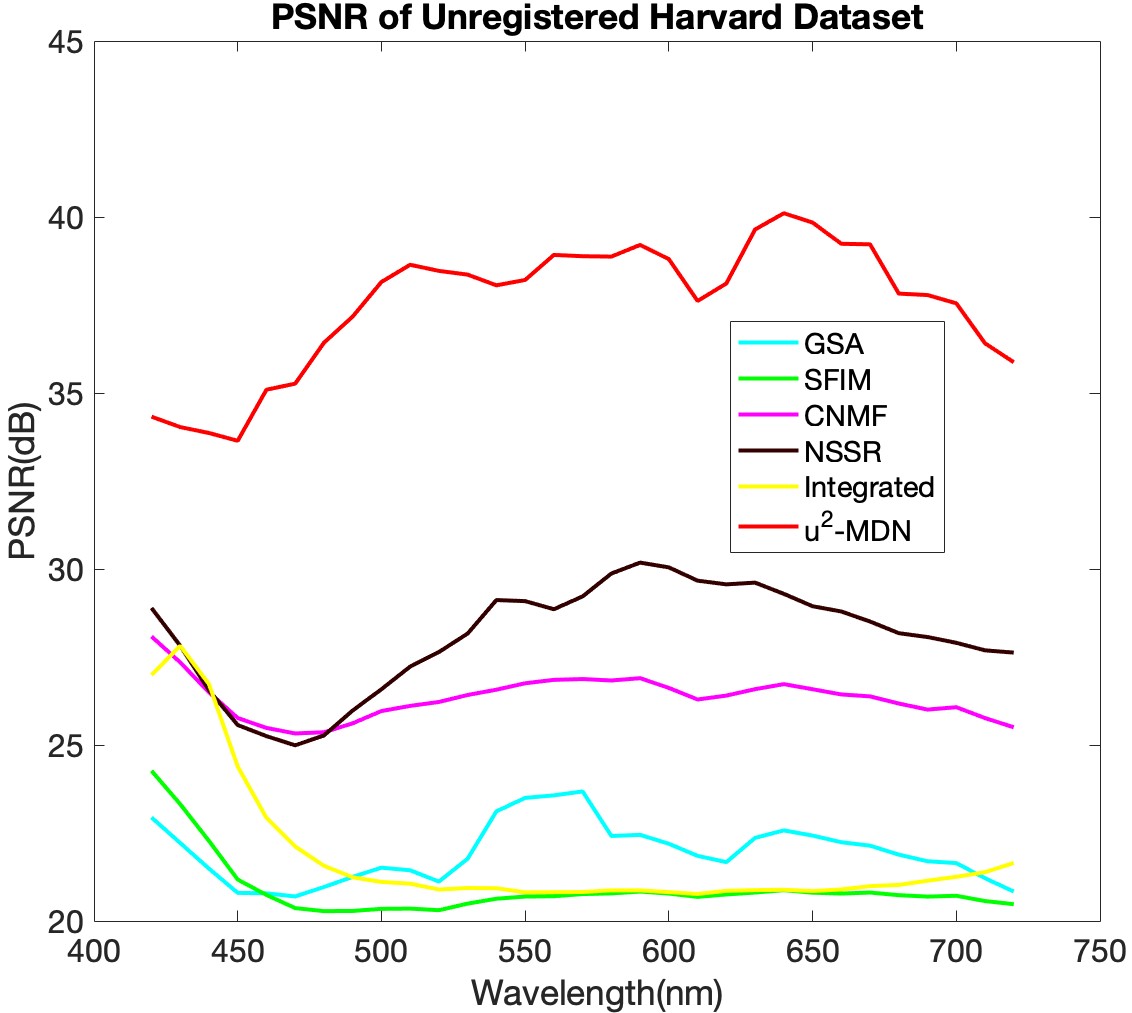}\label{fig:avg_psnr:b}}\hfill
	\end{center}
	\caption{The average PSNR of different wavelengths for the reconstructed HSI from the unregistered rigid distorted (a) CAVE dataset and (b) Harvard dataset, respectively.}
	\label{fig:avg_psnr}
\end{figure}

\begin{figure}[htbp]
\setlength{\abovecaptionskip}{0.cm}
\setlength{\belowcaptionskip}{0.cm}
	\begin{center}
		\subfloat[]{\includegraphics[width=0.5\linewidth]{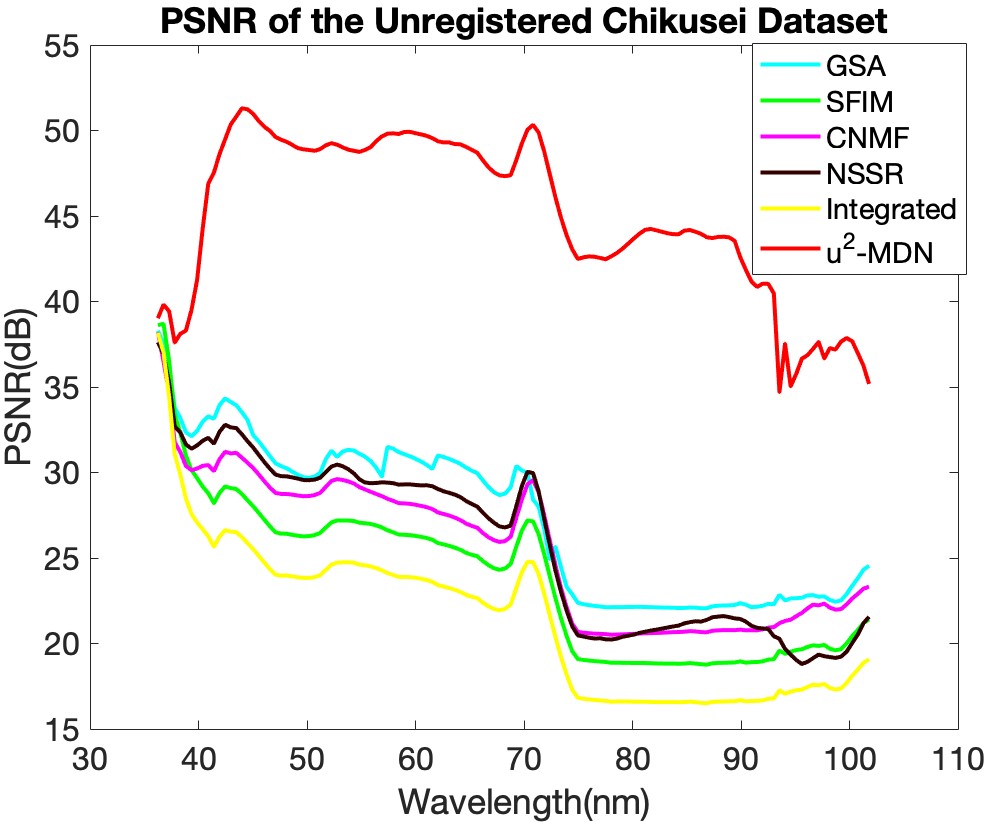}\label{fig:rs_psnr:a}}
		\subfloat[]{\includegraphics[width=0.5\linewidth]{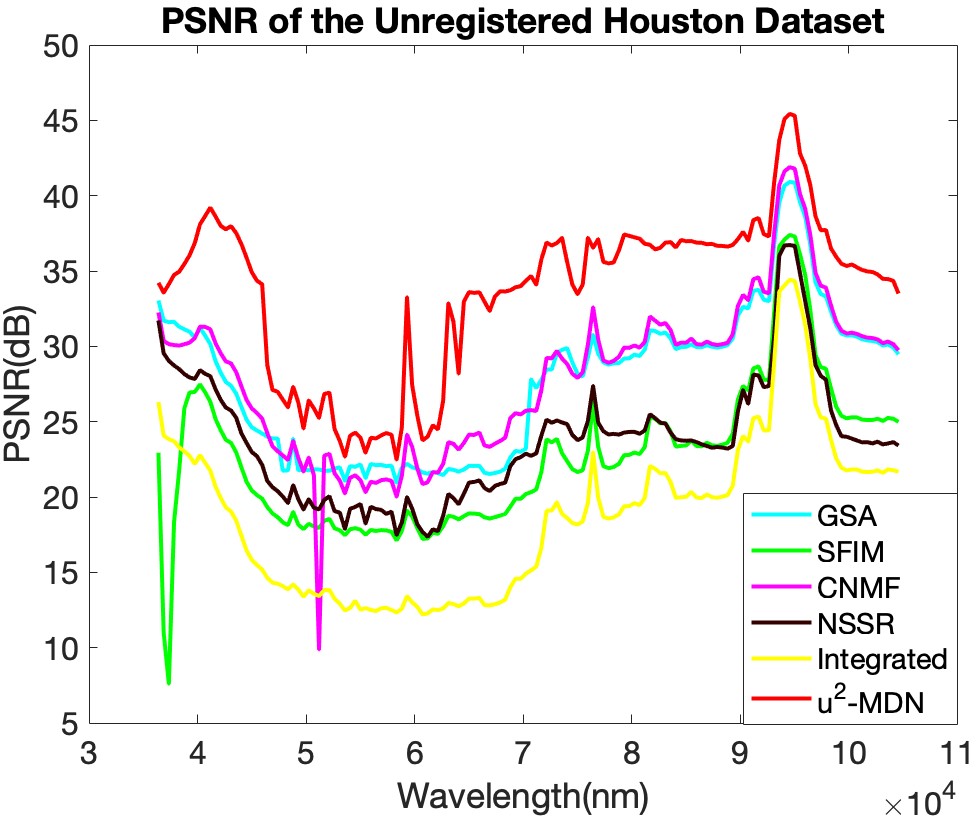}\label{fig:rs_psnr:b}}\\\hfill
		\subfloat[]{\includegraphics[width=0.5\linewidth]{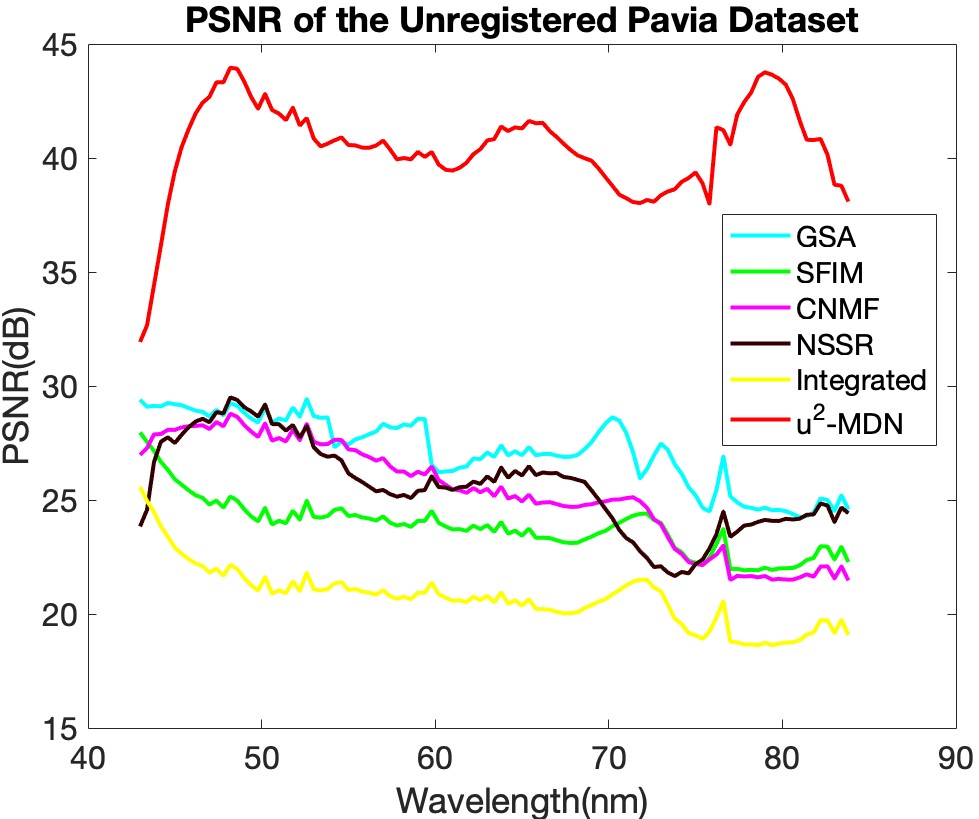}\label{fig:rs_psnr:c}}
		\subfloat[]{\includegraphics[width=0.5\linewidth]{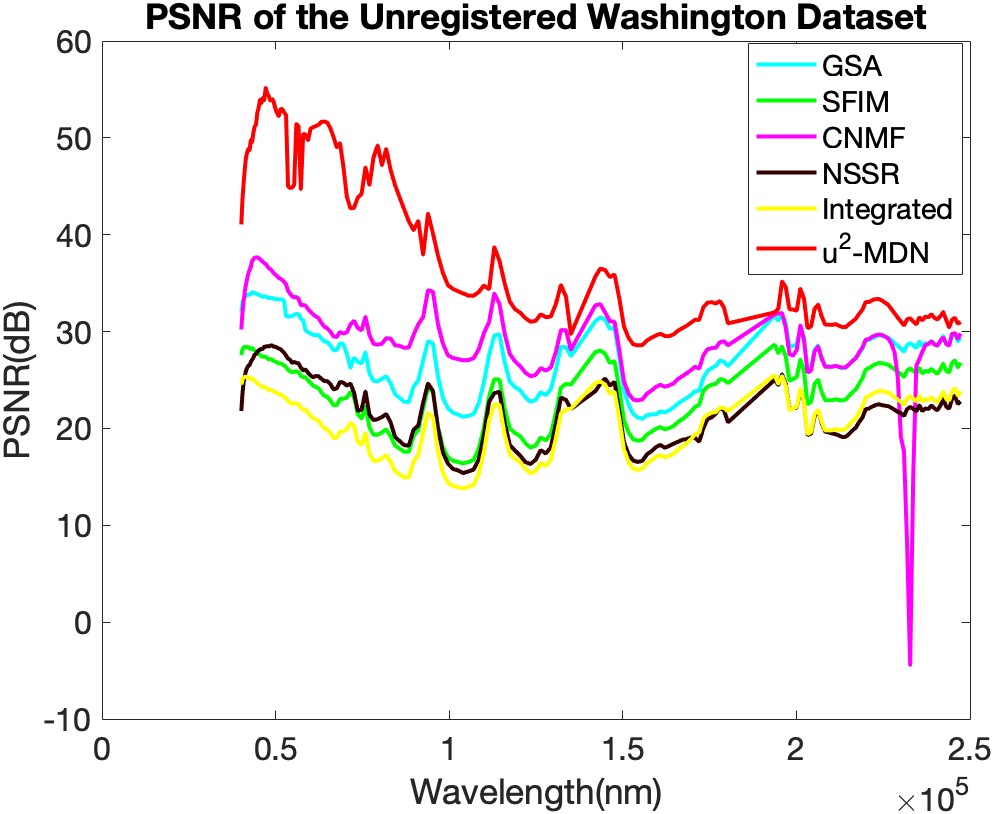}\label{fig:rs_psnr:d}}\hfill
	\end{center}
	\caption{The PSNR of different wavelengths for the reconstructed HSI from unregistered nonrigid distorted from (a) Chikusei, (b) Houston, (c) Pavia and (d) Washington datasets, respectively.}
	\label{fig:rs_psnr}
\end{figure}

The performance of different methods on unregistered image pairs are reported in Tables~\ref{tab:un_evaluate} and~\ref{tab:rs_unevaluate}. Note that, only the methods that are able to work with unregistered image pairs  are chosen in this group of experiments. Thus five state-of-the-art methods are compared in the tables. The traditional CS-based GSA~\cite{aiazzi2007improving} and MTF-based SFIM~\cite{liu2000smoothing} fail in this scenario. This is because when the given two modalities are unregistered, the spatial details could not be directly added to improve the spatial resolution of LR HSI. The matrix-factorization-based CNMF and sparse-coding based NSSR are more robust than the traditional methods. However, their performance also drops for both benchmarked and remote sensing datasets. The reason is that the adopted predefined down-sampling function will introduce significant spectral distortion when the LR HSI and HR MSI are unregistered. The integrated fusion method could achieve good performance on remote sensing images with small distortion. However, its performance drops on images with large distortion. This is because the integrated fusion method performs registration before fusion, which may introduce additional distortion during optimization. The proposed $u^2$-MDN is able to handle challenging scenarios much better than the state-of-the-art. The main reason that contributes to the success of the proposed approach is that, the network is able to extract the optimal and correlated spatial representations from two modalities through mutual information and collaborative loss. In this way, both the spatial and especially the spectral information are effectively preserved. This demonstrates the representation capacity of the proposed structure.

To demonstrate the reconstruction performance in different spectral bands, the average PSNR of the benchmarked datasets on each wavelength is shown in Fig.~\ref{fig:avg_psnr}. Since the numbers of spectral bands of the remote sensing datasets are different, we show their individual PSNR on each band in Fig.~\ref{fig:rs_psnr}. We can observe that, regardless of the type of the datasets, the proposed method consistently outperforms the other methods for all the spectral bands on unregistered image pairs. 

\begin{figure*}[htbp]
\setlength{\abovecaptionskip}{0.cm}
\setlength{\belowcaptionskip}{0.cm}
	\begin{center}
		\begin{minipage}{1\linewidth}
			\subfloat[Ref HR HSI]{\includegraphics[width=0.14\linewidth]{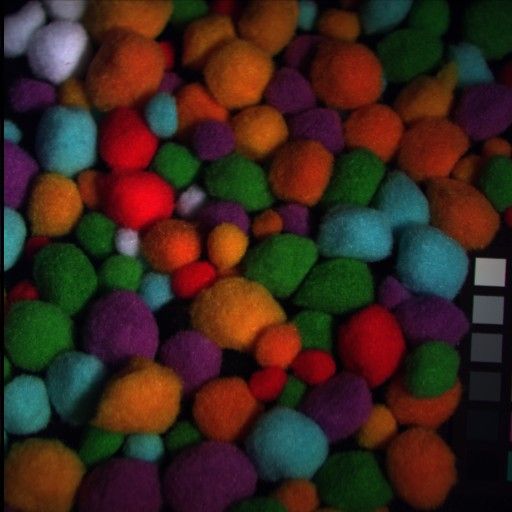}
				\label{fig:pom:a}}
			\subfloat[GSA]{\includegraphics[width=0.14\linewidth]{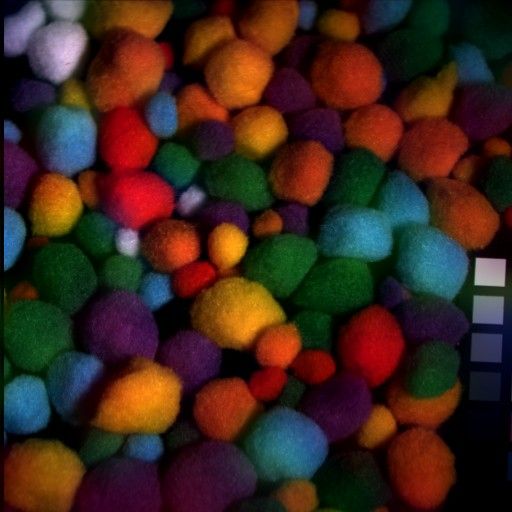} 
				\label{fig:pom:b}}
			\subfloat[SFIM]{\includegraphics[width=0.14\linewidth]{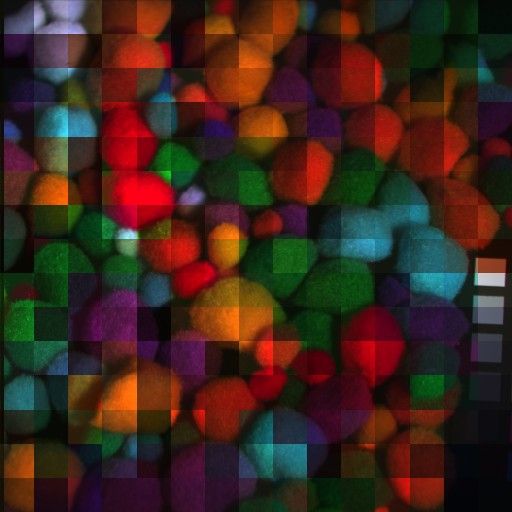}
				\label{fig:pom:c}}
			\subfloat[CNMF]{\includegraphics[width=0.14\linewidth]{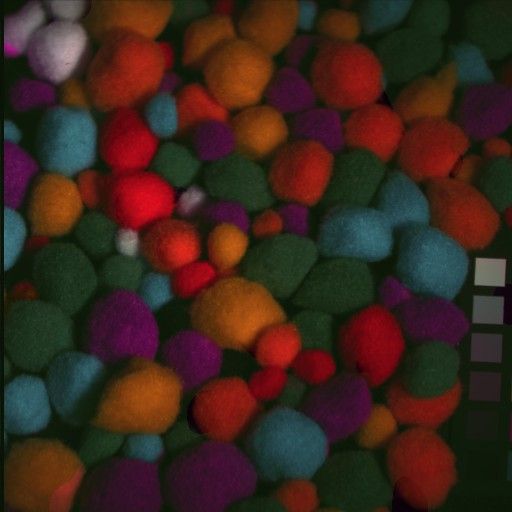}
				\label{fig:pom:d}}
			\subfloat[NSSR]{\includegraphics[width=0.14\linewidth]{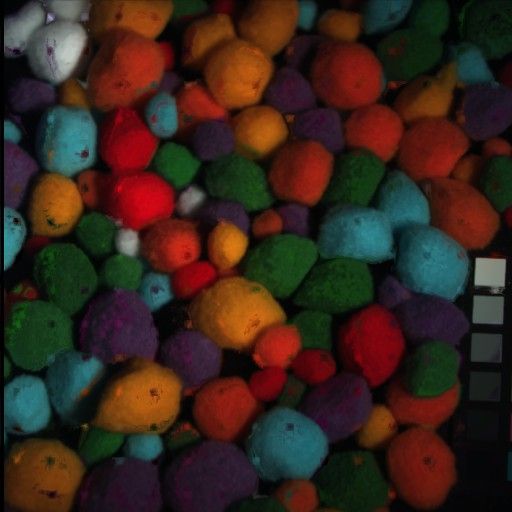}     
				\label{fig:pom:e}}
			\subfloat[Integrated]{\includegraphics[width=0.14\linewidth]{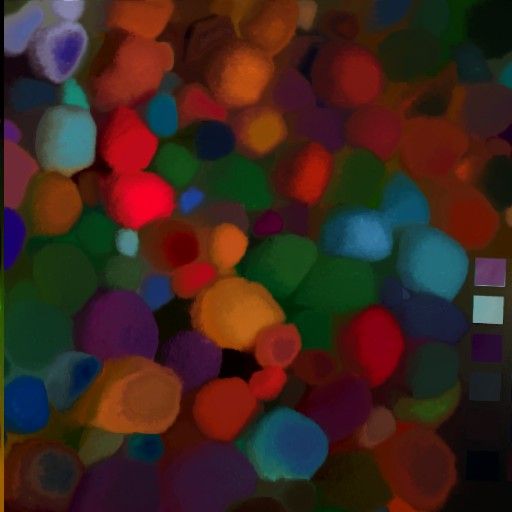}
				\label{fig:pom:f}}
			\subfloat[$u^2$-MDN]{\includegraphics[width=0.14\linewidth]{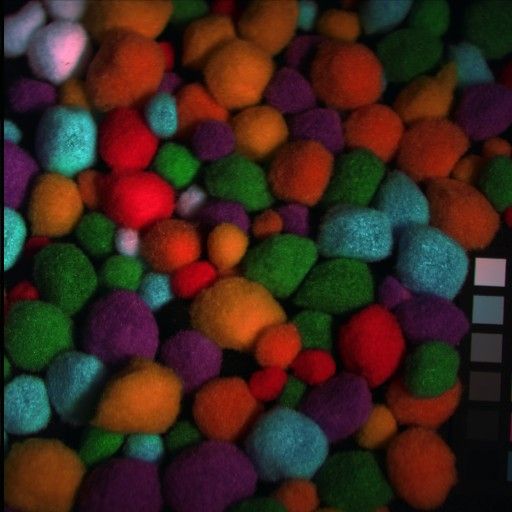}  
				\label{fig:pom:g}}\hfill\\
			\subfloat[Distorted LR HSI]{\includegraphics[width=0.14\linewidth]{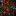}         
				\label{fig:pom:a2}}
			\subfloat[Difference of GSA]{\includegraphics[width=0.14\linewidth]{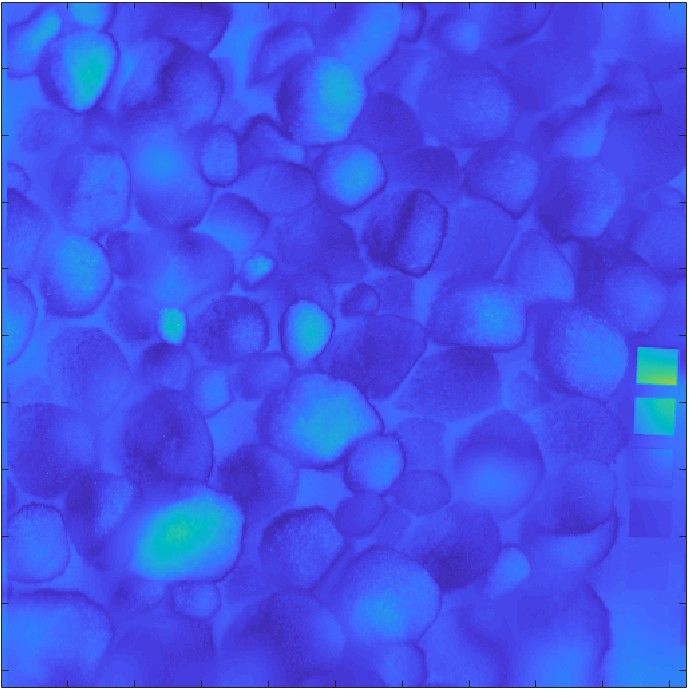}
				\label{fig:pom:b2}}
			\subfloat[Difference of SFIM]{\includegraphics[width=0.14\linewidth]{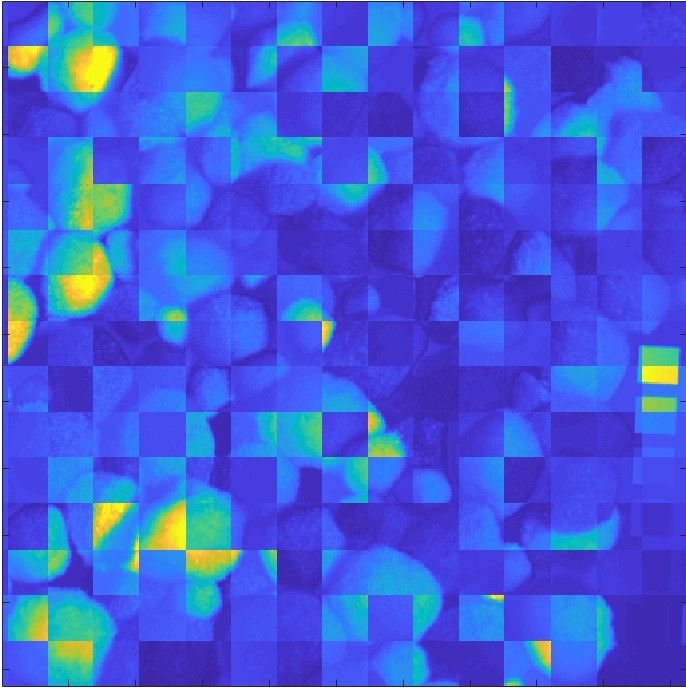}
				\label{fig:pom:c2}}
			\subfloat[Difference of CNMF]{\includegraphics[width=0.14\linewidth]{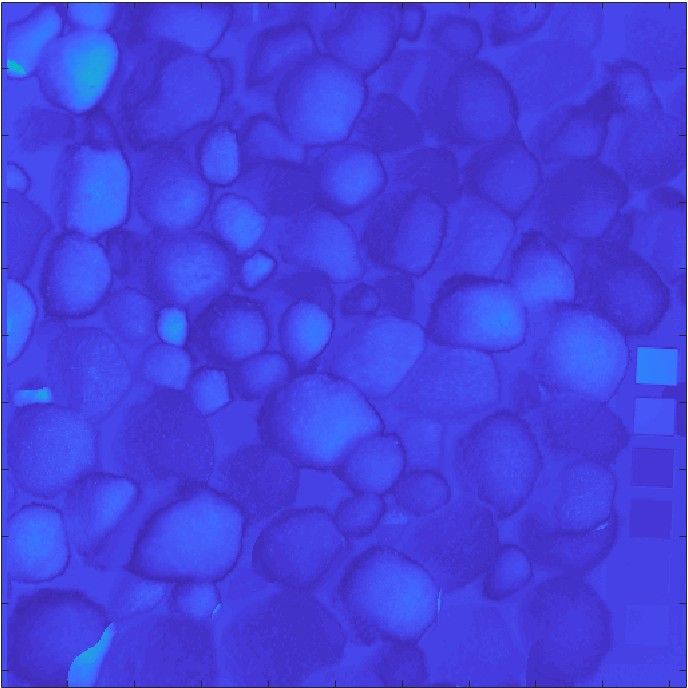}
				\label{fig:pom:d2}}
			\subfloat[Difference of NSSR]{\includegraphics[width=0.14\linewidth]{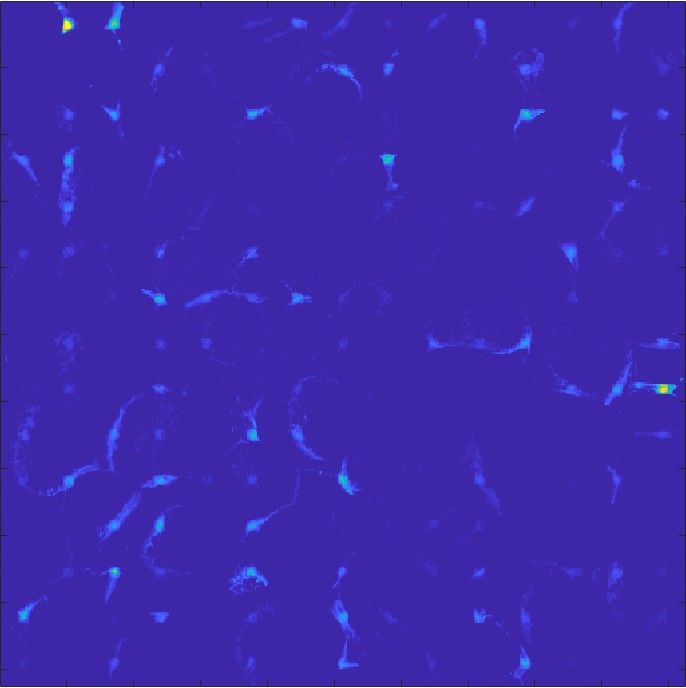}
				\label{fig:pom:e2}}
			\subfloat[Difference of Integrated]{\includegraphics[width=0.14\linewidth]{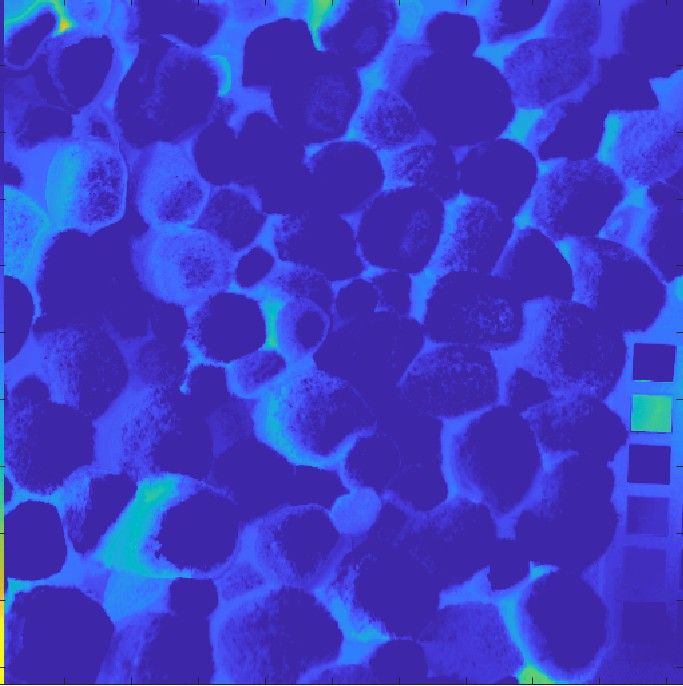}
				\label{fig:pom:f2}}
			\subfloat[Difference of $u^2$-MDN]{\includegraphics[width=0.165\linewidth]{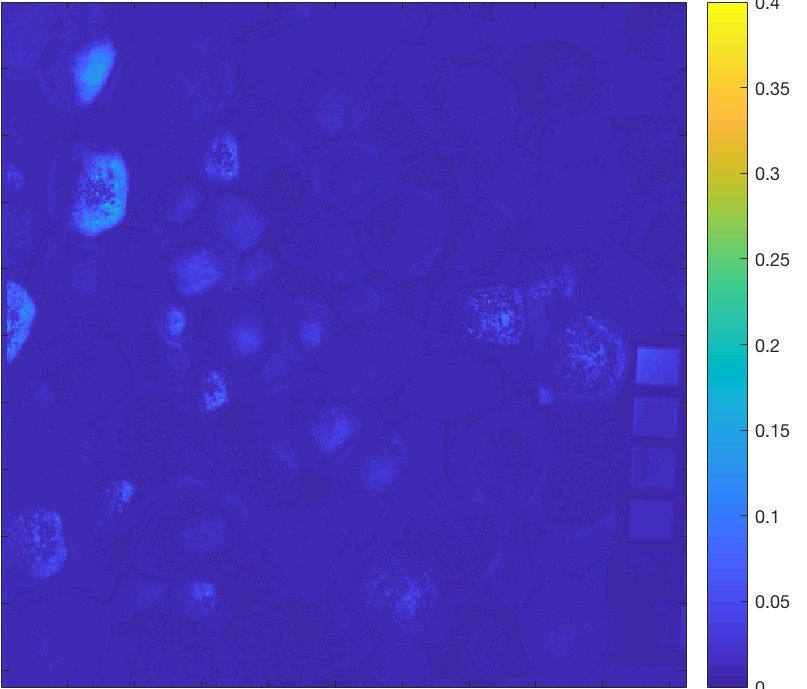}
				\label{fig:pom:g2}}\hfill\\
			\subfloat[LR HSI]{\includegraphics[width=0.14\linewidth]{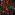}
				\label{fig:pom:a3}}
			\subfloat[SAM of GSA]{\includegraphics[width=0.14\linewidth]{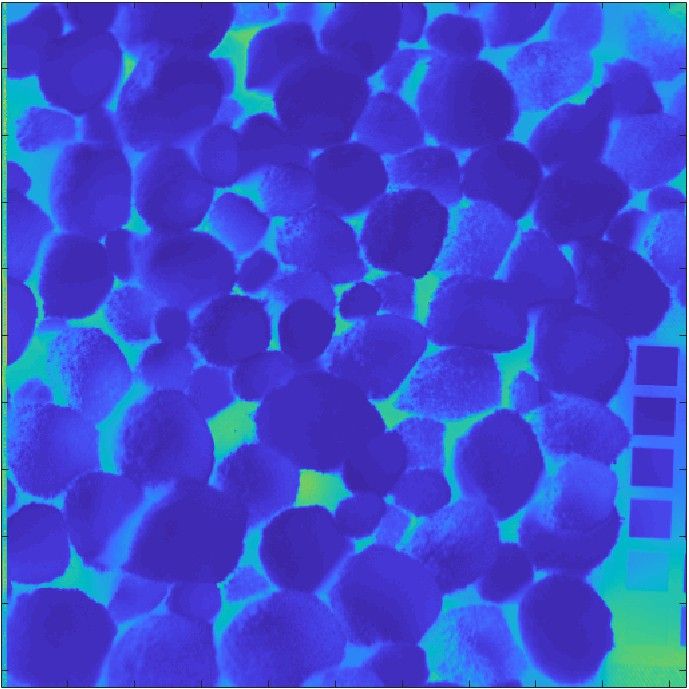}
				\label{fig:pom:b3}}
			\subfloat[SAM of SFIM]{\includegraphics[width=0.14\linewidth]{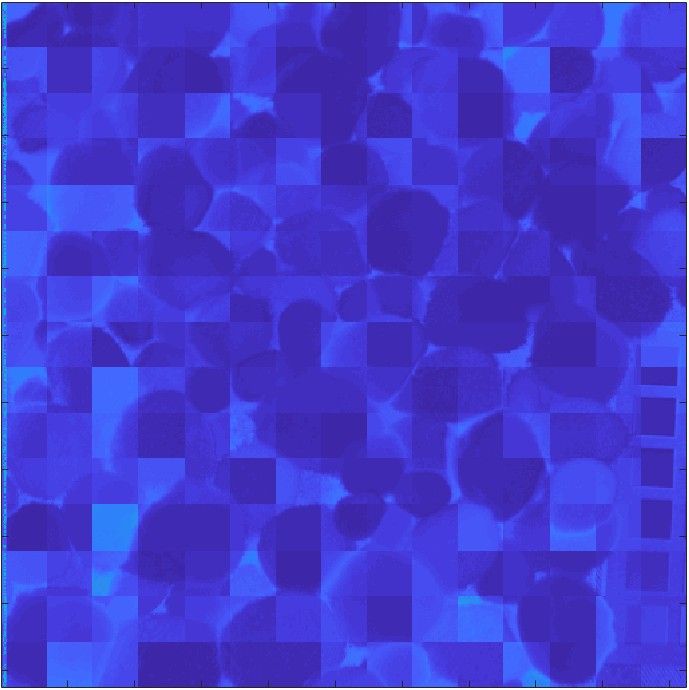}
				\label{fig:pom:c3}}
			\subfloat[SAM of CNMF]{\includegraphics[width=0.14\linewidth]{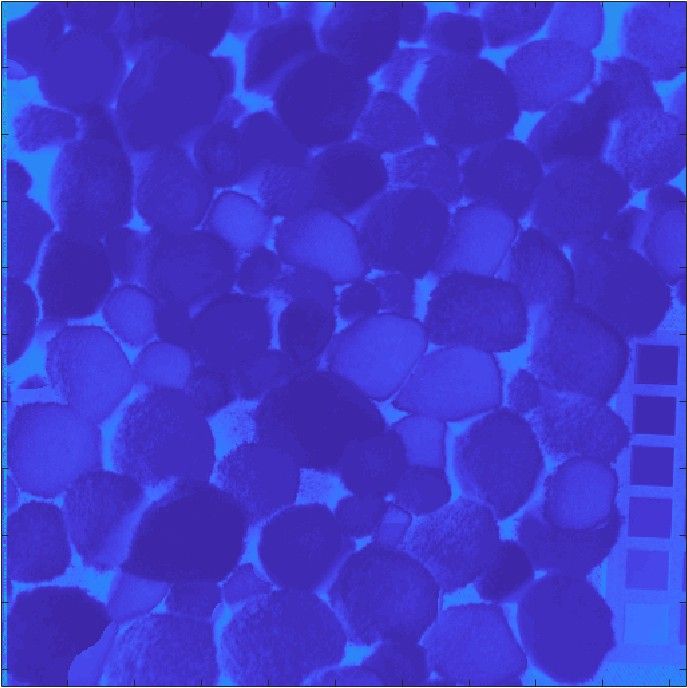}
				\label{fig:pom:d3}}
			\subfloat[SAM of NSSR]{\includegraphics[width=0.14\linewidth]{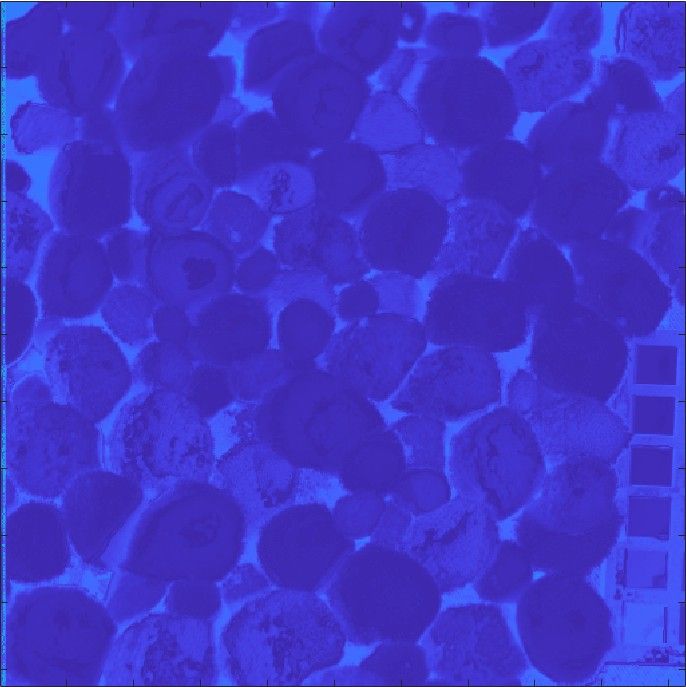}
				\label{fig:pom:e3}}
			\subfloat[SAM of Integrated]{\includegraphics[width=0.14\linewidth]{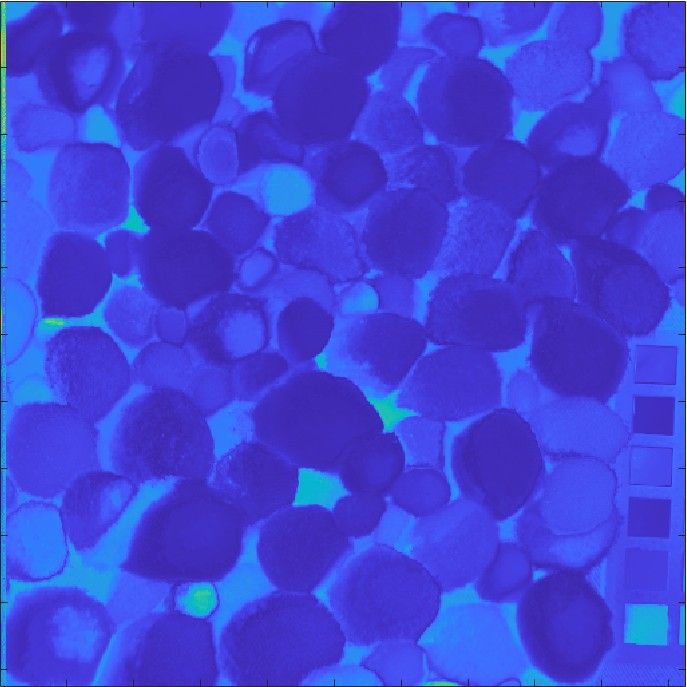}
				\label{fig:pom:e3}}
			\subfloat[SAM of $u^2$-MDN]{\includegraphics[width=0.16\linewidth]{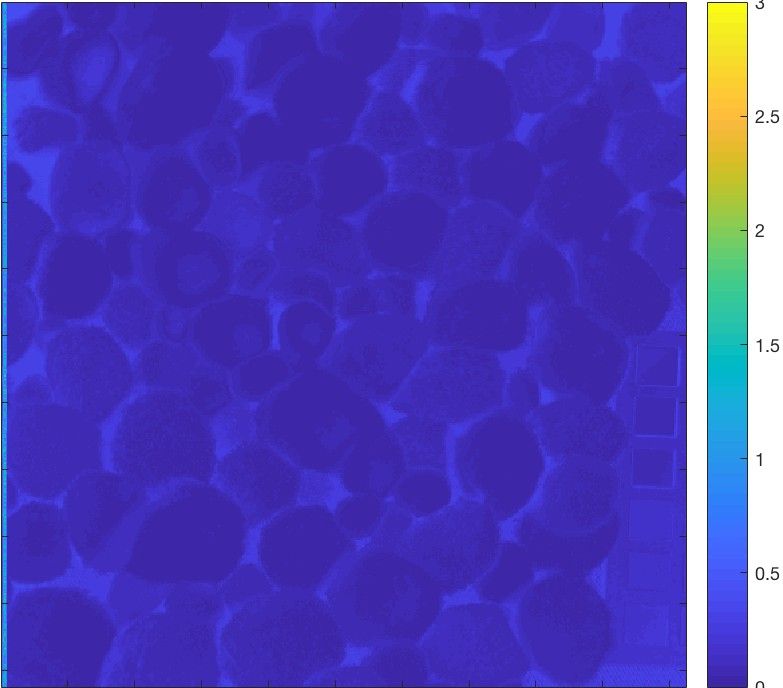}
				\label{fig:pom:g3}}\hfill\\
		\end{minipage}
	\end{center}
	\caption{Reconstructed results given unregistered rigid distorted image pairs from the CAVE dataset. (a) Color composite of the reference HR HSI. (h) Color composite of the distorted LR HSI. (o) Color composite of the LR HSI. (b)-(g): reconstructed results. (i)-(n): average absolute difference between the reconstructed HSI and reference HSI over different spectral bands, from different methods. (p)-(u) SAM of each pixel between the reconstructed HSI and reference HSI from different methods.}
	\label{fig:pom}
\end{figure*}

\begin{figure*}[htbp]
\setlength{\abovecaptionskip}{0.cm}
\setlength{\belowcaptionskip}{0.cm}
	\begin{center}
		\begin{minipage}{1\linewidth}
			\subfloat[Ref HR HSI]{\includegraphics[width=0.14\linewidth]{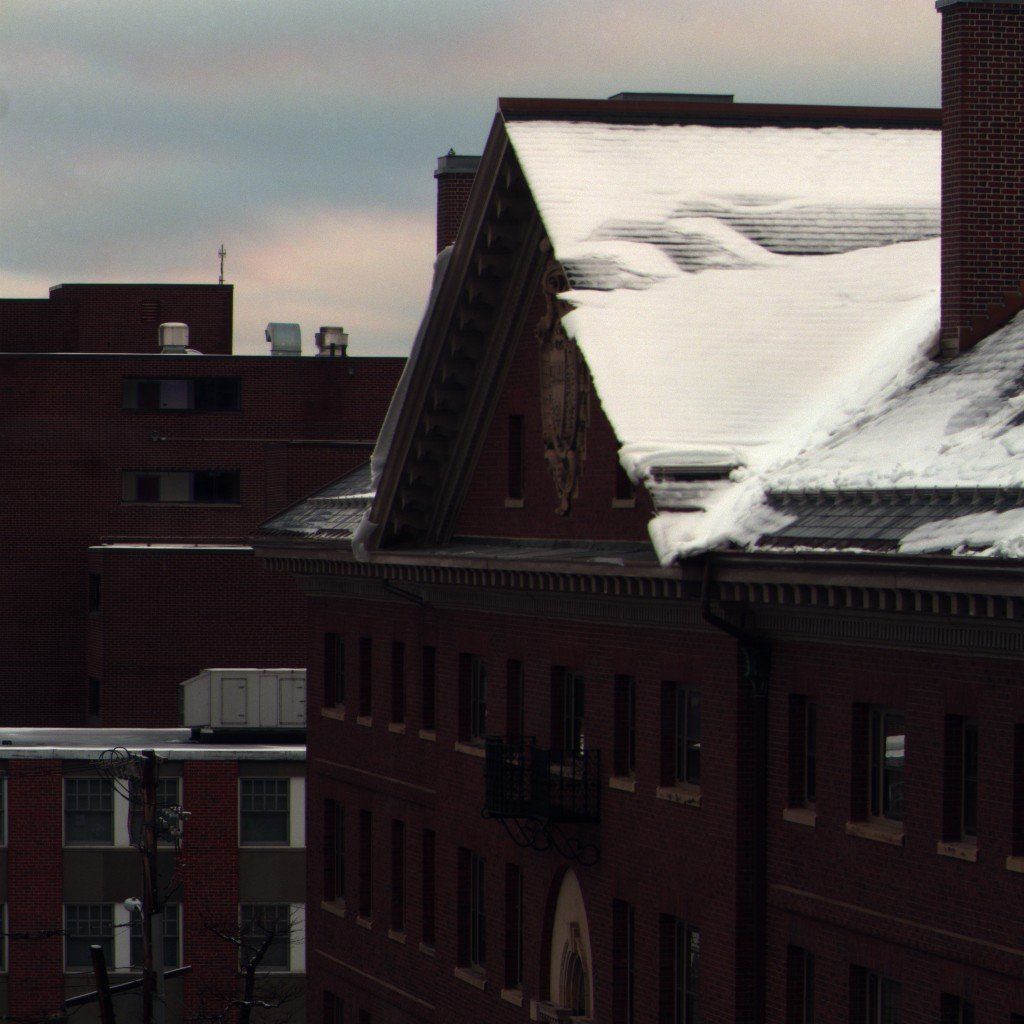}
				\label{fig:img1:a}}
			\subfloat[GSA]{\includegraphics[width=0.14\linewidth]{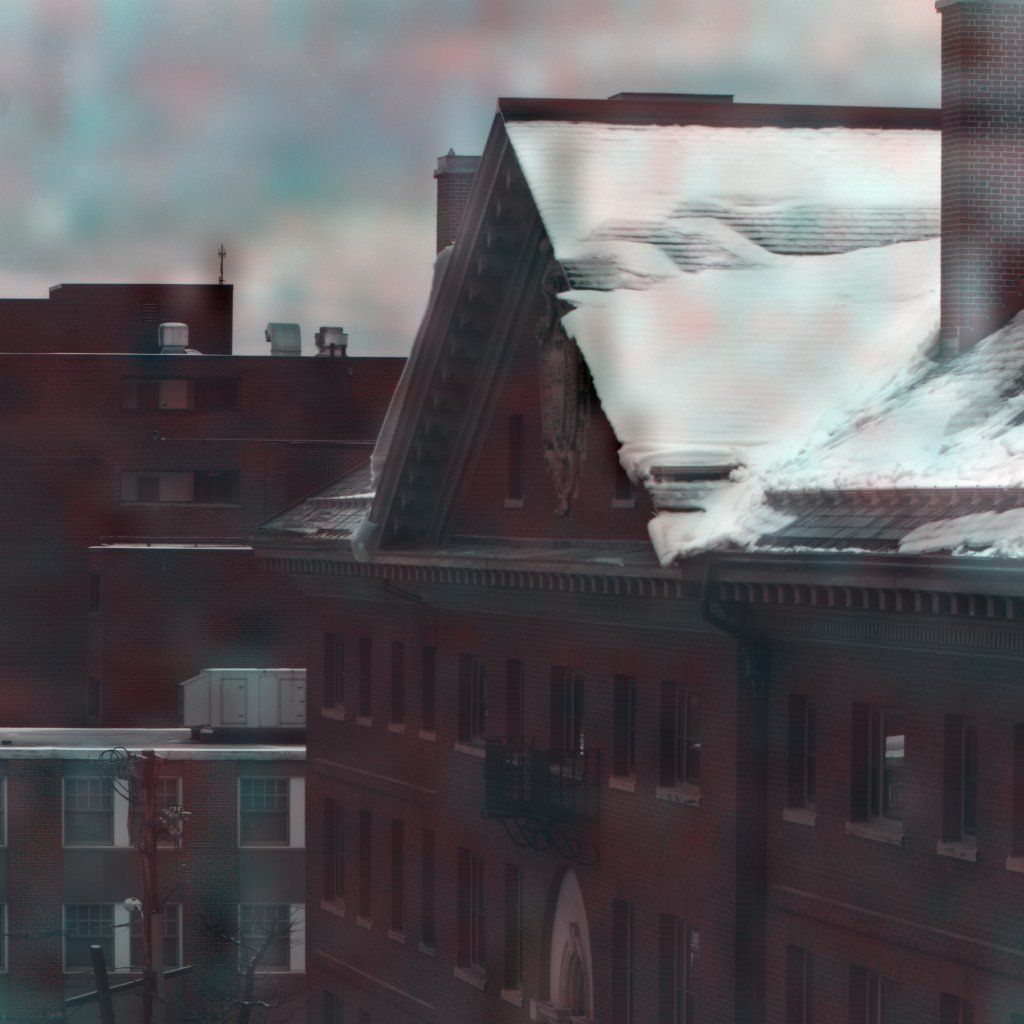} 
				\label{fig:img1:b}}
			\subfloat[SFIM]{\includegraphics[width=0.14\linewidth]{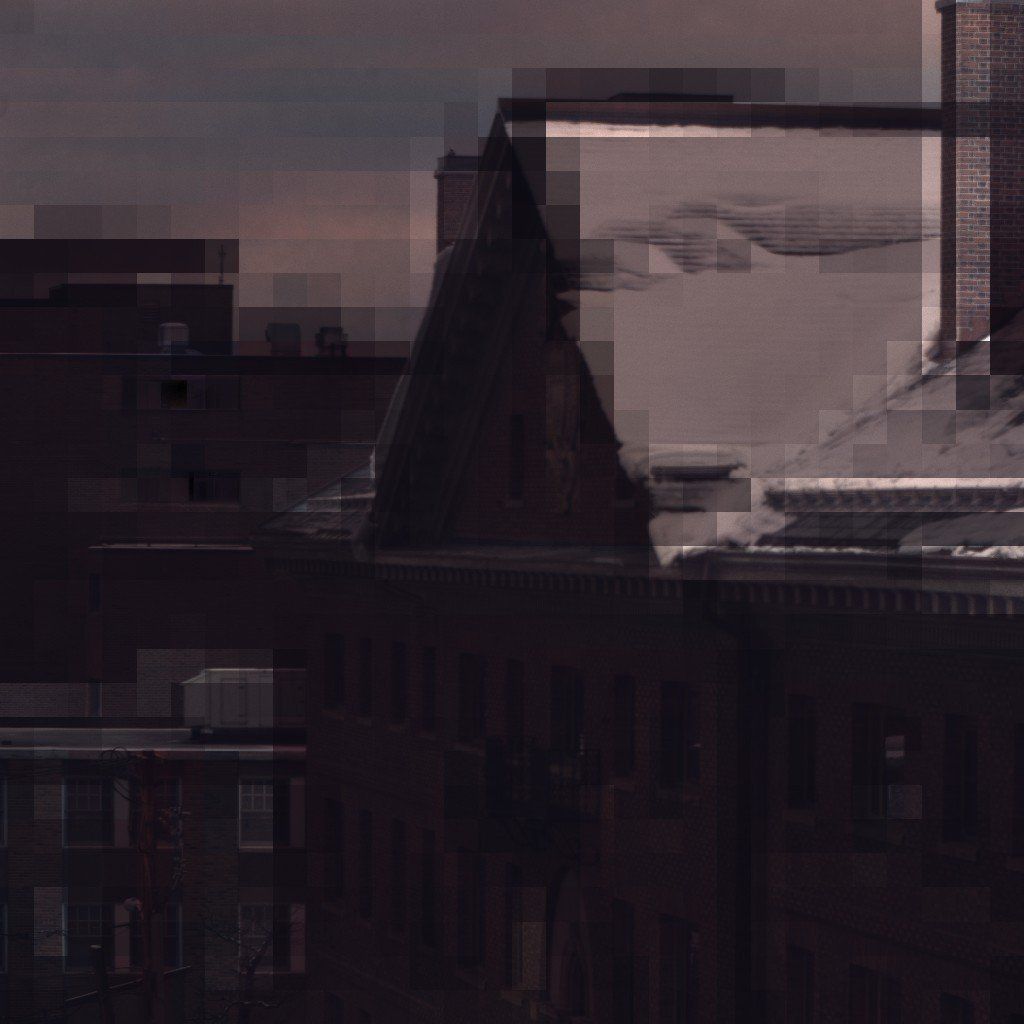}
				\label{fig:img1:c}}
			\subfloat[CNMF]{\includegraphics[width=0.14\linewidth]{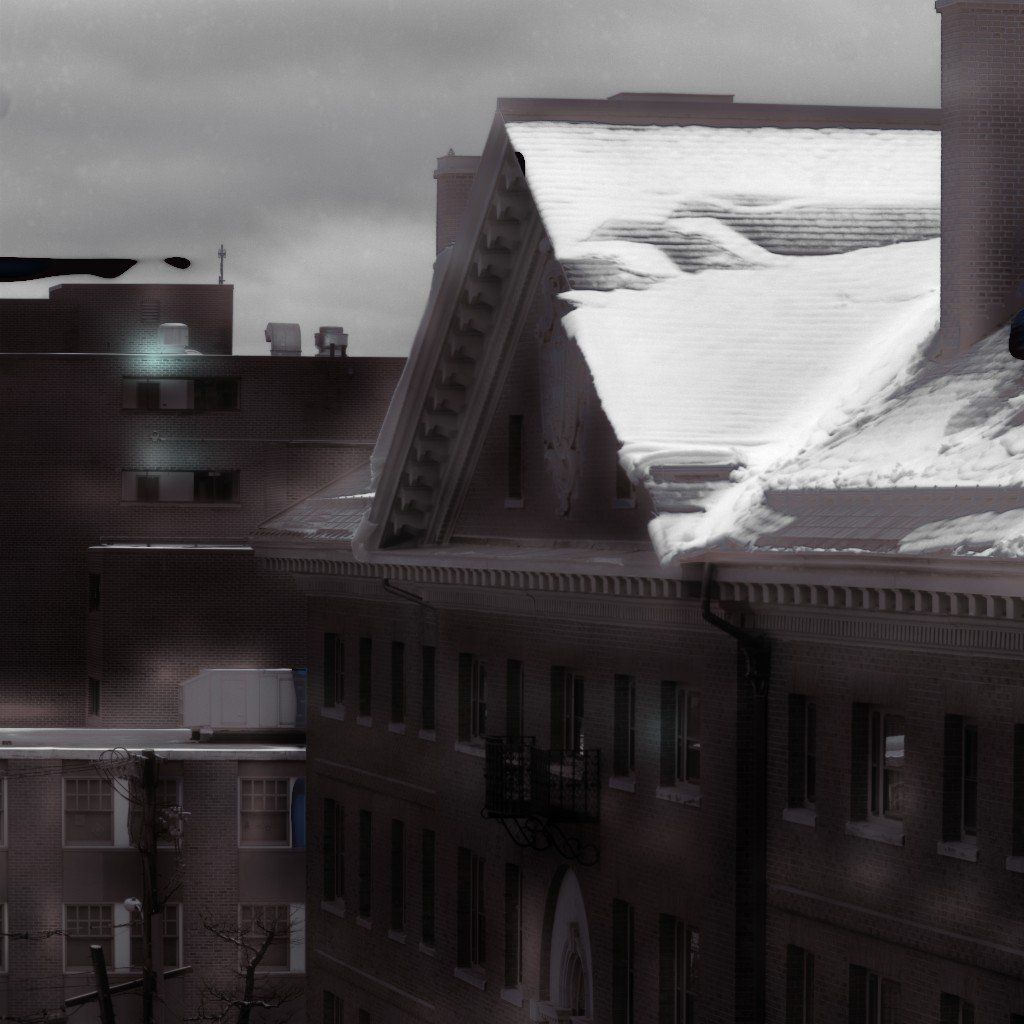}
				\label{fig:img1:d}}
			\subfloat[NSSR]{\includegraphics[width=0.14\linewidth]{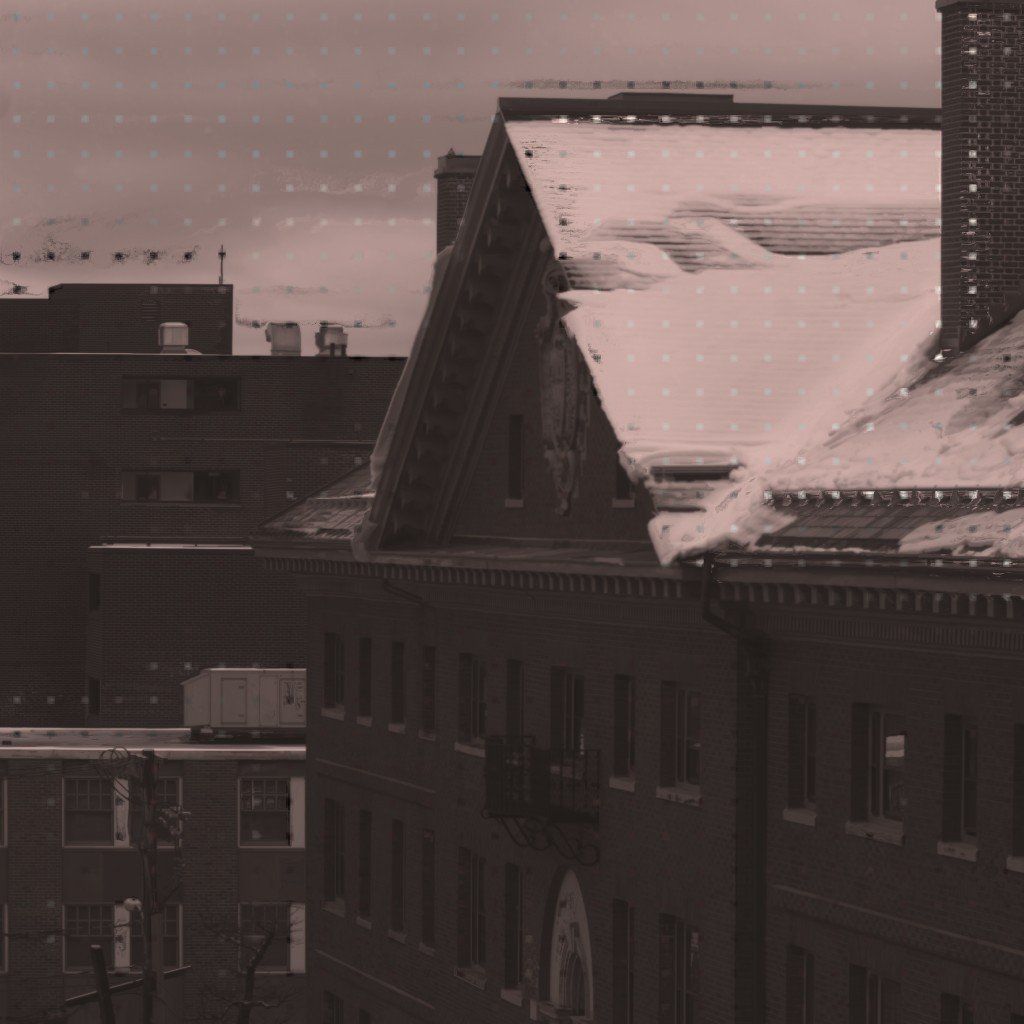}        
				\label{fig:img1:e}}
			\subfloat[Integrated]{\includegraphics[width=0.14\linewidth]{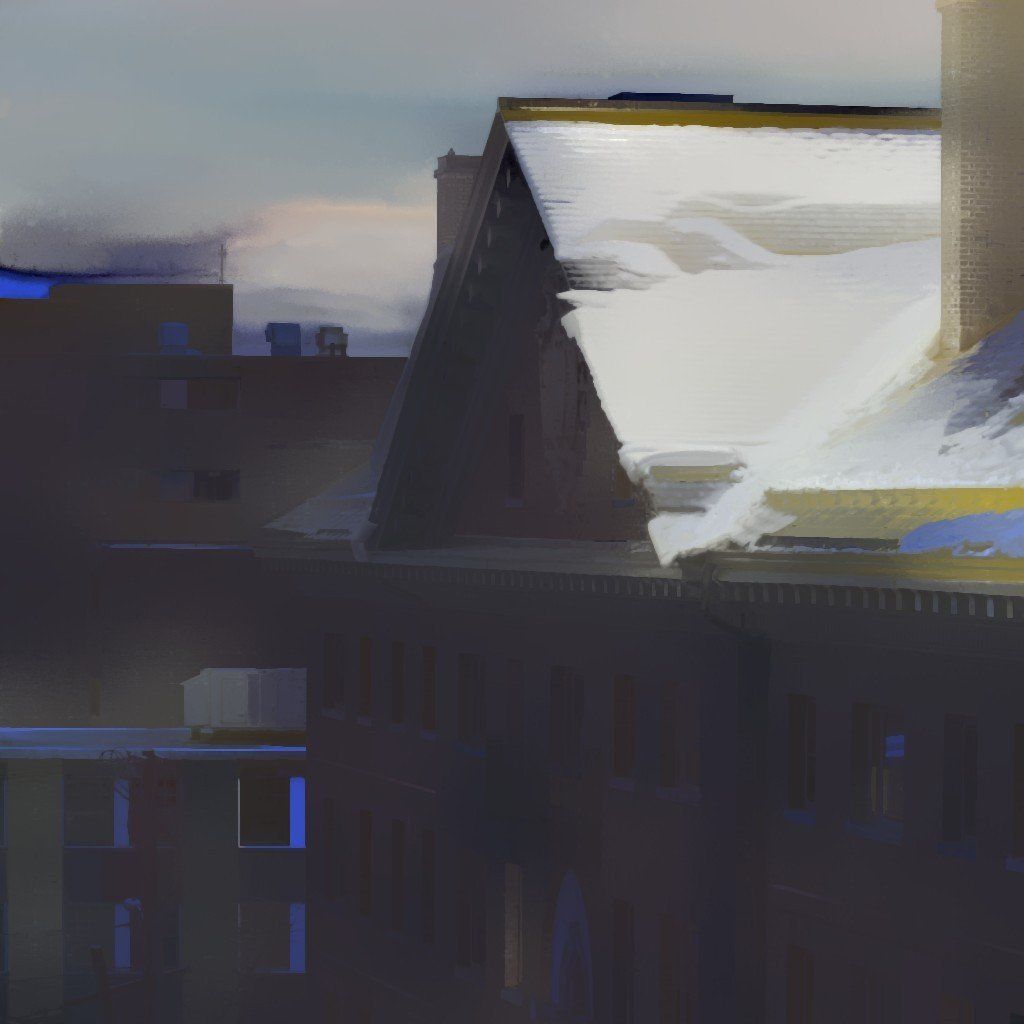}        
				\label{fig:img1:f}}
			\subfloat[$u^2$-MDN]{\includegraphics[width=0.14\linewidth]{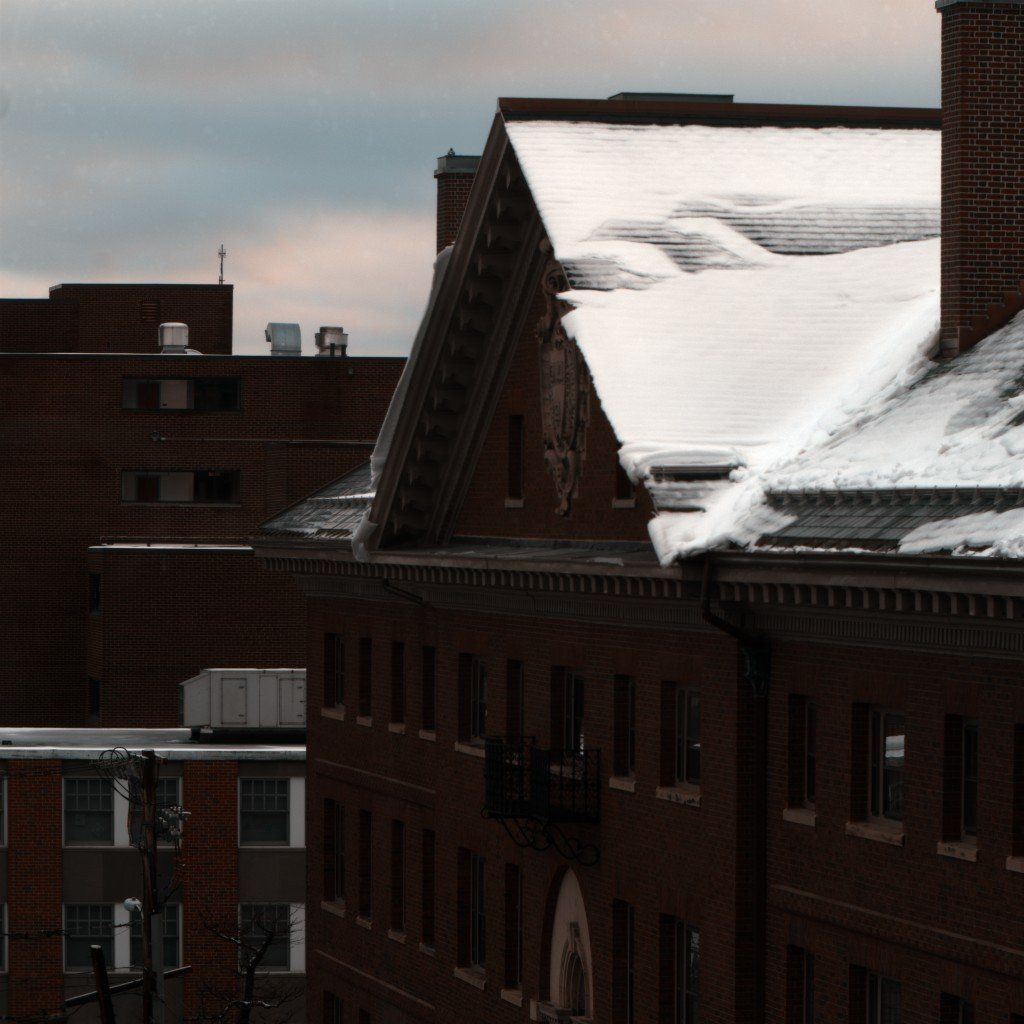}  
				\label{fig:img1:g}}\hfill\\
			\subfloat[Distorted LR HSI]{\includegraphics[width=0.14\linewidth]{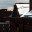}         
				\label{fig:img1:a2}}
			\subfloat[Difference of GSA]{\includegraphics[width=0.14\linewidth]{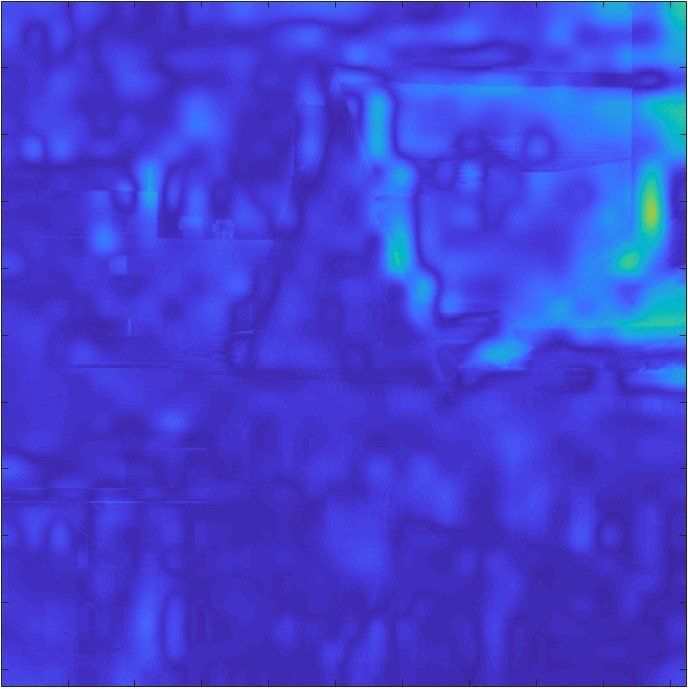}
				\label{fig:img1:b2}}
			\subfloat[Difference of SFIM]{\includegraphics[width=0.14\linewidth]{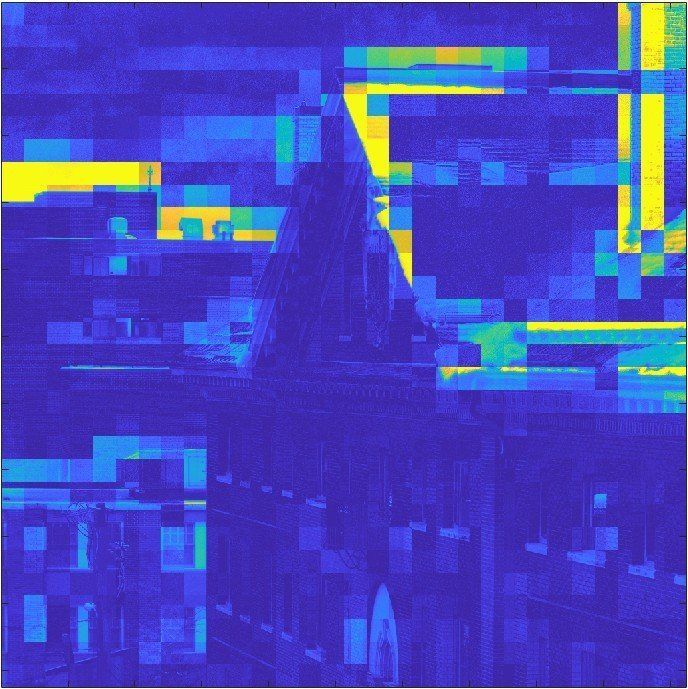}
				\label{fig:img1:c2}}
			\subfloat[Difference of CNMF]{\includegraphics[width=0.14\linewidth]{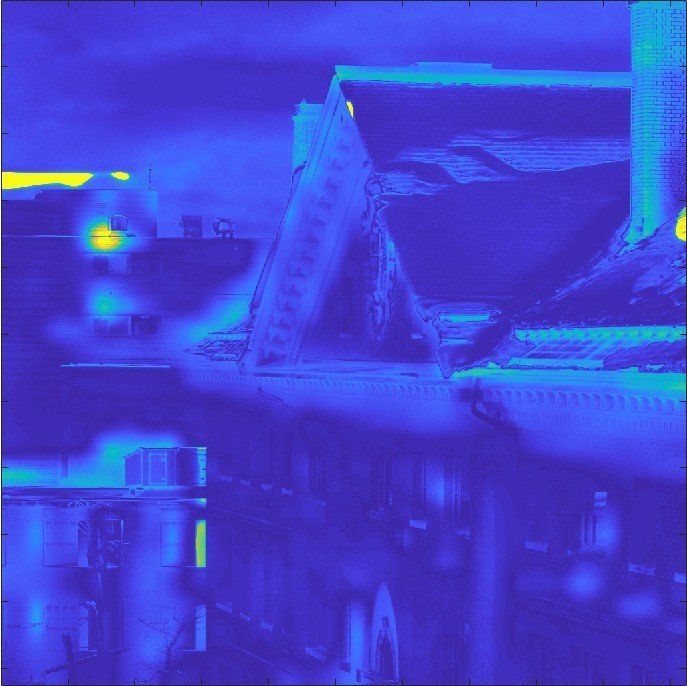}
				\label{fig:img1:d2}}
			\subfloat[Difference of NSSR]{\includegraphics[width=0.14\linewidth]{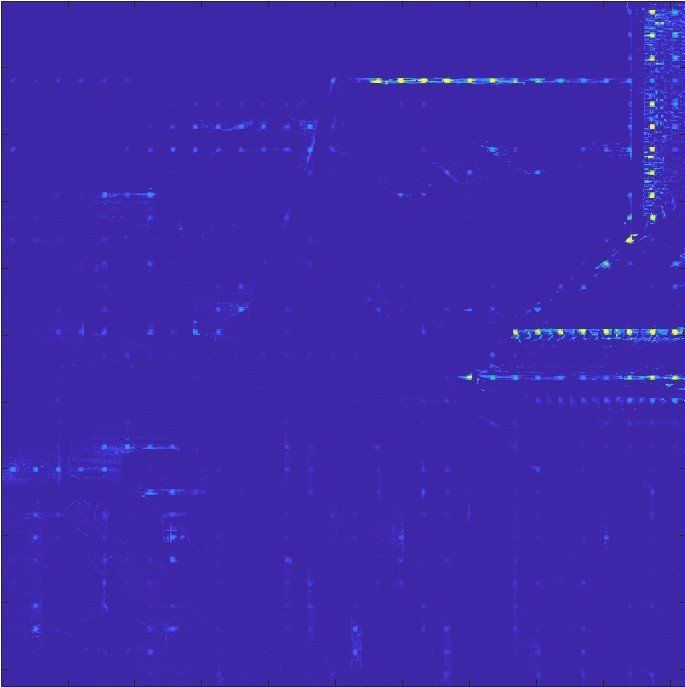}
				\label{fig:img1:e2}}
			\subfloat[Difference of Integrated]{\includegraphics[width=0.14\linewidth]{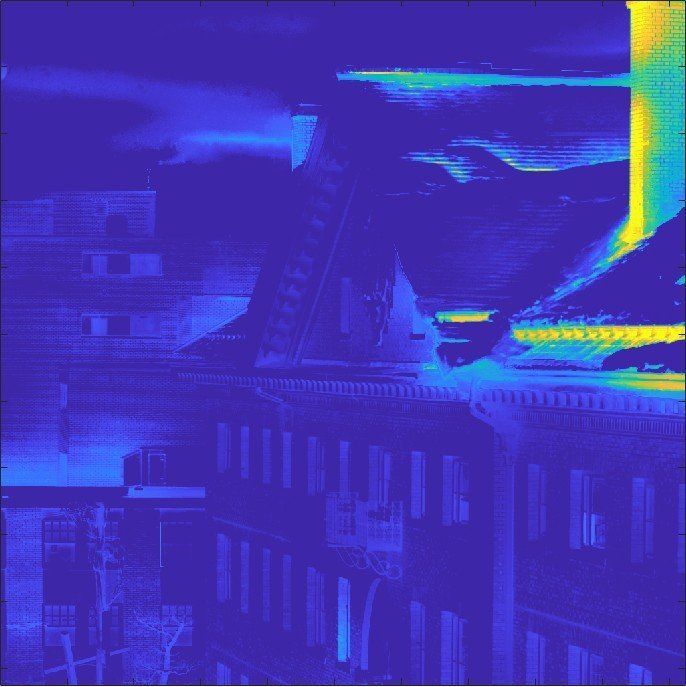}
				\label{fig:img1:f2}}
			\subfloat[Difference of $u^2$-MDN]{\includegraphics[width=0.165\linewidth]{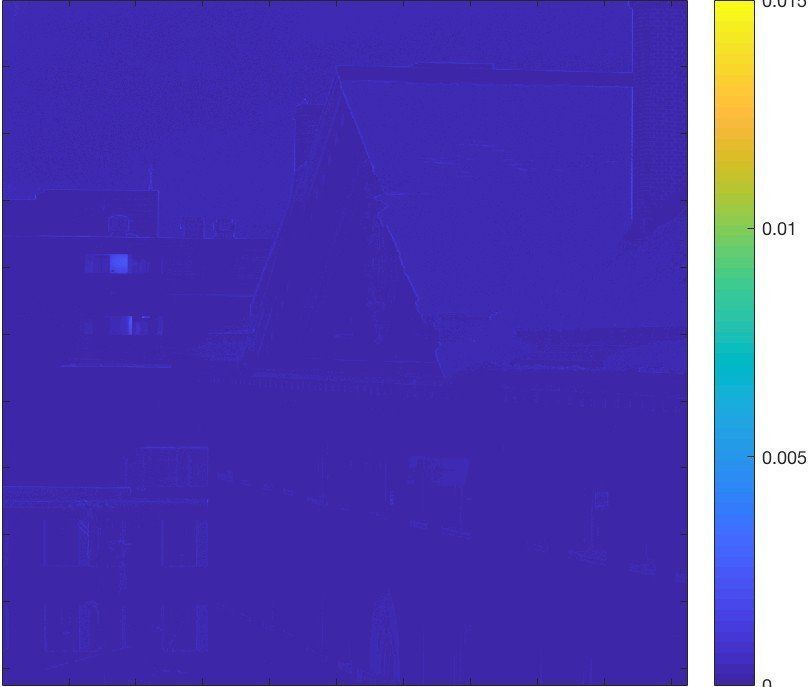}
				\label{fig:img1:g2}}\hfill\\
			\subfloat[LR HSI]{\includegraphics[width=0.14\linewidth]{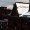}
				\label{fig:img1:a3}}
			\subfloat[SAM of GSA]{\includegraphics[width=0.14\linewidth]{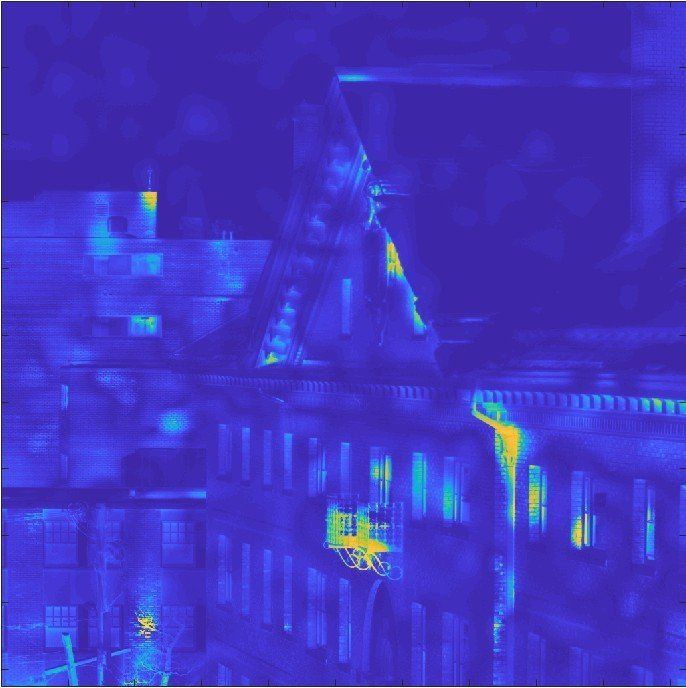}
				\label{fig:img1:b3}}
			\subfloat[SAM of SFIM]{\includegraphics[width=0.14\linewidth]{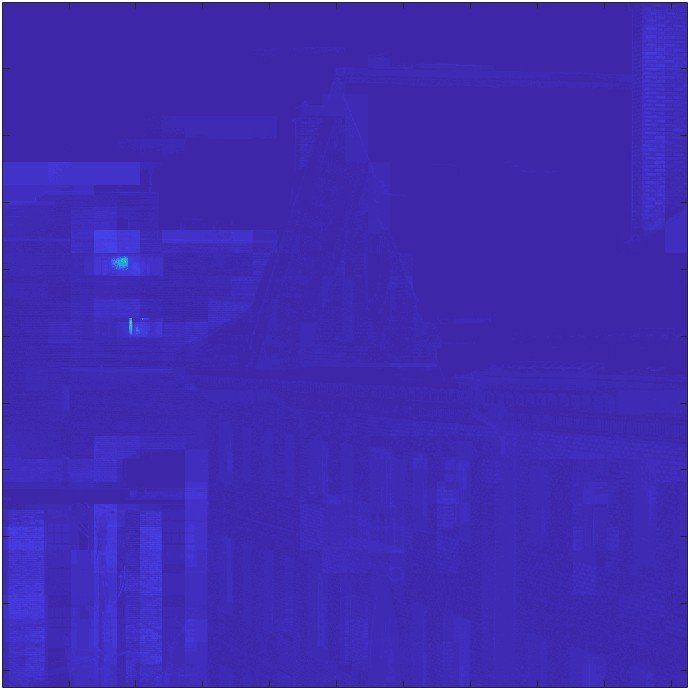}
				\label{fig:img1:c3}}
			\subfloat[SAM of CNMF]{\includegraphics[width=0.14\linewidth]{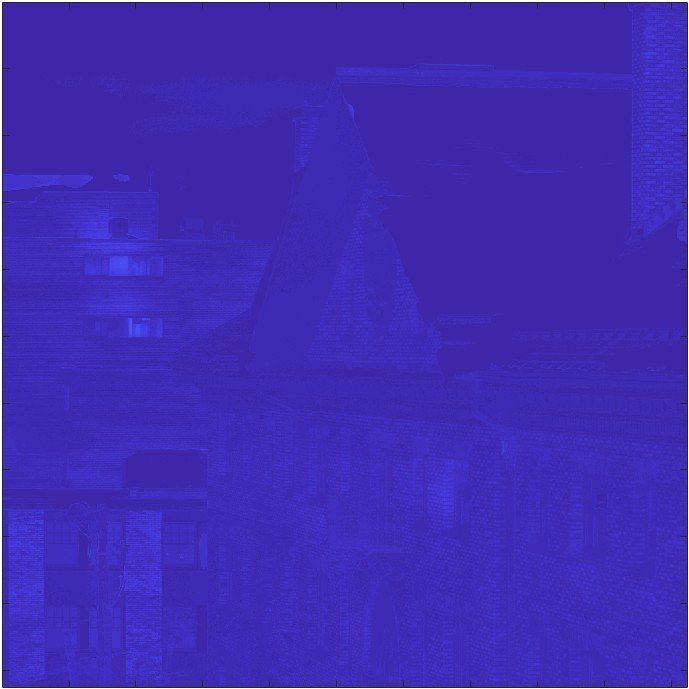}
				\label{fig:img1:d3}}
			\subfloat[SAM of NSSR]{\includegraphics[width=0.14\linewidth]{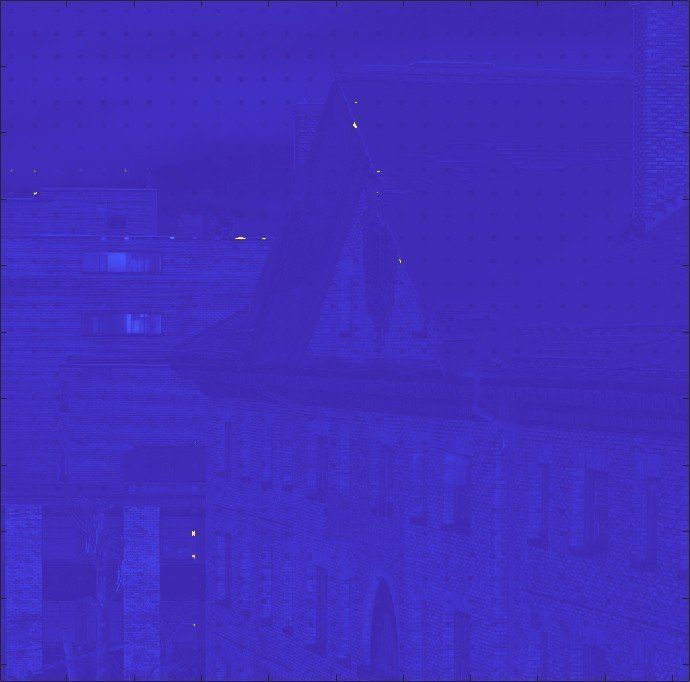}
				\label{fig:img1:e3}}
			\subfloat[SAM of Integrated]{\includegraphics[width=0.14\linewidth]{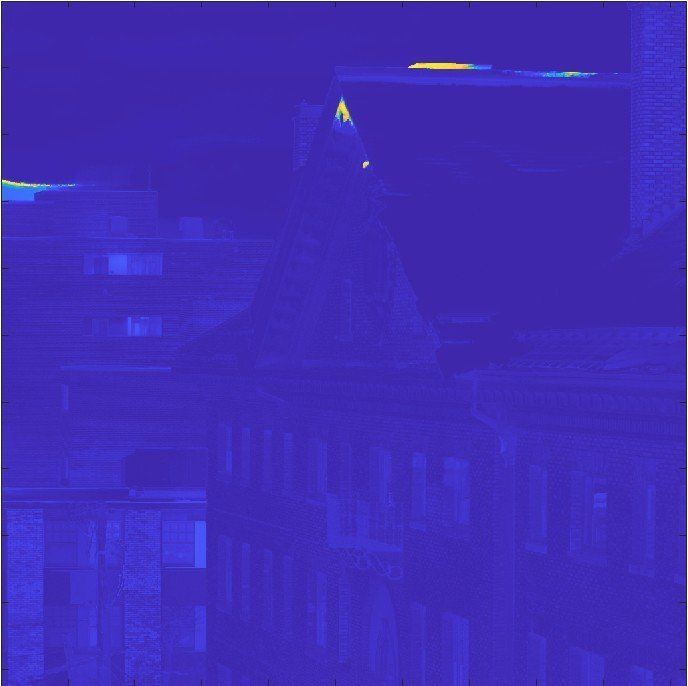}
				\label{fig:img1:f3}}
			\subfloat[SAM of $u^2$-MDN]{\includegraphics[width=0.16\linewidth]{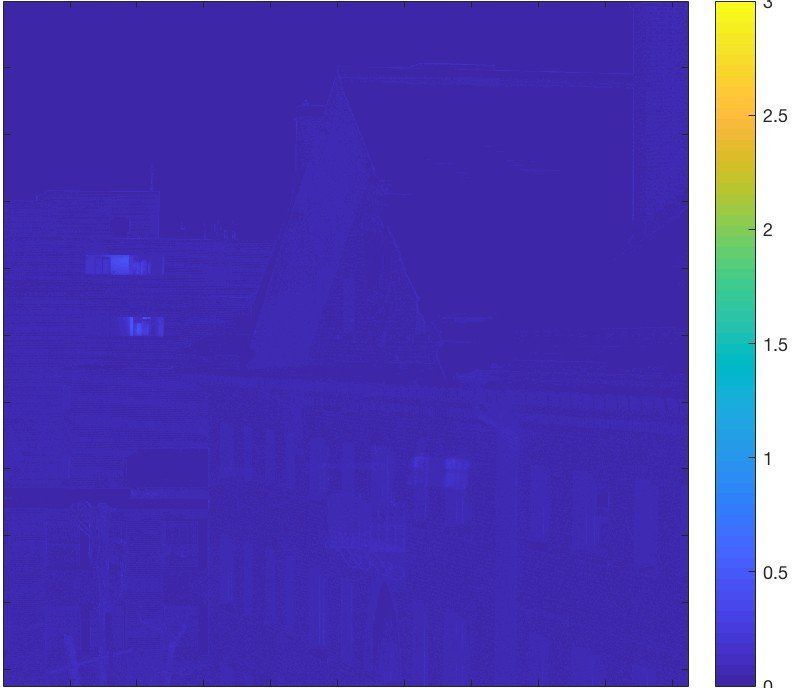}
				\label{fig:img1:g3}}\hfill\\
		\end{minipage}
	\end{center}
	\caption{Reconstructed results given unregistered rigid distorted  image pairs from the Harvard dataset. (a) Color composite of the reference HR HSI. (h) Color composite of the distorted LR HSI. (o) Color composite of the LR HSI. (b)-(g): reconstructed results. (i)-(n): average absolute difference between the reconstructed HSI and reference HSI over different spectral bands, from different methods. (p)-(u) SAM of each pixel between the reconstructed HSI and reference HSI from different methods.}
	\label{fig:img1}
\end{figure*}

\begin{figure*}[htbp]
\setlength{\abovecaptionskip}{0.cm}
\setlength{\belowcaptionskip}{0.cm}
	\begin{center}
		\begin{minipage}{1\linewidth}
			\subfloat[Ref HR HSI]{\includegraphics[width=0.14\linewidth]{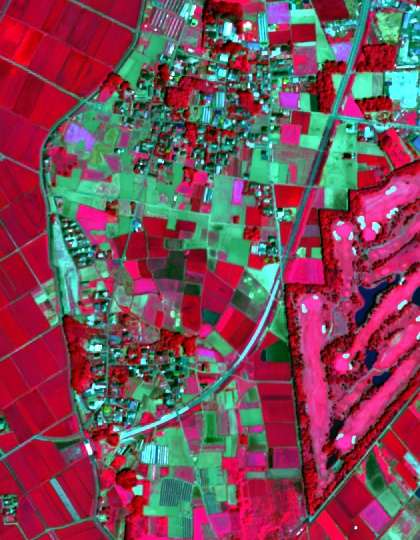}
				\label{fig:chiku:a}}
			\subfloat[GSA]{\includegraphics[width=0.14\linewidth]{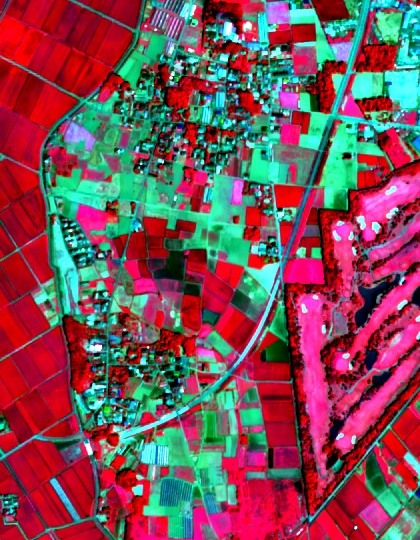} 
				\label{fig:chiku:b}}
			\subfloat[SFIM]{\includegraphics[width=0.14\linewidth]{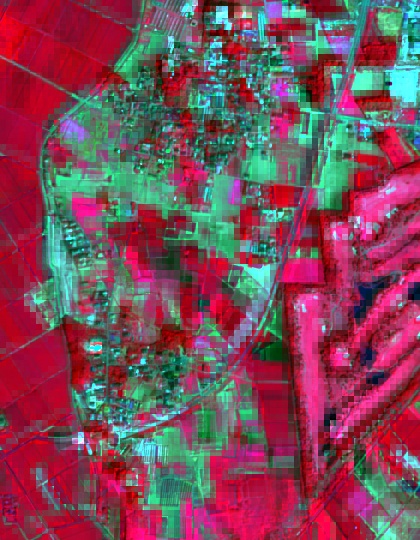}
				\label{fig:chiku:c}}
			\subfloat[CNMF]{\includegraphics[width=0.14\linewidth]{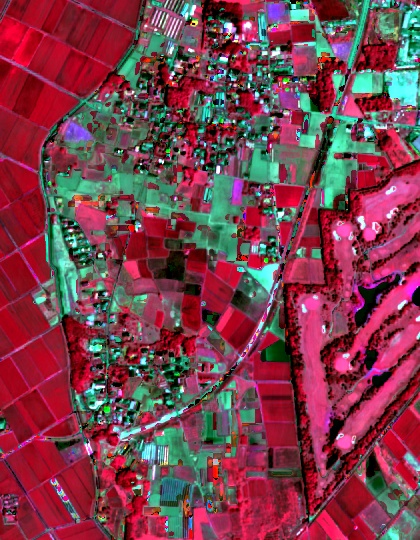}
				\label{fig:chiku:d}}
			\subfloat[NSSR]{\includegraphics[width=0.14\linewidth]{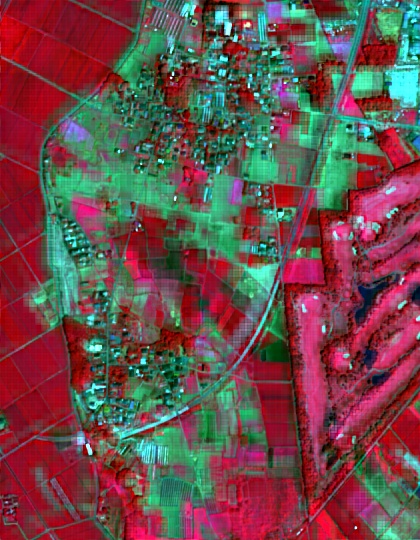}        
				\label{fig:chiku:e}}
			\subfloat[Integrated]{\includegraphics[width=0.14\linewidth]{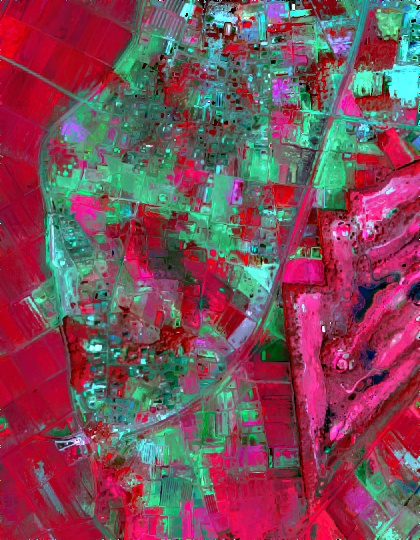}        
				\label{fig:chiku:f}}
			\subfloat[$u^2$-MDN]{\includegraphics[width=0.14\linewidth]{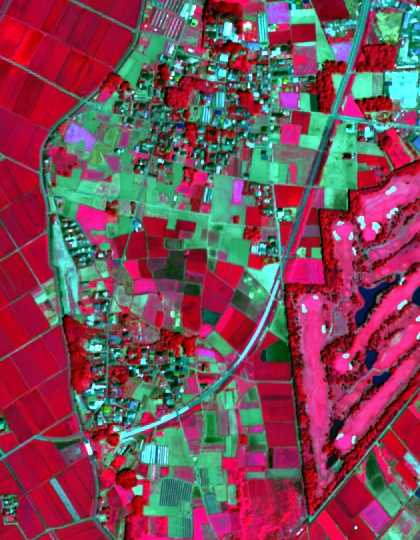}  
				\label{fig:chiku:g}}\hfill\\
			\subfloat[Distorted LR HSI]{\includegraphics[width=0.14\linewidth]{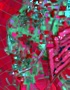}         
				\label{fig:chiku:a2}}
			\subfloat[Difference of GSA]{\includegraphics[width=0.14\linewidth]{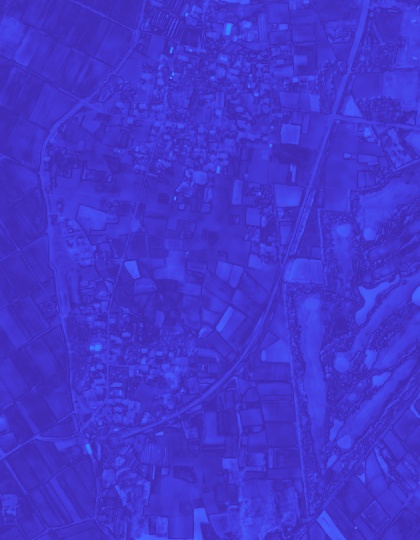}
				\label{fig:chiku:b2}}
			\subfloat[Difference of SFIM]{\includegraphics[width=0.14\linewidth]{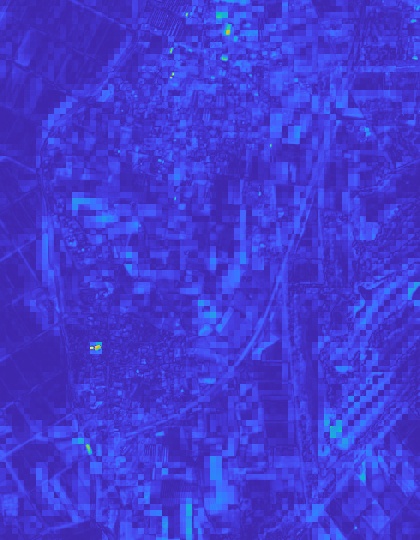}
				\label{fig:chiku:c2}}
			\subfloat[Difference of CNMF]{\includegraphics[width=0.14\linewidth]{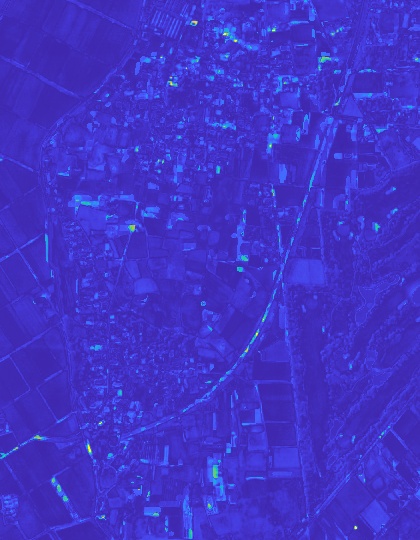}
				\label{fig:chiku:d2}}
			\subfloat[Difference of NSSR]{\includegraphics[width=0.14\linewidth]{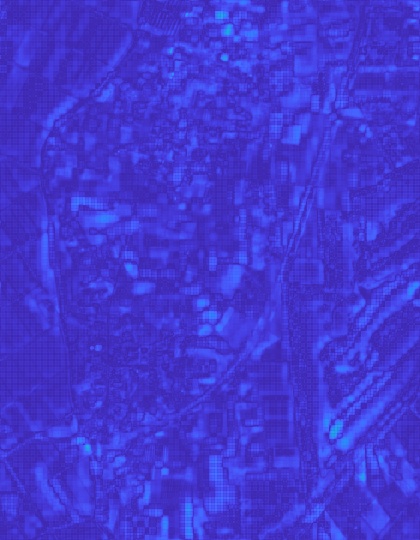}
				\label{fig:chiku:e2}}
			\subfloat[Difference of integrated]{\includegraphics[width=0.14\linewidth]{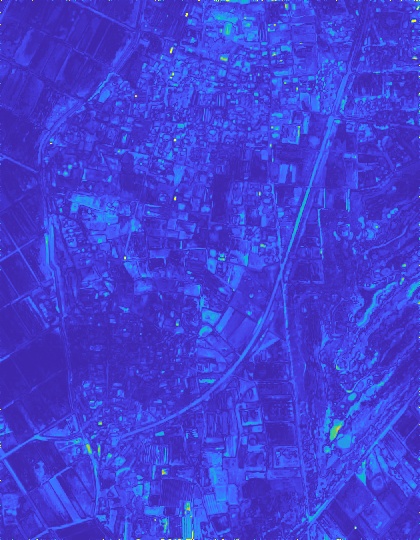}
				\label{fig:chiku:f2}}
			\subfloat[Difference of $u^2$-MDN]{\includegraphics[width=0.165\linewidth]{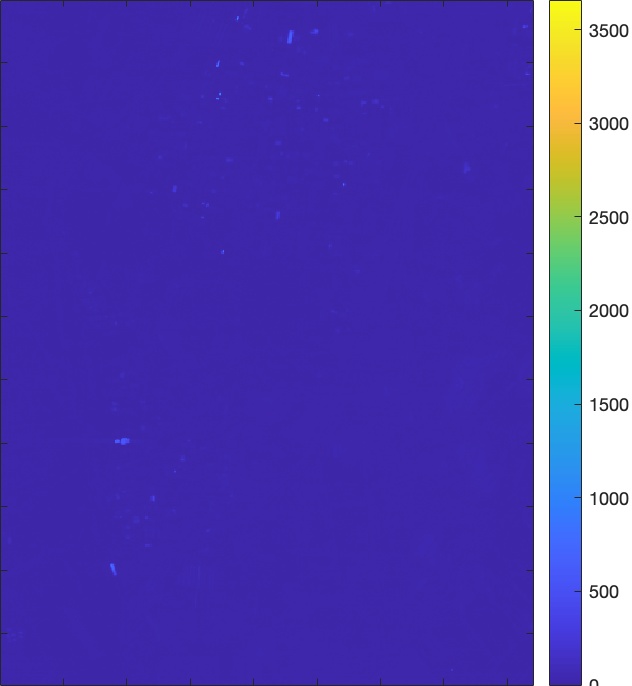}
				\label{fig:chiku:g2}}\hfill\\
			\subfloat[LR HSI]{\includegraphics[width=0.14\linewidth]{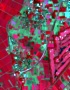}
				\label{fig:chiku:a3}}
			\subfloat[SAM of GSA]{\includegraphics[width=0.14\linewidth]{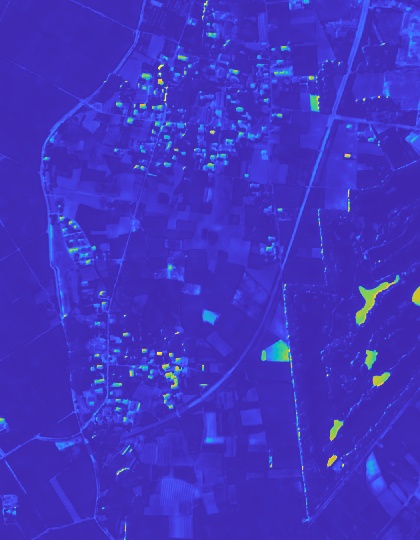}
				\label{fig:chiku:b3}}
			\subfloat[SAM of SFIM]{\includegraphics[width=0.14\linewidth]{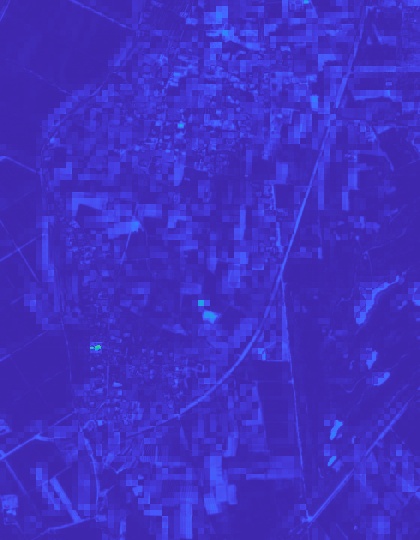}
				\label{fig:chiku:c3}}
			\subfloat[SAM of CNMF]{\includegraphics[width=0.14\linewidth]{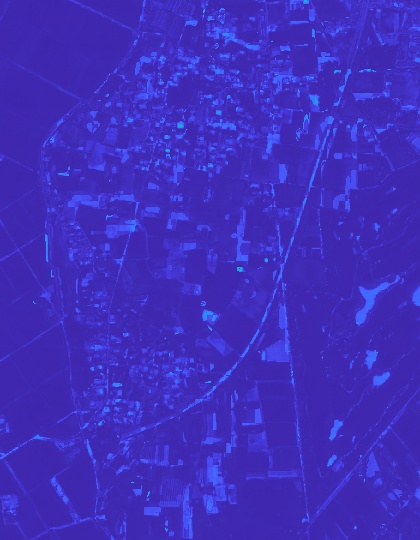}
				\label{fig:chiku:d3}}
			\subfloat[SAM of NSSR]{\includegraphics[width=0.14\linewidth]{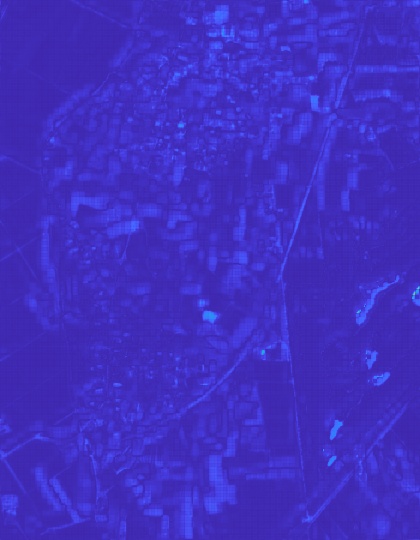}
				\label{fig:chiku:e3}}
			\subfloat[SAM of Integrated]{\includegraphics[width=0.14\linewidth]{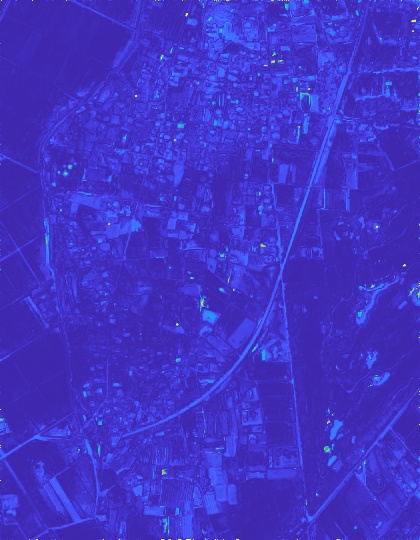}
				\label{fig:chiku:f3}}
			\subfloat[SAM of $u^2$-MDN]{\includegraphics[width=0.16\linewidth]{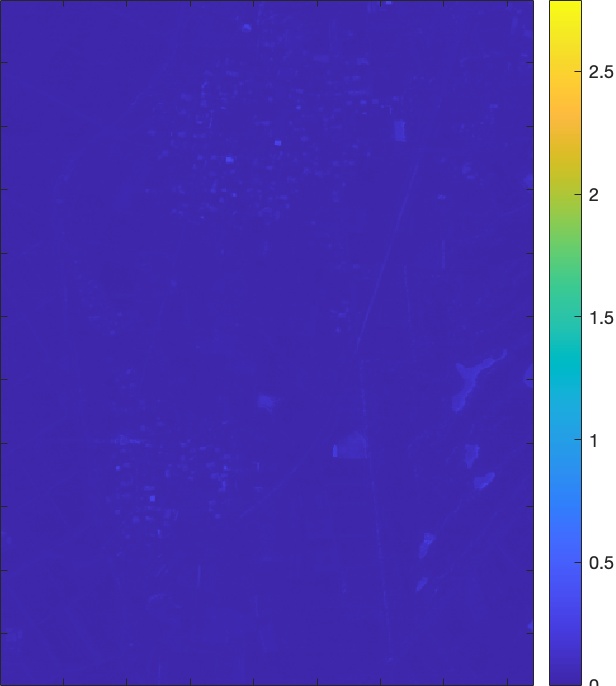}
				\label{fig:chiku:g3}}\hfill\\
		\end{minipage}
	\end{center}
	\caption{Reconstructed results given unregistered nonrigid distorted image pairs from the Chikusei dataset. (a) Color composite of the reference HR HSI. (h) Color composite of the distorted LR HSI. (o) Color composite of the LR HSI. (b)-(g): reconstructed results. (i)-(n): average absolute difference between the reconstructed HSI and reference HSI over different spectral bands, from different methods. (p)-(u) SAM of each pixel between the reconstructed HSI and reference HSI from different methods.}
	\label{fig:chiku}
\end{figure*}

\begin{figure*}[htbp]
\setlength{\abovecaptionskip}{0.cm}
\setlength{\belowcaptionskip}{0.cm}
	\begin{center}
		\begin{minipage}{1\linewidth}
			\subfloat[Ref HR HSI]{\includegraphics[width=0.14\linewidth]{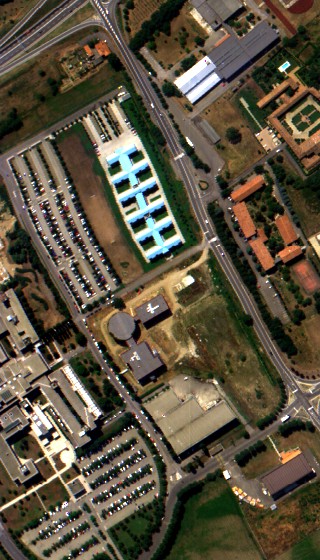}
				\label{fig:pavia:a}}
			\subfloat[GSA]{\includegraphics[width=0.14\linewidth]{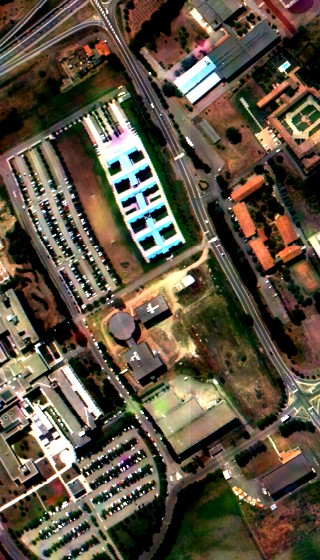} 
				\label{fig:pavia:b}}
			\subfloat[SFIM]{\includegraphics[width=0.14\linewidth]{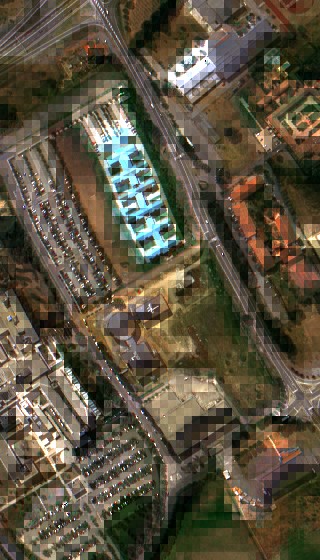}
				\label{fig:pavia:c}}
			\subfloat[CNMF]{\includegraphics[width=0.14\linewidth]{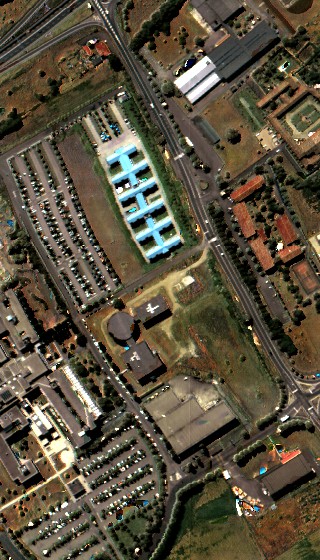}
				\label{fig:pavia:d}}
			\subfloat[NSSR]{\includegraphics[width=0.14\linewidth]{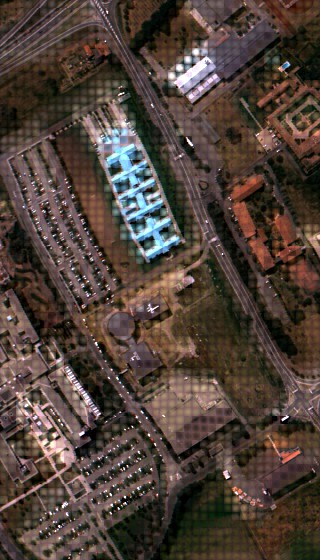}        
				\label{fig:pavia:e}}
			\subfloat[Integrated]{\includegraphics[width=0.14\linewidth]{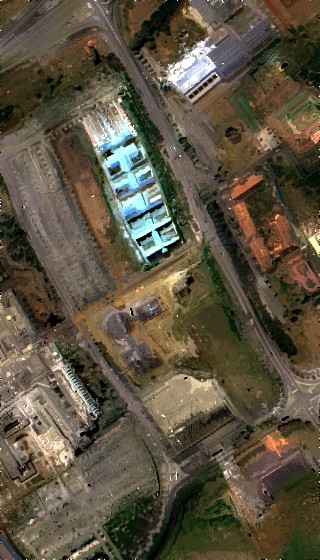}        
				\label{fig:pavia:f}}
			\subfloat[$u^2$-MDN]{\includegraphics[width=0.14\linewidth]{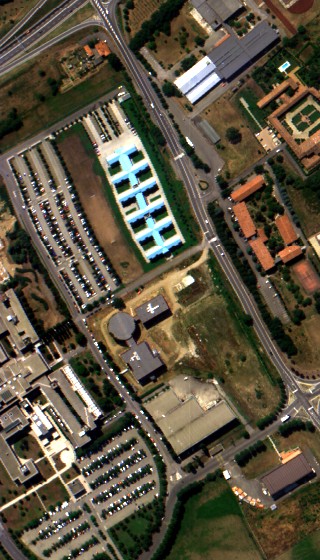}  
				\label{fig:pavia:g}}\hfill\\
			\subfloat[Distorted LR HSI]{\includegraphics[width=0.14\linewidth]{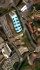}         
				\label{fig:pavia:a2}}
			\subfloat[Difference of GSA]{\includegraphics[width=0.14\linewidth]{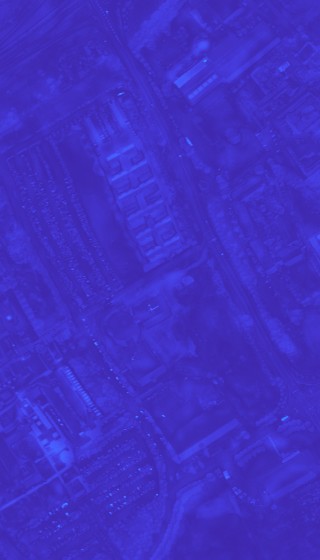}
				\label{fig:pavia:b2}}
			\subfloat[Difference of SFIM]{\includegraphics[width=0.14\linewidth]{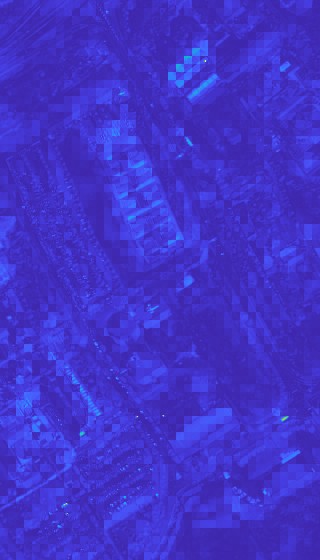}
				\label{fig:pavia:c2}}
			\subfloat[Difference of CNMF]{\includegraphics[width=0.14\linewidth]{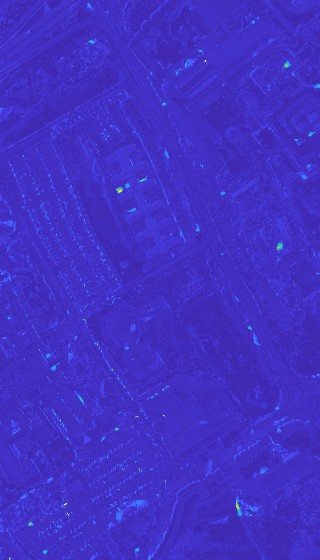}
				\label{fig:pavia:d2}}
			\subfloat[Difference of NSSR]{\includegraphics[width=0.14\linewidth]{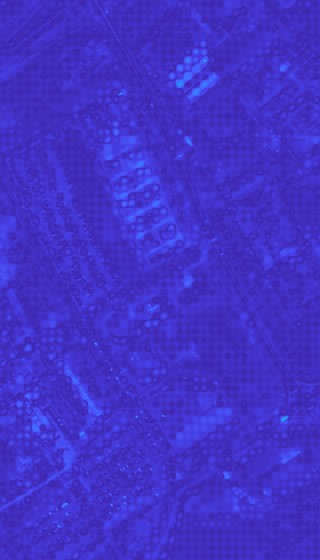}
				\label{fig:pavia:e2}}
			\subfloat[Difference of integrated]{\includegraphics[width=0.14\linewidth]{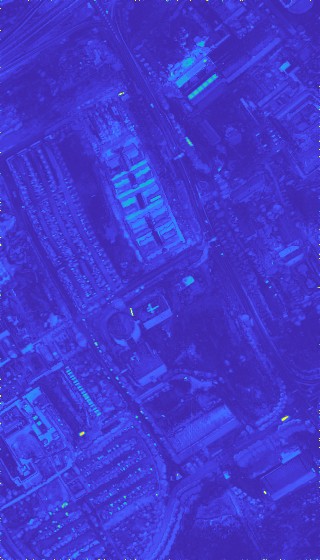}
				\label{fig:pavia:f2}}
			\subfloat[Difference of $u^2$-MDN]{\includegraphics[width=0.175\linewidth]{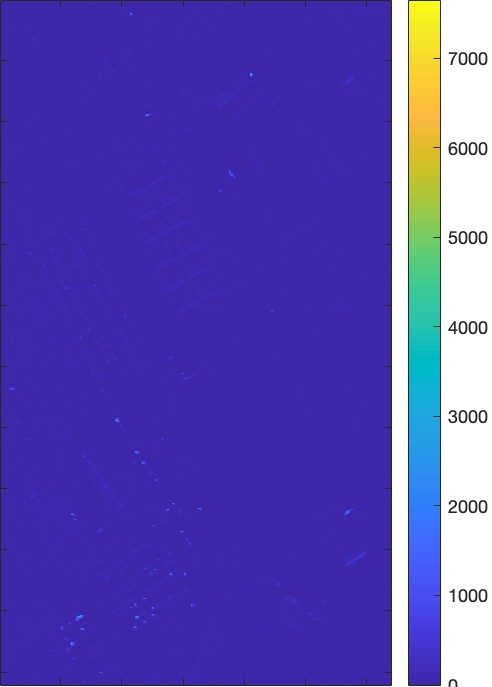}
				\label{fig:pavia:g2}}\hfill\\
			\subfloat[LR HSI]{\includegraphics[width=0.14\linewidth]{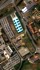}
				\label{fig:pavia:a3}}
			\subfloat[SAM of GSA]{\includegraphics[width=0.14\linewidth]{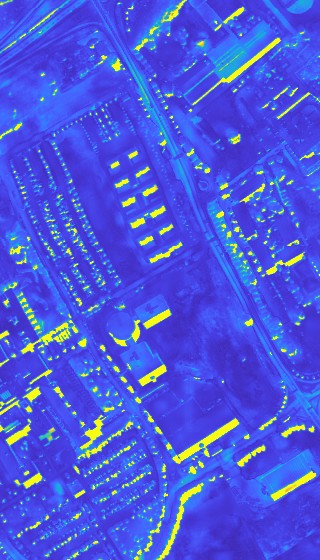}
				\label{fig:pavia:b3}}
			\subfloat[SAM of SFIM]{\includegraphics[width=0.14\linewidth]{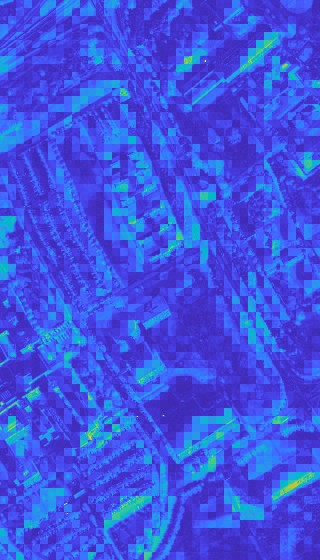}
				\label{fig:pavia:c3}}
			\subfloat[SAM of CNMF]{\includegraphics[width=0.14\linewidth]{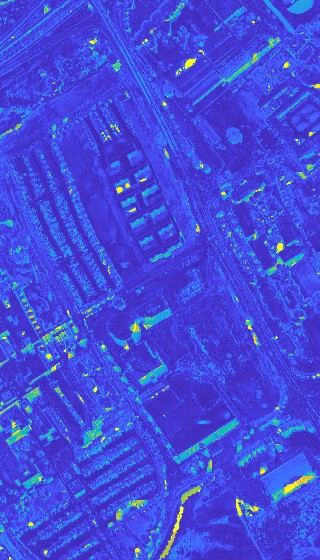}
				\label{fig:pavia:d3}}
			\subfloat[SAM of NSSR]{\includegraphics[width=0.14\linewidth]{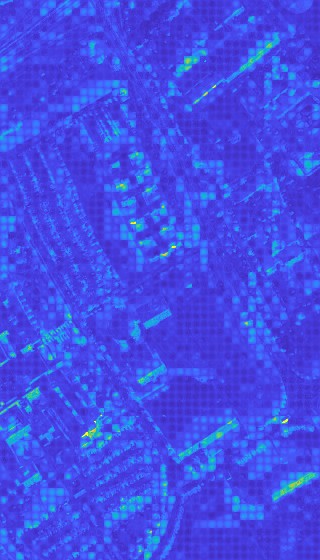}
				\label{fig:pavia:e3}}
			\subfloat[SAM of Integrated]{\includegraphics[width=0.14\linewidth]{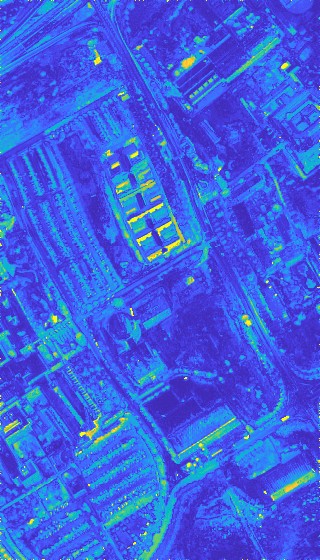}
				\label{fig:pavia:f3}}
			\subfloat[SAM of $u^2$-MDN]{\includegraphics[width=0.17\linewidth]{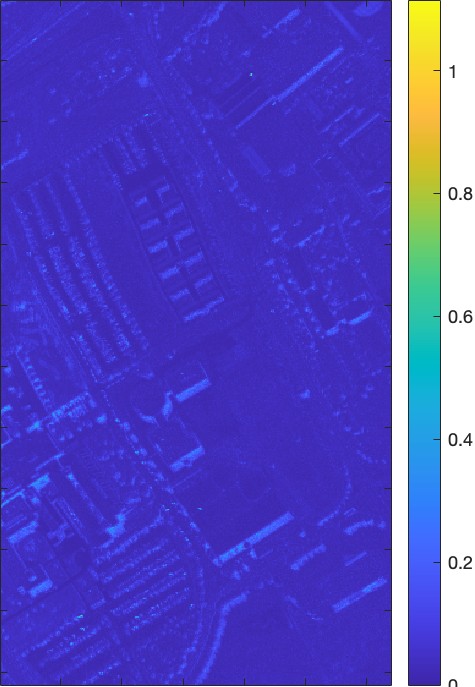}
				\label{fig:pavia:g3}}\hfill\\
		\end{minipage}
	\end{center}
	\caption{Reconstructed results given unregistered nonrigid distorted image pairs from the Pavia dataset. (a) Color composite of the reference HR HSI. (h) Color composite of the distorted LR HSI. (o) Color composite of the LR HSI. (b)-(g): reconstructed results. (i)-(n): average absolute difference between the reconstructed HSI and reference HSI over different spectral bands, from different methods. (p)-(u) SAM of each pixel between the reconstructed HSI and reference HSI from different methods.}
	\label{fig:pavia}
\end{figure*}

To visualize the reconstructed results for unregistered image pairs, we show the color composition of the reconstructed HR HSI in Figs.~\ref{fig:pom}-\ref{fig:pavia}, among which Figs.~\ref{fig:pom},~\ref{fig:img1} demonstrate the results of the rigid distorted image pairs, while  Figs.~\ref{fig:chiku},~\ref{fig:pavia} demonstrate the results of the nonrigid distorted image pairs. The first column of each figure presents the reference HR HSI, the distorted LR HSI, and the original LR HSI in (a), (h), and (o), respectively. The first through third rows show the reconstructed images, the absolute difference, and the spectral map of the results from different methods. We can observe that most approaches could not handle unregistered images pairs with large displacement well. The reconstructed results from SFIM have some blocking artifacts in most scenarios. The integrated fusion method has some smear effects on the reconstructed images due to the large displacement as shown in Fig.~\ref{fig:pom:f} and~\ref{fig:pavia:f}.  NSSR fails on the remote sensing datasets as shown in Figs.~\ref{fig:chiku:e} and~\ref{fig:pavia:e}, but it suffers relatively smaller spatial distortion on the benchmarked datasets. GSA could produce clear reconstructed images in most cases even though the images are unregistered, as shown in Figs.~\ref{fig:pom:b},~\ref{fig:chiku:b} and~\ref{fig:pavia:b}. This observation is consistent with the conclusions drawn in~\cite{baronti2011theoretical}. However, we observe from the SAM maps that, it suffers from spectral distortion. The CNMF method handles unregistered image pairs better than the other approaches as shown in Figs.~\ref{fig:pom:d},~\ref{fig:img1:d},~\ref{fig:chiku:d} and~\ref{fig:pavia:d}. But its performance is limited by the predefined down-sampling function. The effectiveness of the proposed method can be readily observed from the reconstructed results of difference images shown in Figs.~\ref{fig:pom:g},~\ref{fig:img1:g},~\ref{fig:chiku:g} and~\ref{fig:pavia:g}, where the proposed approach has much less spectral and spatial distortion as compared to the state-of-the-art, regardless of the type of input images.

\subsection{Experimental Results on Unregistered Real Image Pairs}

\begin{figure*}
\setlength{\abovecaptionskip}{0.cm}
\setlength{\belowcaptionskip}{0.cm}
	\centering
	\subfloat[Real LR HSI]{\includegraphics[width=0.33\linewidth]
			{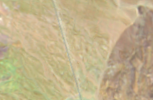}\label{fig:real:a}}\hspace{1mm}
	\subfloat[Real HR MSI]{\includegraphics[width=0.315\linewidth]
			{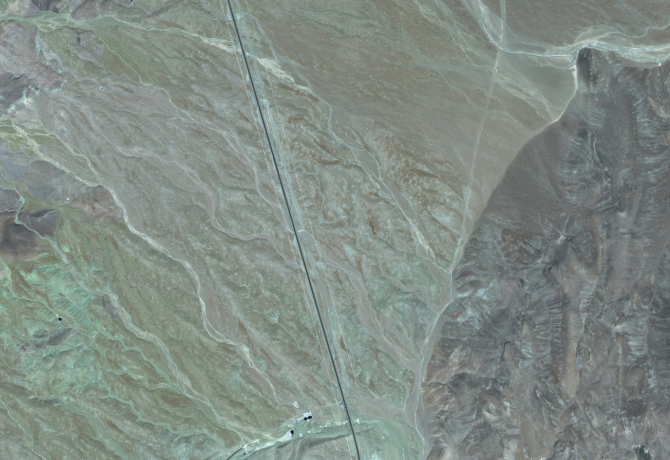}\label{fig:real:b}}\hspace{1mm}
	\subfloat[Reconstructed HR HSI]{\includegraphics[width=0.315\linewidth]
			{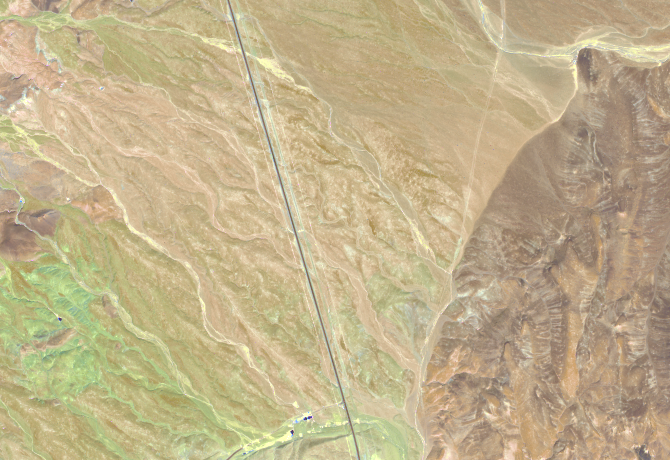}\label{fig:real:c}}\hfill
	\caption{Color composite of (a) the LR HSI of the real data from Hyperion, (b) the HR MSI of the real data from WorldView3 (images courtesy Maxar), and (c) the reconstructed HR HSI from the proposed method.}
	\label{fig:real}
\end{figure*}

We further evaluate the proposed method on the real unregistered image pairs with both rigid and nonrigid distortions. Since there is no ground truth HR HSI in real applications, we provide a visual inspection of the reconstructed results in Fig.~\ref{fig:real}. We can observe that, as long as the LR HSI includes all the spectral bases of HR MSI, the proposed method powered with mutual information is able to increase the spatial resolution of the LR HSI while preserving its spectral resolution well, even when the LR HSI and HR MSI have large pixel displacement.

\subsection{Ablation and Parameter Study}
Taking the challenging rotated `pompom' image from the CAVE dataset as an example, we further evaluate 1) the necessity of maximizing the mutual information between representations and input images and 2) the usage of collaborative $l_{2,1}$ loss. Since they are all designed to reduce the spectral distortion of the reconstructed image, we use SAM as the evaluation metric. 

\begin{figure*}[htb]
\setlength{\abovecaptionskip}{0.cm}
\setlength{\belowcaptionskip}{0.cm}
	\begin{center}
		\begin{minipage}{0.25\linewidth}
			{\includegraphics[width=1\linewidth]{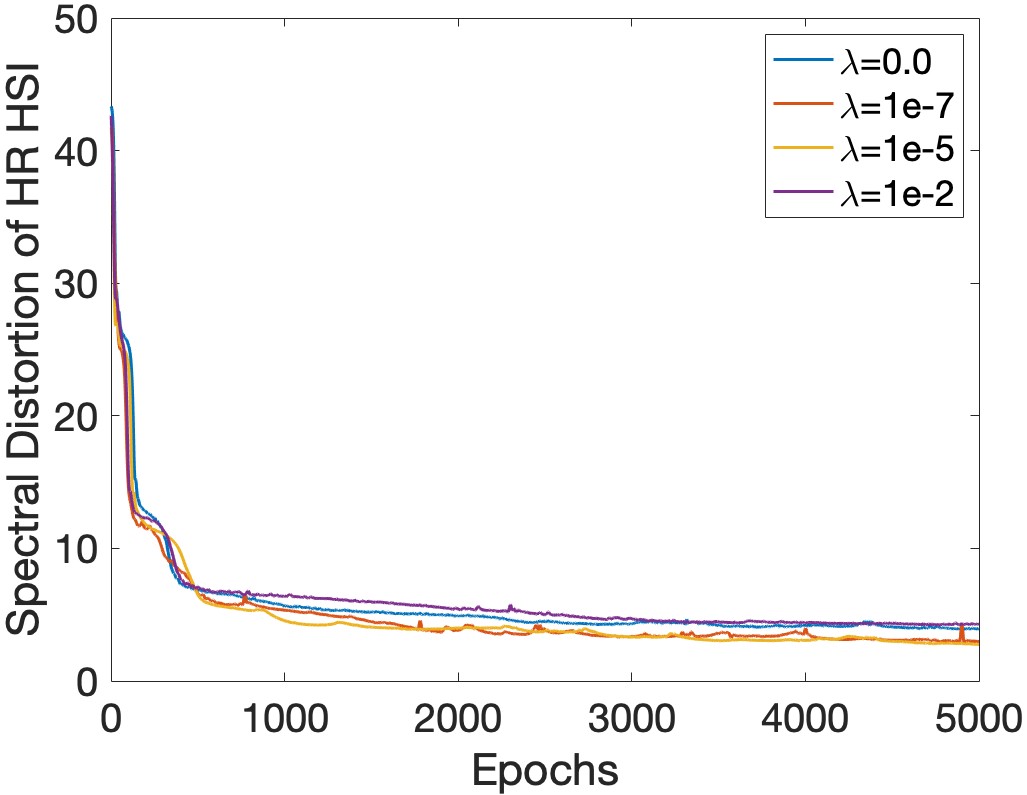}}
			\caption{Influence of MI}
			\label{fig:mutual}
		\end{minipage}
		\begin{minipage}{0.25\linewidth}
			{\includegraphics[width=1\linewidth]{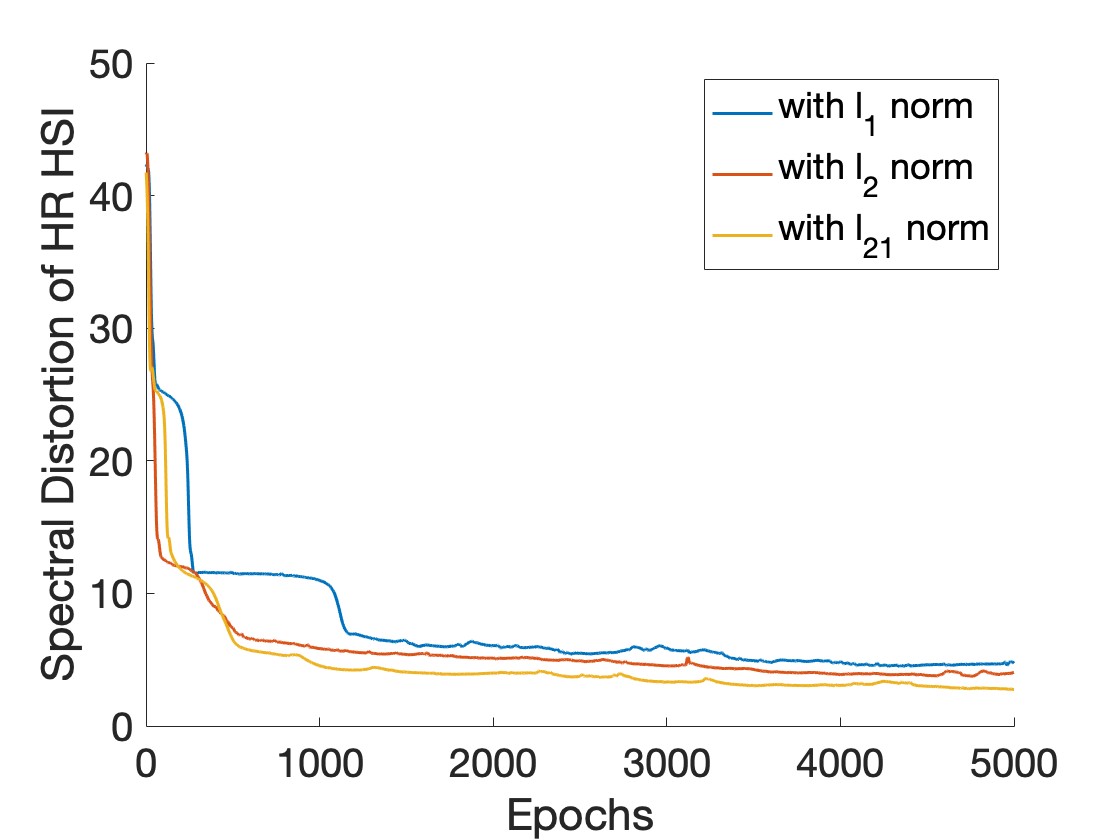}}\hfill
			\caption{The effect of $l_{2,1}$}
			\label{fig:l21}
		\end{minipage}
		\begin{minipage}{0.25\linewidth}
			{\includegraphics[width=1\linewidth]{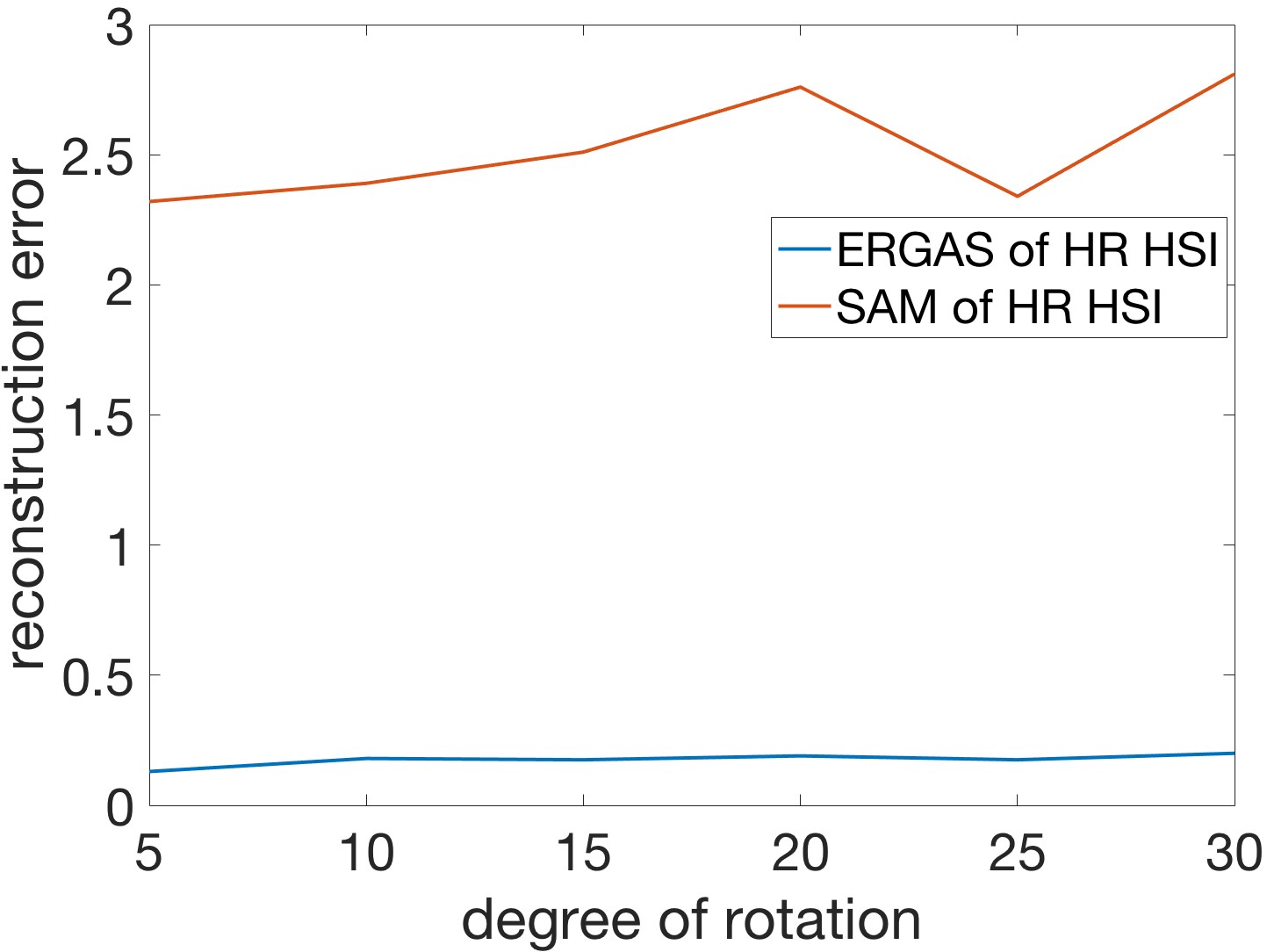}}\hfill
			\caption{Tolerance study}
			\label{fig:tol}
		\end{minipage}\hfill
	\end{center}
\end{figure*}

Fig. \ref{fig:mutual} illustrates the SAM of the reconstructed HR HSI when increasing the parameters of mutual information $\lambda$ in Eq.~\ref{equ:optall}. We can observe that, if there is no mutual information maximization, \ie $\lambda=0$, the spectral information would not be preserved well. When we gradually increase  $\lambda$, the reconstructed HR HSI preserves better spectral information, \ie, the SAM is largely reduced. The reason for that is, when we maximize the MI between the representations and their own inputs, it actually maximizes the mutual information of the representations of two modalities. Therefore, the network is able to correlate the extracted spectral and spatial information from unregistered HR MSI and LR MSI in an effective way, to largely reduce the spectral distortion. However, when the parameters are too large, it may hinder the reconstruction procedure of the image pairs. Therefore, we need to choose the proper parameters for the network. In our experiments, we keep $\mu=1\times 10^{-4}$ during the experiments to reduce over-fitting.We set $\lambda=1\times10^{-5}$ for general HSI dataset with less spectral bands and $\lambda=1\times10^{-1}$ for remote sensing HSI with more spectral bands.

The effectiveness evaluation of the collaborative $l_{2,1}$ norm is demonstrated in Fig.~\ref{fig:l21}. We can observe that with $l_1$ norm, the network converges much slower as compared to those using the $l_2$ norm and $l_{21}$ norm, and the $l_{21}$ norm converges to smaller spectral distortions than using the $l_2$ norm or the $l_1$ norm. Thus, $l_{2,1}$ norm can preserve the spectral information better and significantly reduce the spectral distortion of the restored HR HSI.

\subsection{Tolerance Study}
At last, we would like to examine how much spectral information can be preserved when the network deals with unregistered images. To preserve spectral information, the input LR HSI should cover all the spectral signatures of HR MSI. Thus, we choose the image in Fig.~\ref{fig:sample} from the Harvard dataset which has most of the spectral signatures centered in the image. The results are shown in Fig.~\ref{fig:tol}. The image is rotated from 5 degrees to 30 degrees with 15\% to 48\% percent of information missing. We can observe that as long as the spectral bases are included in the LR HSI, no matter how small the overlapped region is between the LR HSI and HR MSI, we could always achieve the reconstructed image with small spectral distortion even for unregistered input images. 

\section{Conclusion}
We proposed an unsupervised encoder-decoder network $u^2$-MDN to solve the problem of hyperspectral image super-resolution without multi-modality registration. The unique structure stabilizes the network training by projecting both modalities into the same space and extracting the spectral basis from LR HSI with rich spectral information as well as spatial representations from HR MSI with high-resolution spatial information simultaneously. The network learns correlated spatial information from two unregistered modalities by maximizing the mutual information between the representations and their own raw inputs. In this way, it maximizes the MI between the two representations that largely reduces the spectral distortion. In addition, the collaborative $l_{2,1}$ norm is adopted to encourage the network to further preserve spectral information. Extensive experiments on two benchmark datasets demonstrated the superiority of the proposed approach over the state-of-the-art.

\section*{Acknowledgment}
The authors would like to thank all the developers of the evaluated methods who kindly offered their codes,  and Dr. Danfeng Hong and Dr. Ke Zhang who provided  suggestions on synthetic data generation. This publication was made possible by NASA grant NNX12CB05C and NNX16CP38P. %The statements made herein are solely the responsibility of the authors. 

\ifCLASSOPTIONcaptionsoff
  \newpage
\fi



%\bibliographystyle{IEEEtran}
%\begin{thebibliography}{1}
%  0.5em minus 0.4em\relax Harlow, England: Addison-Wesley, 1999.\\
%\end{thebibliography}

%\pagebreak
\bibliographystyle{IEEEtran}
\bibliography{refs}


%\begin{IEEEbiography}[{\includegraphics[width=1\linewidth]}{fig/yingqu}]{Ying Qu}
%\end{IEEEbiography}

\begin{IEEEbiography}[{\includegraphics[width=1in,keepaspectratio]{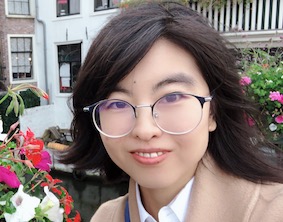}}]{Ying Qu} (S'16--M'18)
received the B.S. degree in automatics and M.S. degree in pattern recognition \& artificial intelligence from Northeastern University, Shenyang, China in 2008 and 2010, respectively, and the Ph.D. degree in computer engineering from the University of Tennessee, Knoxville, in 2017. She is currently a research associate with the Department of Electrical Engineering and Computer Science at the University of Tennessee, Knoxville. Her current research interests are remote sensing, artificial intelligence and computer vision. She was the recipient of the Best Student Paper Awards at The International Geoscience and Remote Sensing Symposium (IGARSS) in 2016.
\end{IEEEbiography}\vfill

\begin{IEEEbiography}[{\includegraphics[width=1in,keepaspectratio]{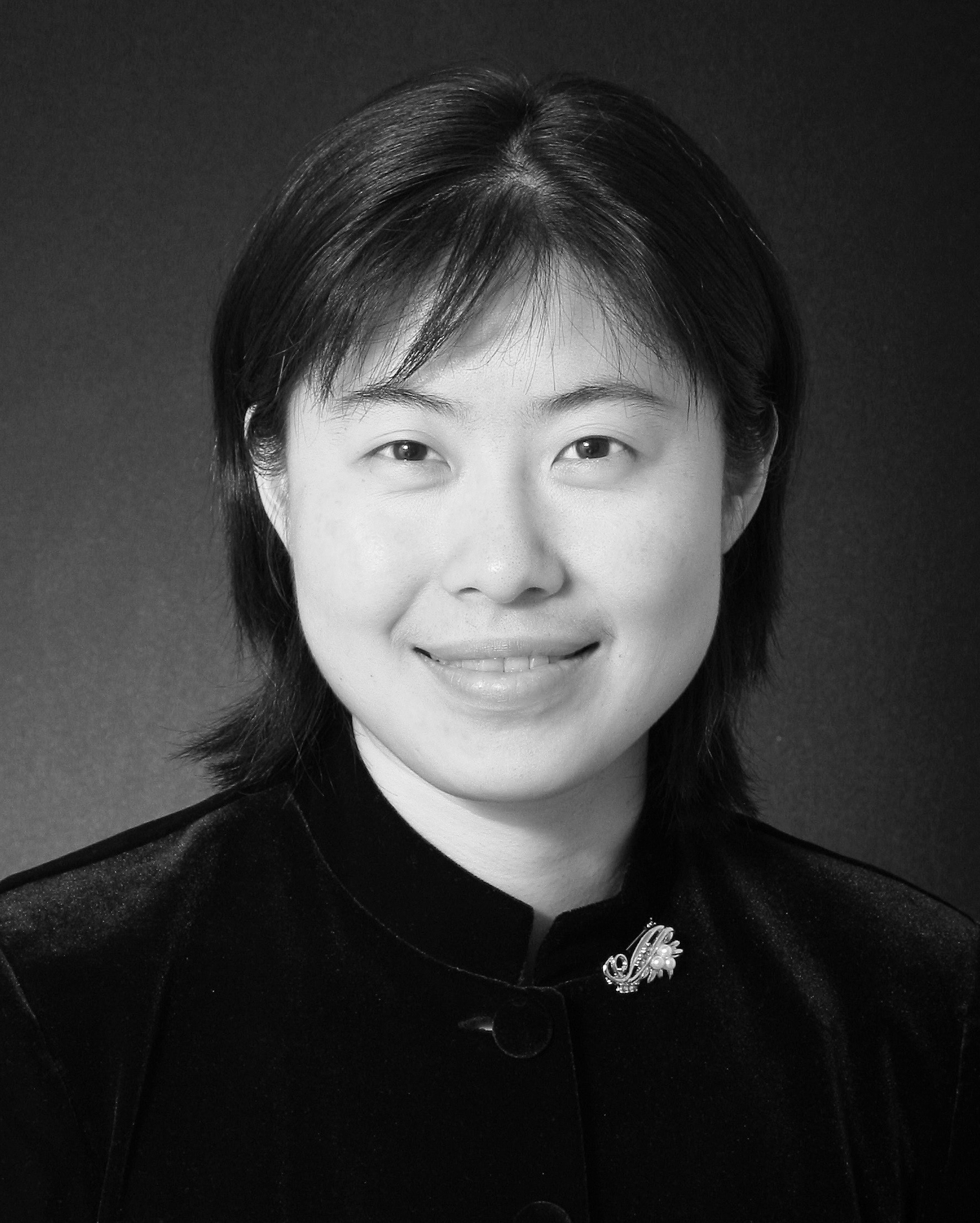}}]{Hairong Qi} (IEEE Fellow since 2017) received the B.S. and M.S. degrees in computer science from Northern JiaoTong University, Beijing, China in 1992 and 1995, respectively, and the Ph.D. degree in computer engineering from North Carolina State University, Raleigh, in 1999. She is currently the Gonzalez Family Professor with the Department of Electrical Engineering and Computer Science at the University of Tennessee, Knoxville. Her current research interests are in advanced imaging and collaborative processing in resource-constrained distributed environment, hyperspectral image analysis, and automatic target recognition. Dr. Qi's research is supported by National Science Foundation (NSF), DARPA, Office of Naval Research (ONR), Department of Homeland Security (DHS), U.S. Army Space and Missile Defense Command, and U.S. Army Medical Research and Materiel Command. Dr. Qi is the recipient of the NSF CAREER Award. She also received the Best Paper Awards at the 18th International Conference on Pattern Recognition (ICPR) in 2006, the 3rd ACM/IEEE International Conference on Distributed Smart Cameras (ICDSC) in 2009, and IEEE Workshop on Hyperspectral Image and Signal Processing: Evolution in Remote Sensor (WHISPERS) in 2015. She is awarded the Highest Impact Paper from the IEEE Geoscience and Remote Sensing Society in 2012.
\end{IEEEbiography}\vfill

\begin{IEEEbiography}[{\includegraphics[width=1in,keepaspectratio]{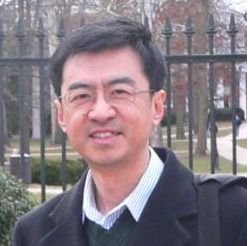}}]{Chiman Kwan}
(S’85-M’93-SM’98) received his BS with honors in Electronics from the Chinese University of Hong Kong in 1988, and MS and Ph.D. degrees in electrical engineering from the University of Texas at Arlington in 1989 and 1993, respectively. Currently, he is the Chief Technology Officer of Signal Processing, Inc. and Applied Research LLC, leading research and development efforts in chemical agent detection, biometrics, speech processing, image fusion, mission planning, and fault diagnostics and prognostics. His primary research areas include robust and adaptive control methods, signal and image processing, communications, neural networks, and pattern recognition applications. 
From April 1991 to February 1994, he worked in the Beam Instrumentation Department of the SSC (Superconducting Super Collider Laboratory) in Dallas, Texas, where he was heavily involved in the modeling, simulation and design of modern digital controllers and signal processing algorithms for the beam control and synchronization system. He later joined the Automation and Robotics Research Institute in Fort Worth, where he applied intelligent control methods such as neural networks and fuzzy logic to the control of power systems, robots, and motors. Between July 1995 and April 2006, he was with Intelligent Automation, Inc. in Rockville, Maryland. He has served as Principal Investigator/Program Manager for more than 120 different projects. Dr. Kwan has 15 patents, 65 invention disclosures, more than 120 papers in archival journals and more than 250 additional papers published in major conference proceedings. He is listed in the New Millennium edition of Who’s Who in Science and Engineering and is a member of Tau Beta Pi. He also received several awards from IEEE related to fault diagnostics and prognostics, and a certificate of recognition from NASA for the health monitoring of Auxiliary Power Units in the Space Shuttle. 
\end{IEEEbiography}\vfill

\begin{IEEEbiography}[{\includegraphics[width=1in,keepaspectratio]{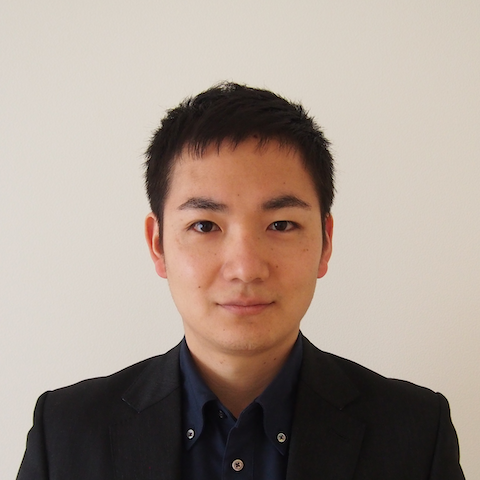}}]{Naoto Yokoya} (S'10--M'13)
received the M.Eng. and Ph.D. degrees from the Department of Aeronautics and Astronautics, the University of Tokyo, Tokyo, Japan, in 2010 and 2013, respectively. He is currently a Lecturer at the University of Tokyo and a Unit Leader at the RIKEN Center for Advanced Intelligence Project, Tokyo, Japan, where he leads the Geoinformatics Unit. He was an Assistant Professor at the University of Tokyo from 2013 to 2017. In 2015-2017, he was an Alexander von Humboldt Fellow, working at the German Aerospace Center (DLR), Oberpfaffenhofen, and Technical University of Munich (TUM), Munich, Germany. His research is focused on the development of image processing, data fusion, and machine learning algorithms for understanding remote sensing images, with applications to disaster management. Dr. Yokoya won the first place in the 2017 IEEE Geoscience and Remote Sensing Society (GRSS) Data Fusion Contest organized by the Image Analysis and Data Fusion Technical Committee (IADF TC). He is the Chair (2019-2021) and was a Co-Chair (2017-2019) of IEEE GRSS IADF TC and also the secretary of the IEEE GRSS All Japan Joint Chapter since 2018. He is an Associate Editor for the IEEE Journal of Selected Topics in Applied Earth Observations and Remote Sensing (JSTARS) since 2018. He is/was a Guest Editor for the IEEE JSTARS in 2015-2021, for Remote Sensing in 2016-2021, and for the IEEE Geoscience and Remote Sensing Letters (GRSL) in 2018-2019.
\end{IEEEbiography}\vfill

\begin{IEEEbiography}[{\includegraphics[width=1in,keepaspectratio]{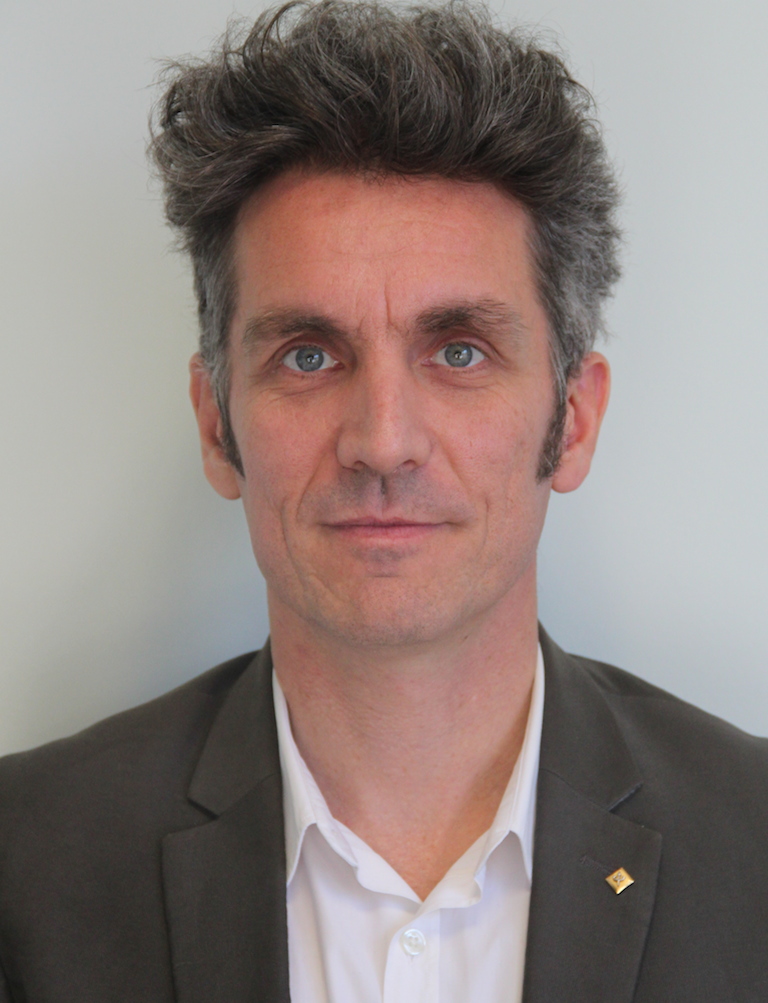}}]{Jocelyn Chanussot} (M'04–SM'04–F'12)
received the M.Sc. degree in electrical engineering from the Grenoble Institute of Technology (Grenoble INP), Grenoble, France, in 1995, and the Ph.D. degree from the Université de Savoie, Annecy, France, in 1998. Since 1999, he has been with Grenoble INP, where he is currently a Professor of signal and image processing. His research interests include image analysis, hyperspectral remote sensing, data fusion, machine learning and artificial intelligence. He has been a visiting scholar at Stanford University (USA), KTH (Sweden) and NUS (Singapore). Since 2013, he is an Adjunct Professor of the University of Iceland. In 2015-2017, he was a visiting professor at the University of California, Los Angeles (UCLA).  He holds the AXA chair in remote sensing and is an Adjunct professor at the Chinese Academy of Sciences, Aerospace Information research Institute, Beijing.
Dr. Chanussot is the founding President of IEEE Geoscience and Remote Sensing French chapter (2007-2010) which received the 2010 IEEE GRSS Chapter Excellence Award. He has received multiple outstanding paper awards. He was the Vice-President of the IEEE Geoscience and Remote Sensing Society, in charge of meetings and symposia (2017-2019). He was the General Chair of the first IEEE GRSS Workshop on Hyperspectral Image and Signal Processing, Evolution in Remote sensing (WHISPERS). He was the Chair (2009-2011) and  Cochair of the GRS Data Fusion Technical Committee (2005-2008). He was a member of the Machine Learning for Signal Processing Technical Committee of the IEEE Signal Processing Society (2006-2008) and the Program Chair of the IEEE International Workshop on Machine Learning for Signal Processing (2009). He is an Associate Editor for the IEEE Transactions on Geoscience and Remote Sensing, the IEEE Transactions on Image Processing and the Proceedings of the IEEE. He was the Editor-in-Chief of the IEEE Journal of Selected Topics in Applied Earth Observations and Remote Sensing (2011-2015). In 2014 he served as a Guest Editor for the IEEE Signal Processing Magazine. He is a Fellow of the IEEE, a member of the Institut Universitaire de France (2012-2017) and a Highly Cited Researcher (Clarivate Analytics/Thomson Reuters, since 2018).
\end{IEEEbiography}\vfill

%\newpage



%\vfill


\end{document}